\documentclass{article}

\usepackage{arxiv}

\usepackage[utf8]{inputenc} 
\usepackage[T1]{fontenc}    
\usepackage{hyperref}       
\usepackage{url}            
\usepackage{booktabs}       
\usepackage{amsfonts}       
\usepackage{nicefrac}       
\usepackage{microtype}      
\usepackage{lipsum}		
\usepackage{graphicx}
\usepackage{doi}
\usepackage{caption}
\usepackage{listings}
\usepackage{xcolor}
\usepackage{fancyhdr}       
\usepackage{graphicx}       
\usepackage{fancyvrb}        
\usepackage{pdfpages}
\usepackage{hyperref} 
\usepackage{caption} 

\usepackage{setspace} 

\usepackage{soul}
\usepackage{amsmath}
\usepackage{listings}
\usepackage{caption}

\usepackage{multirow}
\usepackage{helvet} 
\usepackage{tikz}
\usetikzlibrary{shapes.geometric, arrows, arrows.meta, positioning,fit,backgrounds, shadows}
\usepackage{enumitem}

\usetikzlibrary{decorations.pathreplacing}

\newcounter{textbox}
\renewcommand{\thetextbox}{\arabic{textbox}} 

\newcommand{\textboxcaption}[1]{
  \refstepcounter{textbox}
  \noindent\small\textbf{Text Box \thetextbox:} #1\par\medskip
}

\usepackage{sansmath} 

\tikzstyle{process} = [rectangle, rounded corners, minimum width=3cm, minimum height=1cm, text centered, draw=black, fill=orange!30]
\tikzstyle{start} = [ellipse, minimum width=3cm, minimum height=1cm, text centered, draw=black, fill=yellow!30]
\tikzstyle{decision} = [diamond, minimum width=3cm, minimum height=1cm, text centered, draw=black, fill=green!30]
\tikzstyle{arrow} = [thick,->,>=stealth]

\tikzstyle{phase} = [rectangle, rounded corners, minimum width=3cm, minimum height=1cm, text centered, draw=black, fill=orange!30]
\tikzstyle{result} = [ellipse, minimum width=3cm, minimum height=1cm, text centered, draw=black, fill=green!30]

\tikzstyle{dataset} = [rectangle, rounded corners, minimum width=2cm, minimum height=1cm, text centered, draw=black, fill=blue!30]
\tikzstyle{masking} = [rectangle, rounded corners, minimum width=2cm, minimum height=1cm, text centered, draw=black, fill=red!30]

\usepackage[skins,breakable]{tcolorbox}

\definecolor{aigold}{RGB}{244,210, 1} 
\definecolor{aigreen}{RGB}{210,244,211} 

\sethlcolor{aigreen}

\definecolor{aired}{RGB}{255,180,181}

\definecolor{aigold}{RGB}{255,180,181}

\definecolor{aiblue}{RGB}{173,216,230} 
 
\definecolor{lightred}{rgb}{1,0.9,0.9} 

\newcommand{\hlred}[1]{{%
  \sethlcolor{lightred}
  \hl{#1}
}}

\tcbset{
  customboxsmalll/.style={
    width=0.95\textwidth,  
    fonttitle=\ttfamily\scriptsize, 
    colback=white,  
    colframe=black,  
    colbacktitle=blue!75!black, 
    coltitle=white,  
    boxrule=0.5pt,  
    rounded corners,  
    boxed title style={
        colback=blue!75!black, 
        colframe=black, 
        boxrule=0.5pt  
            before={\setlength{\parskip}{0pt} \setlength{\parindent}{-6pt}},  
    after={\setlength{\parskip}{0pt} \setlength{\parindent}{-6pt}},   
    halign=center,  
    before skip=-5pt,  
    after skip=-5pt,   
    innertop=-3pt,     
    innerbottom=-3pt,  
    valign=center,    
    }
    
  }
}

\tcbset{
  custombox/.style={
    width=512.18663pt,
    top=3pt,
    fonttitle=\bfseries\scriptsize\sffamily, 
    colback=white,
    colframe=black,
    colbacktitle=black,
    boxrule=0.5pt, 
    colback=white, 
    enhanced,
    rounded corners, 
    center,
    boxed title style={
            sharp corners,
            size=small,
            colframe=blue!75!black, 
        },
    attach boxed title to top left={yshift=-0.1in,xshift=0.15in},
    boxed title style={boxrule=0pt,colframe=white,},
    before=\setlength{\baselineskip}{2pt}, 
    before upper=\setlength{\baselineskip}{2pt}, 
    after=\setlength{\baselineskip}{2pt}, 
  }
} 

\newtcolorbox{LLMbox}[2][]{custombox, title=#2,#1}
\newtcolorbox{LLMboxSmall}[2][]{customboxsmalll, title=#2,#1}

\DefineVerbatimEnvironment{thinkingcolor}{Verbatim}{
  formatcom=\color{blue}, 
  fontsize=\scriptsize  
}
\DefineVerbatimEnvironment{reflectcolor}{Verbatim}{
  formatcom=\color{orange}, 
  fontsize=\scriptsize  
}

\title{\textbf{Agentic Deep Graph Reasoning Yields Self-Organizing Knowledge Networks}
}

\author{ \href{https://orcid.org/0000-0002-4173-9659}{\includegraphics[scale=0.06]{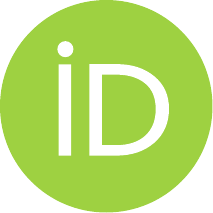}\hspace{1mm}Markus J. Buehler}\thanks{Corresponding author.} \\
	Laboratory for Atomistic and Molecular Mechanics\\Center for Computational Science and Engineering \\
        Schwarzman College of Computing \\
	Massachusetts Institute of Technology\\
	Cambridge, MA 02139, USA \\
        \\
	\texttt{mbuehler@MIT.EDU} \\
}

\newtcbox{\mybox}[1][green]{on line,
arc=0pt,outer arc=0pt,colback=#1!10!white,colframe=#1!50!black,
boxsep=0pt,left=0pt,right=0pt,top=0pt,bottom=0pt,
boxrule=0pt,bottomrule=0pt,toprule=0pt}

\tcbset{
  customboxmultipage/.style={
    width=512.18663pt,
    top=3pt,
    fonttitle=\ttfamily\scriptsize, 
    colback=white,
    breakable,
    colframe=black,
    colbacktitle=black,
    boxrule=0.5pt,
    rounded corners,
    center,
    boxed title style={
        sharp corners,
        size=small,
        colframe=blue!75!black,
    },
    attach boxed title to top left={yshift=-0.1in,xshift=0.15in},
    boxed title style={boxrule=0pt,colframe=white,},
    before={\par\noindent},
    after={\par},
    before upper={\setstretch{0.3}}, 
    after lower={\nointerlineskip},
    every page on break/.style={
      interior style={top color=white, bottom color=white},
    },
  }
}

\newtcolorbox{LLMboxmultipage}[2][]{customboxmultipage,title=#2,#1}

\renewcommand{\headeright}{}
\renewcommand{\shorttitle}{Agentic Deep Graph Reasoning }

\hypersetup{
pdftitle={
Autonomous Knowledge Gardening at Scale},
pdfsubject={q-bio.NC, q-bio.QM},
pdfauthor={Markus J. Buehler, MIT, mbuehler@MIT.EDU},
pdfkeywords={Language Modeling, Graph Theory, Category Theory, Materials Science, Materiomics, Symbolic Reasoning},
}

\begin{document}
\maketitle

\begin{abstract} 


We present an agentic, autonomous graph expansion framework that iteratively structures and refines knowledge \textit{in situ}. Unlike conventional knowledge graph construction methods relying on static extraction or single-pass learning, our approach couples a reasoning-native large language model with a continually updated graph representation. At each step, the system actively generates new concepts and relationships, merges them into a global graph, and formulates subsequent prompts based on its evolving structure. Through this feedback-driven loop, the model organizes information into a scale-free network characterized by hub formation, stable modularity, and bridging nodes that link disparate knowledge clusters. Over hundreds of iterations, new nodes and edges continue to appear without saturating, while centrality measures and shortest path distributions evolve to yield increasingly distributed connectivity. Our analysis reveals emergent patterns—such as the rise of highly connected ``hub'' concepts and the shifting influence of ``bridge'' nodes—indicating that agentic, self-reinforcing graph construction can yield open-ended, coherent knowledge structures. Applied to materials design problems, we present compositional reasoning experiments by extracting node-specific and synergy-level principles to foster genuinely novel knowledge synthesis, yielding cross-domain ideas that transcend rote summarization and strengthen the framework’s potential for open-ended scientific discovery. We discuss other applications in scientific discovery and outline future directions for enhancing scalability and interpretability.
\end{abstract}


\section{Introduction}

Scientific inquiry often proceeds through an interplay of incremental refinement and transformative leaps, evoking broader questions of how knowledge evolves under continual reflection and questioning. In many accounts of discovery, sustained progress arises not from isolated insights but from an iterative process in which prior conclusions are revisited, expressed as generalizable ideas, refined, or even reorganized as new evidence and perspectives emerge~\cite{kuhn1962structure}. Foundational work in category theory has formalized aspects of this recursive structuring, showing how hierarchical representations can unify diverse knowledge domains and enable higher-level abstractions in both the natural and social sciences \cite{Spivak2011CategoryNetworks,Giesa2011ReoccurringAnalogies,Giesa2012CategoryDesign}. Across engineering disciplines including materials science, such iterative integration of information has proven essential in synthesizing deeply interlinked concepts. 

Recent AI methods, however, often emphasize predictive accuracy and single-step outputs over the layered, self-reflective processes that characterize human problem-solving. Impressive gains in natural language processing, multimodal reasoning \cite{Vaswani2017AttentionNeed,AlecRadfordImprovingPre-Training,Xue2021ByT5:Models,Jiang2023Mistral7Bb,Phi-2:Research,dubey2024llama3herdmodels,Brown2020LanguageLearners,salinas2025exoplanettransitcandidateidentification}, and materials science \cite{Schmidt2019MLMaterials,Buehler2024X-LoRA:Design,Arevalo2023LearningMaterials,Hu2023DeepScience,D1MH00495F}, including breakthroughs in molecular biology \cite{Vamathevan2019ApplicationsDevelopment} and protein folding \cite{Jumper2021HighlyAlphaFold,ProteinTrRosetta,Wu2022High-resolutionSequencelocal}, showcase the prowess of large-scale models trained on vast datasets. Yet most of the early systems generate answers in a single pass, sidestepping the symbolic, stepwise reasoning that often underpins scientific exploration. This gap has prompted a line of research into modeling that explicitly incorporates relational modeling, reflection or multi-step inferences \cite{Spivak2011CategoryNetworks,Giesa2011ReoccurringAnalogies,Giesa2012CategoryDesign,abbott2024flashattentionnapkindiagrammaticapproach,Buehler2025GraphAwareGPT,Miconi2025,o1-model-card-2024,buehler2024preflexorpreferencebasedrecursivelanguage,buehler2025insitugraphreasoningknowledge,reddy2024towards}, hinting at a transition from single-shot pattern recognition to more adaptive synthesis of answers from first principles in ways that more closely resemble compositional mechanisms. Thus, a fundamental challenge now is how can we build scientific AI systems that synthesize information rather than memorizing it. 

Graphs offer a natural substrate for this kind of iterative knowledge building. By representing concepts and their relationships as a network, it becomes possible to capture higher-order structure—such as hubs, bridging nodes, or densely interconnected communities—that might otherwise remain implicit. This explicit relational format also facilitates systematic expansion: each newly added node or edge can be linked back to existing concepts, reshaping the network and enabling new paths of inference~\cite{BuehlerGraphReasoning2024,Buehler2025GraphAwareGPT,buehler2025insitugraphreasoningknowledge}. Moreover, graph-based abstractions can help large language models move beyond memorizing discrete facts; as nodes accumulate and form clusters, emergent properties may reveal cross-domain synergies or overlooked gaps in the knowledge space.

Recent work suggests that standard Transformer architectures can be viewed as a form of Graph Isomorphism Network (GIN), where attention operates over relational structures rather than raw token sequences~\cite{Buehler2025GraphAwareGPT}. Under this lens, each attention head effectively tests for isomorphisms in local neighborhoods of the graph, offering a principled way to capture both global and local dependencies. A category-theoretic perspective further bolsters this approach by providing a unified framework for compositional abstractions: nodes and edges can be treated as objects and morphisms, respectively, while higher-level concepts emerge from functorial mappings that preserve relational structure \cite{Spivak2011CategoryNetworks,Giesa2011ReoccurringAnalogies,Giesa2012CategoryDesign}. Taken together, these insights hint at the potential for compositional capabilities in AI systems, where simpler building blocks can be combined and reconfigured to form increasingly sophisticated representations, rather than relying on one-pass computations or static ontologies. By using graph-native modeling and viewing nodes and edges as composable abstractions, such a model may be able to recognize and reapply learned configurations in new contexts—akin to rearranging building blocks to form unanticipated solutions. This compositional approach, strengthened by category-theoretic insights, allows the system to not only interpolate among known scenarios but to extrapolate to genuinely novel configurations. In effect, graph-native attention mechanisms treat interconnected concepts as first-class entities, enabling the discovery of new behaviors or interactions that purely sequence-based methods might otherwise overlook.

A fundamental challenge remains: How can we design AI systems that, rather than merely retrieving or matching existing patterns, build and refine their own knowledge structures across iterations. Recent work proposes that graphs can be useful strategies to endow AI models with relational capabilities\cite{BuehlerGraphReasoning2024,Buehler2025GraphAwareGPT,buehler2025insitugraphreasoningknowledge} both within the framework of creating graph-native attention mechanisms and by training models to use graphs as native abstractions during learned reasoning phases. Addressing this challenge requires not only methods for extracting concepts but also mechanisms for dynamically organizing them so that new information reshapes what is already known. By endowing large language models with recursively expanding knowledge graph capabilities, we aim to show how stepwise reasoning can support open-ended discovery and conceptual reorganization. The work presented here explores how such feedback-driven graph construction may lead to emergent, self-organizing behaviors, shedding light on the potential for truly iterative AI approaches that align more closely with the evolving, integrative nature of human scientific inquiry. Earlier work on graph-native reasoning has demonstrated that models explicitly taught how to reason in graphs and abstractions can lead to systems that generalize better and are more interpretable~\cite{buehler2025insitugraphreasoningknowledge}. 

Here we explore whether we can push this approach toward ever-larger graphs, creating extensive \textit{in situ} graph reasoning loops where models spend hours or days developing complex relational structures before responding to a task. Within such a vision, several key issues arise: Will repeated expansions naturally preserve the network’s relational cohesion, or risk splintering into disconnected clusters? Does the continuous addition of new concepts and edges maintain meaningful structure, or lead to saturation and redundancy? And to what extent do bridging nodes, which may initially spark interdisciplinary links, remain influential over hundreds of iterations? In the sections ahead, we investigate these questions by analyzing how our recursively expanded knowledge graphs grow and reorganize at scale—quantifying hub formation, modular stability, and the persistence of cross-domain connectors. Our findings suggest that, rather than collapsing under its own complexity, the system retains coherent, open-ended development, pointing to new possibilities for large-scale knowledge formation in AI-driven research for scientific exploration.

\begin{figure}
\centering

\sffamily
\scriptsize


\begin{tikzpicture}[
    node distance=0.6cm and 1cm, auto,
    task/.style={rectangle, draw, fill=blue!20, text width=5cm, text centered, rounded corners, minimum height=1.2cm},
    generation/.style={rectangle, draw, fill=cyan!30, text width=5cm, text centered, rounded corners, minimum height=1.2cm},
    extraction/.style={rectangle, draw, fill=violet!40, text width=5cm, text centered, rounded corners, minimum height=1.2cm},
    merge_graph/.style={rectangle, draw, fill=violet!20, text width=5cm, text centered, rounded corners, minimum height=1.5cm},
    question/.style={rectangle, draw, fill=green!20, text width=5cm, text centered, rounded corners, minimum height=1.5cm},
    visualization/.style={rectangle, draw, fill=yellow!30, text width=5cm, text centered, rounded corners, minimum height=1.2cm},
    final_output/.style={rectangle, draw, fill=orange!30, text width=5cm, text centered, rounded corners, minimum height=1.2cm},
    line/.style={draw, -stealth', shorten >=1pt},
    dashed line/.style={draw, dashed, shorten >=1pt}]

    \node [task] (start) {\textbf{Define Initial Question} \\ (Broad question or specific topic, e.g., "Impact-Resistant Materials")};
    \node [generation, below=of start] (generate) {\textbf{Generate Graph-native \\ Reasoning Tokens} \\ \texttt{<|thinking|> ... <|/thinking|>}};
    \node [extraction, below=of generate] (extract_graph) {\textbf{Parse Graph} \( \mathcal{G}_{\text{local}}^i \) \\ (Extract Nodes and Relations)};
    \node [merge_graph, below=of extract_graph] (merge_graph) {\textbf{Merge Extracted Graph with \\ Larger Graph} \\ (Append Newly Added Nodes/Edges) \\  $\mathcal{G} \leftarrow \mathcal{G} \cup \mathcal{G}_{\text{local}}^i$};
    \node [visualization, below=of merge_graph] (save_graph) {\textbf{Save and Visualize}} ;
    \node [final_output, below=of save_graph] (final_result) {\textbf{Final Integrated Graph} $\mathcal{G}$} ;

    \path [line] (start) -- (generate);
    \path [line] (generate) -- (extract_graph);
    \path [line] (extract_graph) -- (merge_graph);
    \path [line] (merge_graph) -- (save_graph);
    \path [line] (save_graph) -- (final_result);

    \node [question, right=of extract_graph, xshift=3cm] (new_question) {\textbf{Generate New Question} \\ 
    Based on Last Extracted Added Nodes/Edges as captured in $\mathcal{G}_{\text{local}}^i$ };
    \path [line] (merge_graph.east) -| (new_question.south);
    \path [line] (new_question.north) |- (generate.east);

    \begin{pgfonlayer}{background}
        \node [draw=orange, thick, dashed, fit=(generate) (extract_graph) (merge_graph) (save_graph) (new_question), 
               inner sep=0.4cm, label={[text=orange]above: \textbf{Iterative Reasoning} $i<N$}] (iteration_box) {};
    \end{pgfonlayer}

\end{tikzpicture}

\rmfamily
\caption{Algorithm used for iterative knowledge extraction and graph refinement. 
At each iteration \( i \), the model generates reasoning tokens (blue). 
From the response, a local graph \( \mathcal{G}_{\text{local}}^i \) is extracted (violet) and merged with the global knowledge graph \( \mathcal{G} \) (light violet). 
The evolving graph is stored in multiple formats for visualization and analysis (yellow). 
Instead of letting the model respond to the task, a follow-up task is generated based on the latest extracted nodes and edges in \( \mathcal{G}_{\text{local}}^i \) (green), ensuring iterative refinement (orange), so that the model generates yet more reasoning tokens, and as part of that process, new nodes and edges. 
The process continues until the stopping condition \( i < N \) is met, yielding a final structured knowledge graph \( \mathcal{G} \) (orange).}
\label{fig:fig_01}
\end{figure}
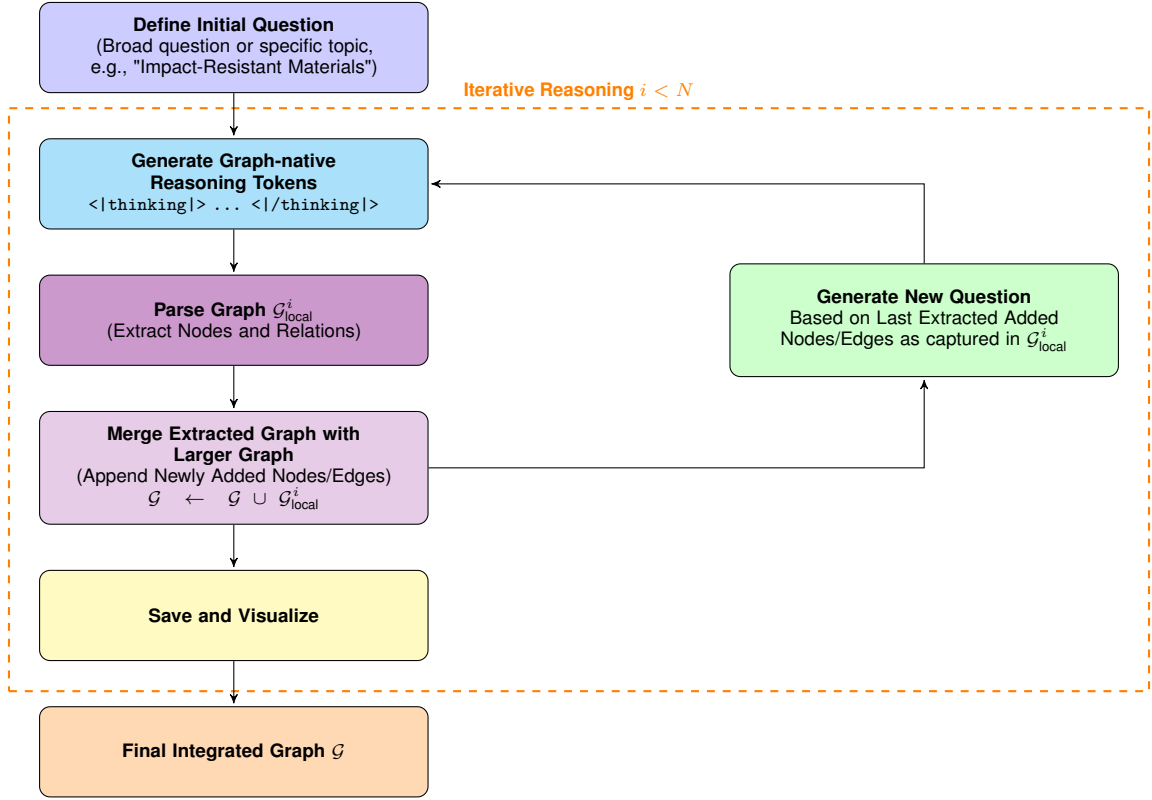


\subsection{Knowledge Graph Expansion Approaches}
Knowledge graphs are one way to organize relational understanding of the world. They have grown from manually curated ontologies decades ago into massive automatically constructed repositories of facts. A variety of methodologies have been developed for expanding knowledge graphs. Early approaches focused on information extraction from text using pattern-based or open-domain extractors. For example, the DIPRE algorithm \cite{brin1998dipre} bootstrapped relational patterns from a few seed examples to extract new facts in a self-reinforcing loop. Similarly, the KnowItAll system \cite{etzioni2004knowitall} introduced an open-ended, autonomous ``generate-and-test'' paradigm to extract entity facts from the web with minimal supervision. Open Information Extraction methods like TextRunner \cite{banko2007openIE} and ReVerb \cite{etzioni2011reverb} further enabled unsupervised extraction of subject–predicate–object triples from large text corpora without requiring a predefined schema. These unsupervised techniques expanded knowledge graphs by harvesting new entities and relations from unstructured data, although they often required subsequent mapping of raw extractions to a coherent ontology.

In parallel, research on knowledge graph completion has aimed to expand graphs by inferring missing links and attributes. Statistical relational learning and embedding-based models (e.g., translational embeddings like TransE \cite{bordes2013translating}) predict new relationships by generalizing from known graph structures. Such approaches, while not fully unsupervised (they rely on an existing core of facts for training), can autonomously suggest plausible new edges to add to a knowledge graph. Complementary to embeddings, logical rule-mining systems such as AMIE \cite{galarraga2013amie} showed that high-confidence Horn rules can be extracted from an existing knowledge base and applied to infer new facts recursively. Traditional link prediction heuristics from network science – for example, preferential attachment and other graph connectivity measures – have also been used as simple unsupervised methods to propose new connections in knowledge networks. Together, these techniques form a broad toolkit for knowledge graph expansion, combining text-derived new content with graph-internal inference to improve a graph’s coverage and completeness.

\subsection{Recursive and Autonomous Expansion Techniques}
\label{sec:rec_review}

A notable line of work seeks to make knowledge graphs growth continuous and self-sustaining – essentially achieving never-ending expansion. The NELL project (Never-Ending Language Learner) \cite{carlson2010nell} pioneered this paradigm, with a system that runs 24/7, iteratively extracting new beliefs from the web, integrating them into its knowledge base, and retraining itself to improve extraction competence each day. Over years of operation, NELL has autonomously accumulated millions of facts by coupling multiple learners (for parsing, classification, relation extraction, etc.) in a semi-supervised bootstrapping loop. This recursive approach uses the knowledge learned so far to guide future extractions, gradually expanding coverage while self-correcting errors; notably, NELL can even propose extensions to its ontology as new concepts emerge.

Another milestone in autonomous knowledge graph construction was Knowledge Vault~\cite{dong2014knowledgevault}, which demonstrated web-scale automatic knowledge base population by fusing facts from diverse extractors with probabilistic inference. Knowledge Vault combined extractions from text, tables, page structure, and human annotations with prior knowledge from existing knowledge graphs, yielding a vast collection of candidate facts (on the order of 300 million) each accompanied by a calibrated probability of correctness. This approach showed that an ensemble of extractors, coupled with statistical fusion, can populate a knowledge graph at scales far beyond what manual curation or single-source extraction can achieve. Both NELL and Knowledge Vault illustrate the power of autonomous or weakly-supervised systems that grow a knowledge graph with minimal human intervention, using recursive learning and data fusion to continually expand and refine the knowledge repository.

More recent research has explored agent-based and reinforcement learning (RL) frameworks for knowledge graph expansion and reasoning. Instead of one-shot predictions, these methods allow an agent to make multi-hop queries or sequential decisions to discover new facts or paths in the graph. For example, some work~\cite{xiong2017deeppath} employ an agent that learns to navigate a knowledge graph and find multi-step relational paths, effectively learning to reason over the graph to answer queries. Such techniques highlight the potential of autonomous reasoning agents that expand knowledge by exploring connections in a guided manner (using a reward signal for finding correct or novel information). This idea of exploratory graph expansion aligns with concepts in network science, where traversing a network can reveal undiscovered links or communities. It also foreshadowed approaches like Graph-PReFLexOR~\cite{buehler2025insitugraphreasoningknowledge} that treat reasoning as a sequential decision process, marked by special tokens, that can iteratively build and refine a task-specific knowledge graph.

Applications of these expansion techniques in science and engineering domains underscore their value for discovery~\cite{BuehlerGraphReasoning2024}. Automatically constructed knowledge graphs have been used to integrate and navigate scientific literature, enabling hypothesis generation by linking disparate findings. A classic example is Swanson’s manual discovery of a connection between dietary fish oil and Raynaud’s disease, which emerged by linking two disjoint bodies of literature through intermediate concepts \cite{swanson1986undiscovered,Cameron2013GraphSwanson}. Modern approaches attempt to replicate such cross-domain discovery in an automated way: for instance, mining biomedical literature to propose new drug–disease links, or building materials science knowledge graphs that connect material properties, processes, and applications to suggest novel materials, engineering concepts, or designs~\cite{nickel2016review,BuehlerGraphReasoning2024}.

\subsection{Relation to Earlier Work and Key Hypothesis}
\label{sec:rel_work}

The prior work discussed in Section~\ref{sec:rec_review} provides a foundation for our approach, which draws on the never-ending learning spirit of NELL~\cite{carlson2010nell} and the web-scale automation of Knowledge Vault~\cite{dong2014knowledgevault} to dynamically grow a knowledge graph \textit{in situ} as it reasons. Like those systems, it integrates information from diverse sources and uses iterative self-improvement. However, rather than relying on passive extraction or purely probabilistic link prediction, our method pairs on-the-fly logical reasoning with graph expansion within the construct of a graph-native reasoning LLM. This means each newly added node or edge is both informed by and used for the model’s next step of reasoning. Inspired in part by category theory and hierarchical inference, we move beyond static curation by introducing a principled, recursive reasoning loop that helps maintain transparency in how the knowledge graph evolves. In this sense, the work can be seen as a synthesis of existing ideas—continuous learning, flexible extraction, and structured reasoning—geared toward autonomous problem-solving in scientific domains.

Despite substantial progress in knowledge graph expansion, many existing methods still depend on predefined ontologies, extensive post-processing, or reinforce only a fixed set of relations. NELL and Knowledge Vault, for instance, demonstrated how large-scale extraction and integration of facts can be automated, but they rely on established schemas or require manual oversight to refine extracted knowledge~\cite{carlson2010nell,dong2014knowledgevault}. Reinforcement learning approaches such as DeepPath~\cite{xiong2017deeppath} can efficiently navigate existing graphs but do not grow them by generating new concepts or hypotheses.

By contrast, the work reported here treats reasoning as an active, recursive process that expands a knowledge graph while simultaneously refining its structure. This aligns with scientific and biological discovery processes, where knowledge is not just passively accumulated but also reorganized in light of new insights. Another key distinction is the integration of preference-based objectives, enabling more explicit interpretability of each expansion step. Methods like TransE~\cite{bordes2013translating} excel at capturing statistical regularities but lack an internal record of reasoning paths; our approach, in contrast, tracks and justifies each newly added node or relation. This design allows for a transparent, evolving representation that is readily applied to interdisciplinary exploration—such as in biomedicine~\cite{swanson1986undiscovered} and materials science~\cite{nickel2016review}—without depending on rigid taxonomies.

Hence, this work goes beyond conventional graph expansion by embedding recursive reasoning directly into the construction process, bridging the gap between passive knowledge extraction and active discovery. As we show in subsequent sections, this self-expanding paradigm yields scale-free knowledge graphs in which emergent hubs and bridge nodes enable continuous reorganization, allowing the system to evolve its understanding without exhaustive supervision and paving the way for scalable hypothesis generation and autonomous reasoning.

\paragraph{Hypothesis.} We hypothesize that recursive graph expansion enables self-organizing knowledge formation, allowing intelligence-like behavior to emerge without predefined ontologies, external supervision, or centralized control. Using a pre-trained model, Graph-PReFLexOR (an autonomous graph-reasoning model trained on a corpus of biological and biologically inspired materials principles) we demonstrate that knowledge graphs can continuously expand in a structured yet open-ended manner, forming scale-free networks with emergent conceptual hubs and interdisciplinary bridge nodes. Our findings suggest that intelligence-like reasoning can arise from recursive self-organization, challenging conventional paradigms and advancing possibilities for autonomous scientific discovery and scalable epistemic reasoning.


\section{Results and Discussion}

\begin{figure}[h!]
    \centering
    \includegraphics[width=1\textwidth]{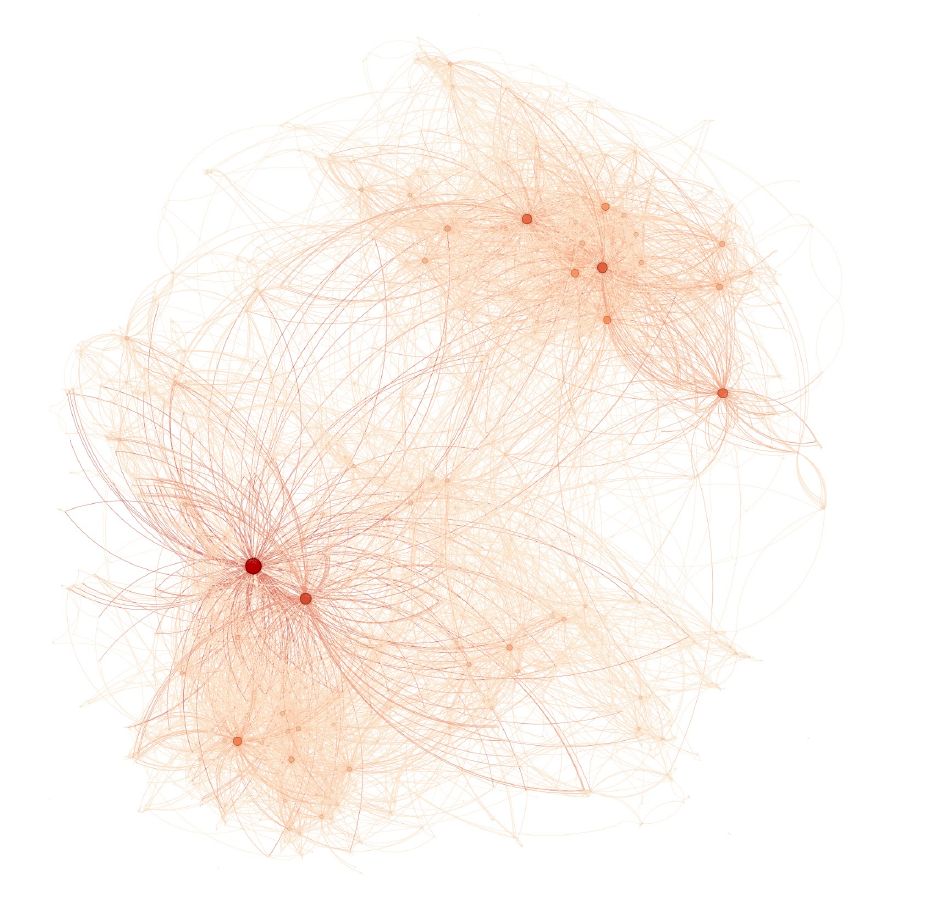}
    \caption{Knowledge graph  $\mathcal{G_1}$ after around 1,000 iterations, under a flexible self-exploration scheme initiated with the prompt \texttt{Discuss an interesting idea in bio-inspired materials science.} We observe the formation of a highly connected graph with multiple hubs and centers. }
    \label{fig:fig_2000}
\end{figure}

We present the results of experiments in which the graph-native reasoning model engages in a continuous, recursive process of graph-based reasoning, expanding its knowledge graph representation autonomously over 1,000 iterations. Unlike prior approaches that rely on a small number of just a few recursive reasoning steps, the experiments reported in this paper explore how knowledge formation unfolds in an open-ended manner, generating a dynamically evolving graph. As the system iterates, it formulates new tasks, refines reasoning pathways, and integrates emerging concepts, progressively structuring its own knowledge representation following the simple algorithmic paradigm delineated in Figure~\ref{fig:fig_01}. The resulting graphs from all iterations form a final integrated knowledge graph, which we analyze for structural and conceptual insights. Figure~\ref{fig:fig_2000} depicts the final state of the graph, referred to as graph $\mathcal{G}_1$, after the full reasoning process. 

\begin{figure}[h!]
    \centering
    \includegraphics[width=1\textwidth]{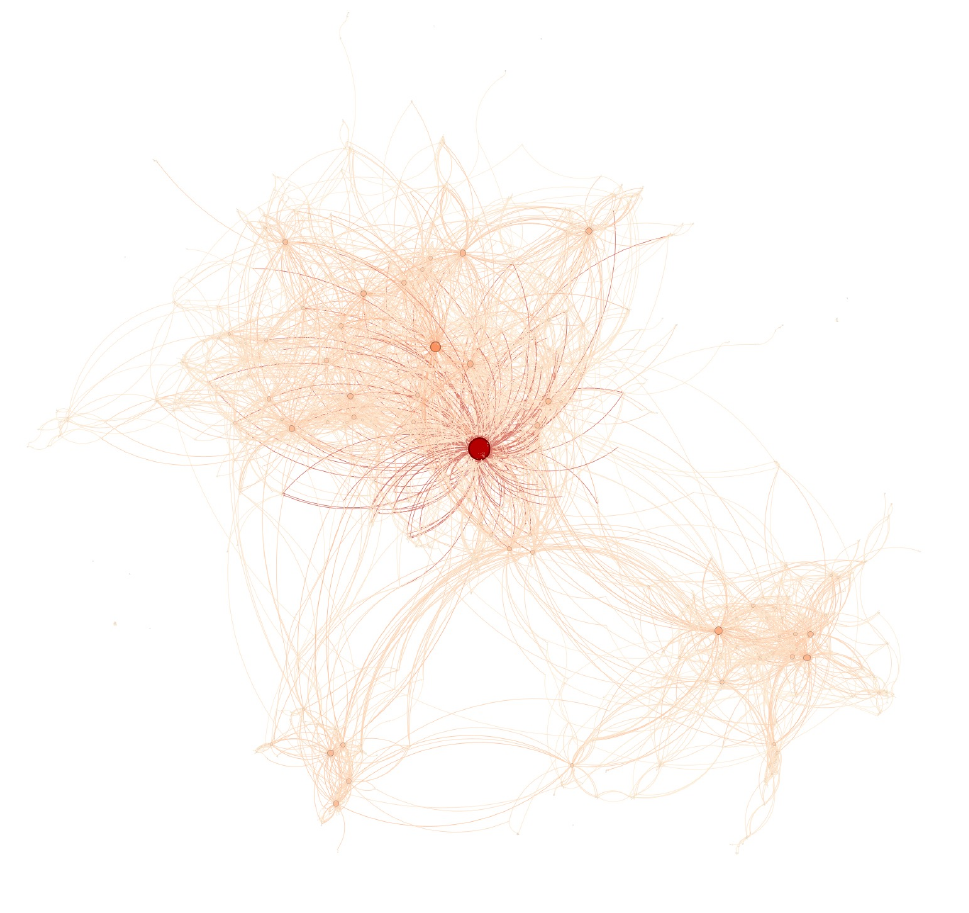}
    \caption{Visualizatrion of the knowledge graph \textbf{Graph 2} after around 500 iterations, under a topic-specific self-exploration scheme initiated with the prompt \texttt{Describe a way to design impact resistant materials.} The graph structure features a complex interwoven but highly connected network with multiple centers.}
    \label{fig:fig_2001}
\end{figure}

The recursive graph reasoning process can be conducted in either an open-ended setting or develoepd into a more tailored manner to address a specific domain or flavor in which reasoning steps are carried out (details, see Materials and Methods). In the example explored here, we focus on designing impact-resistant materials. In this specialized scenario, we initiate the model with a concise, topic-specific prompt -- e.g., \texttt{Describe a way to design impact resistant materials}, and maintain the iterative process of extracting structured knowledge from the model’s reasoning. We refer to the resulting graph as $\mathcal{G}_2$. Despite the narrower focus, the same core principles apply: each new piece of information from the language model is parsed into nodes and edges, appended to a global graph, and informs the next iteration’s query. In this way, $\mathcal{G}_2$ captures a highly directed and domain-specific knowledge space while still exhibiting many of the emergent structural traits—such as hub formation, stable modularity, and growing connectivity—previously seen in the more general graph $\mathcal{G}_1$. Figure~\ref{fig:fig_2001} shows the final snapshot for $\mathcal{G}_2$. To further examine the emergent structural organization of both graphs, Figures~\ref{fig:fig_2000_C} and \ref{fig:fig_2001_C} display the same graphs with nodes and edges colored according to cluster identification, revealing the conceptual groupings that emerge during recursive knowledge expansion. 
 
Table~\ref{tab:graph_comparison} shows a comparison of network properties for two graphs (graph $\mathcal{G_1}$, see Figure~\ref{fig:fig_2000} and graph $\mathcal{G_2}$, see Figure~\ref{fig:fig_2001}), each computed at the end of their iterations.  The scale-free nature of each graph is determined by fitting the degree distribution to a power-law model using the maximum likelihood estimation method. The analysis involves estimating the power-law exponent (\(\alpha\)) and the lower bound (\(x_{\min}\)), followed by a statistical comparison against an alternative exponential distribution. A log-likelihood ratio (LR) greater than zero and a $p$-value below 0.05 indicate that the power-law distribution better explains the degree distribution than an exponential fit, suggesting that the network exhibits scale-free behavior. In both graphs, these criteria are met, supporting a scale-free classification.
We observe that $\mathcal{G_1}$ has a power-law exponent of \(\alpha = 3.0055\), whereas $\mathcal{G_2}$ has a lower \(\alpha = 2.6455\), indicating that Graph 2 has a heavier-tailed degree distribution with a greater presence of high-degree nodes (hubs). The lower bound \(x_{\min}\) is smaller in $\mathcal{G_2}$  (\(x_{\min} = 10.0\)) compared to $\mathcal{G_1}$ (\(x_{\min} = 24.0\)), suggesting that the power-law regime starts at a lower degree value, reinforcing its stronger scale-free characteristics.

Other structural properties provide additional insights into the connectivity and organization of these graphs. The average clustering coefficients (0.1363 and 0.1434) indicate moderate levels of local connectivity, with $\mathcal{G_2}$ exhibiting slightly higher clustering. The average shortest path lengths (5.1596 and 4.8984) and diameters (17 and 13) suggest that both graphs maintain small-world characteristics, where any node can be reached within a relatively short number of steps. The modularity values (0.6970 and 0.6932) indicate strong community structures in both graphs, implying the presence of well-defined clusters of interconnected nodes. These findings collectively suggest that both graphs exhibit small-world and scale-free properties, with $\mathcal{G_2}$ demonstrating a stronger tendency towards scale-free behavior due to its lower exponent and smaller \(x_{\min}\).

Beyond scale-free characteristics, we note that the two graphs exhibit differences in structural properties that influence their connectivity and community organization. We find that $\mathcal{G_1}$, with 3,835 nodes and 11,910 edges, is much larger and more densely connected than $\mathcal{G_2}$, which has 2,180 nodes and 6,290 edges. However, both graphs have similar average degrees (6.2112 and 5.7706), suggesting comparable overall connectivity per node. The number of self-loops is slightly higher in Graph 1 (70 vs. 33), though this does not significantly impact global structure. The clustering coefficients (0.1363 and 0.1434) indicate moderate levels of local connectivity, with Graph 2 exhibiting slightly more pronounced local clustering. The small-world nature of both graphs is evident from their average shortest path lengths (5.1596 and 4.8984) and diameters (17 and 13), implying efficient information flow. Modularity values (0.6970 and 0.6932) suggest both graphs have well-defined community structures, with Graph 1 showing marginally stronger modularity, possibly due to its larger size. Overall, while both graphs display small-world and scale-free properties, $\mathcal{G_2}$ appears to have a more cohesive structure with shorter paths and higher clustering, whereas $\mathcal{G_1}$ is larger with a slightly stronger community division.

\begin{table}[h!]
\centering
\begin{tabular}{|l|p{3cm}|p{3cm}|}
\hline
\textbf{Metric} & \textbf{Graph} $\mathcal{G_1}$ & \textbf{Graph}  $\mathcal{G_2}$ \\ \hline
Number of nodes & 3835 & 2180 \\ \hline
Number of edges & 11910 & 6290 \\ \hline
Average degree & 6.2112 & 5.7706 \\ \hline
Number of self-loops & 70 & 33 \\ \hline
Average clustering coefficient & 0.1363 & 0.1434 \\ \hline
Average shortest path length (LCC) & 5.1596 & 4.8984 \\ \hline
Diameter (LCC) & 17 & 13 \\ \hline
Modularity (Louvain) & 0.6970 & 0.6932 \\ \hline
Log-likelihood ratio (LR) & 15.6952 & 39.6937 \\ \hline
p-value & 0.0250 & 0.0118 \\ \hline
Power-law exponent (\(\alpha\)) & 3.0055 & 2.6455 \\ \hline
Lower bound (\(x_{\min}\)) & 24.0 & 10.0 \\ \hline
Scale-free classification & Yes & Yes \\ \hline
\end{tabular}
\caption{Comparison of network properties for two graphs (graph $\mathcal{G_1}$, see Figure~\ref{fig:fig_2000} and \ref{fig:fig_2000_C} and graph $\mathcal{G_2}$, see Figure~\ref{fig:fig_2001} and \ref{fig:fig_2001_C}), each computed at the end of their iterations. Both graphs exhibit scale-free characteristics, as indicated by the statistically significant preference for a power-law degree distribution over an exponential fit (log-likelihood ratio \(LR > 0\) and \(p < 0.05\)). The power-law exponent (\(\alpha\)) for $\mathcal{G_1}$ is 3.0055, while $\mathcal{G_2}$ has a lower exponent of 2.6455, suggesting a heavier-tailed degree distribution. The clustering coefficients (0.1363 and 0.1434) indicate the presence of local connectivity, while the shortest path lengths (5.1596 and 4.8984) and diameters (17 and 13) suggest efficient global reachability. The high modularity values (0.6970 and 0.6932) indicate strong community structure in both graphs. Overall, both networks exhibit hallmark properties of scale-free networks, with $\mathcal{G_2}$ showing a more pronounced scale-free behavior due to its lower \(\alpha\) and lower \(x_{\min}\).}
\label{tab:graph_comparison}
\end{table}

\subsection{Basic Analysis of Recursive Graph Growth}
We now move on to a detailed analysis of the evolution of the graph as the reasoning process unfolds over thinking iterations. This sheds light into how the iterative process dynamically changes the nature of the graph. The analysis is largely focused on $\mathcal{G_1}$, albeit a few key results are also included for  $\mathcal{G_2}$. Detailed methods about how the various quantities are computed are included in Materials and Methods.

\begin{figure}[h!]
    \centering
    \includegraphics[width=0.9\textwidth]{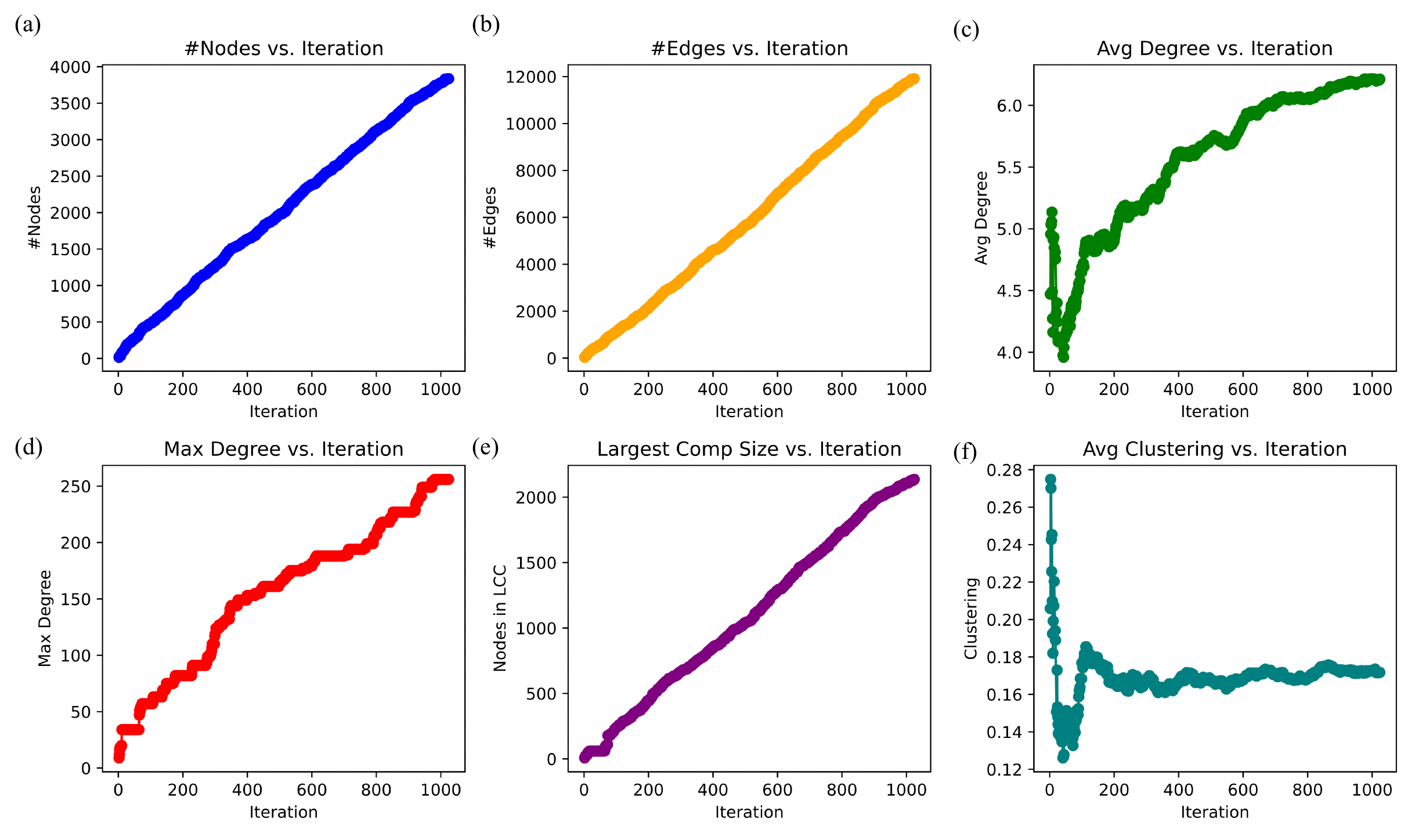}
    \caption{Evolution of basic graph properties over recursive iterations, highlighting the emergence of hierarchical structure, hub formation, and adaptive connectivity, for $\mathcal{G_1}$.}
    \label{fig:graph_analysis}
\end{figure}

Figure~\ref{fig:graph_analysis} illustrates the evolution of key structural properties of the recursively generated knowledge graph. The number of nodes and edges both exhibit linear growth with iterations, indicating that the reasoning process systematically expands the graph without saturation. The increase in edges is slightly steeper than that of nodes, suggesting that each new concept introduced is integrated into an increasingly dense network of relationships rather than remaining isolated. This continuous expansion supports the hypothesis that the model enables open-ended knowledge discovery through recursive self-organization.

The average degree of the graph steadily increases, stabilizing around six edges per node. This trend signifies that the knowledge graph maintains a balance between exploration and connectivity, ensuring that newly introduced concepts remain well-integrated within the broader structure. Simultaneously, the maximum degree follows a non-linear trajectory, demonstrating that certain nodes become significantly more connected over time. This emergent hub formation is characteristic of scale-free networks and aligns with patterns observed in human knowledge organization, where certain concepts act as central abstractions that facilitate higher-order reasoning.

The size of the largest connected component (LCC) grows proportionally with the total number of nodes, reinforcing the observation that the graph remains a unified, traversable structure rather than fragmenting into disconnected subgraphs. This property is crucial for recursive reasoning, as it ensures that the system retains coherence while expanding. The average clustering coefficient initially fluctuates but stabilizes around 0.16, indicating that while localized connections are formed, the graph does not devolve into tightly clustered sub-networks. Instead, it maintains a relatively open structure that enables adaptive reasoning pathways.

These findings highlight the self-organizing nature of the recursive reasoning process, wherein hierarchical knowledge formation emerges without the need for predefined ontologies or supervised corrections. The presence of conceptual hubs, increasing relational connectivity, and sustained network coherence suggest that the model autonomously structures knowledge in a manner that mirrors epistemic intelligence. This emergent organization enables the system to navigate complex knowledge spaces efficiently, reinforcing the premise that intelligence-like behavior can arise through recursive, feedback-driven information processing. Further analysis of degree distribution and centrality metrics would provide deeper insights into the exact nature of this evolving graph topology.

Figure~\ref{fig:graph_analysis_V7} illustrates the same analysis of the evolution of key structural properties of the recursively generated knowledge graph for graph $\mathcal{G_2}$, as a comparison.

\textbf{Structural Evolution of the Recursive Knowledge Graph}

\begin{figure}[h!]
    \centering
    \includegraphics[width=0.9\textwidth]{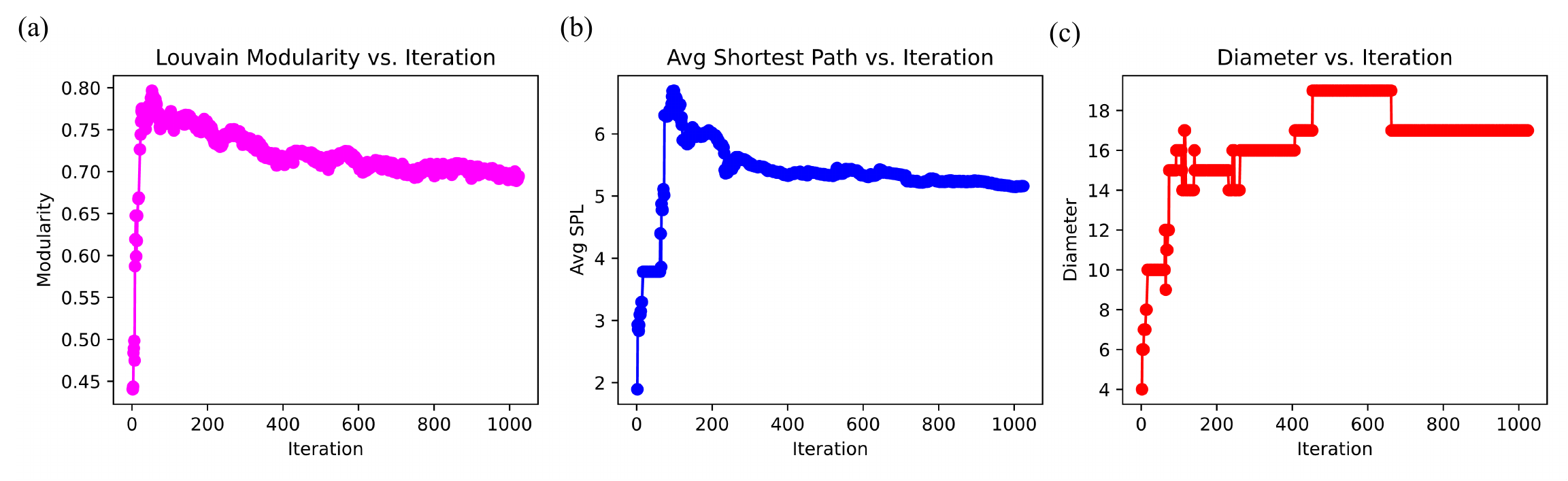}
    \caption{Evolution of key structural properties in the recursively generated knowledge graph $\mathcal{G_1}$: (a) Louvain modularity, showing stable community formation; (b) average shortest path length, highlighting efficient information propagation; and (c) graph diameter, demonstrating bounded hierarchical expansion.}
    \label{fig:graph_modularity}
\end{figure}

Figure~\ref{fig:graph_modularity} presents the evolution of three key structural properties, including Louvain modularity, average shortest path length, and graph diameter, over iterations. These metrics provide deeper insights into the self-organizing behavior of the graph as it expands through iterative reasoning. 
The Louvain modularity, depicted in Figure~\ref{fig:graph_modularity}(a), measures the strength of community structure within the graph. Initially, modularity increases sharply, reaching a peak around 0.75 within the first few iterations. This indicates that the early phases of reasoning lead to the rapid formation of well-defined conceptual clusters. As the graph expands, modularity stabilizes at approximately 0.70, suggesting that the system maintains distinct knowledge domains while allowing new interconnections to form. This behavior implies that the model preserves structural coherence, ensuring that the recursive expansion does not collapse existing conceptual groupings.

The evolution of the average shortest path length (SPL), shown in Figure~\ref{fig:graph_modularity}(b), provides further evidence of structured self-organization. Initially, the SPL increases sharply before stabilizing around 4.5–5.0. The initial rise reflects the introduction of new nodes that temporarily extend shortest paths before they are effectively integrated into the existing structure. The subsequent stabilization suggests that the recursive process maintains an efficient knowledge representation, ensuring that information remains accessible despite continuous expansion. This property is crucial for reasoning, as it implies that the system does not suffer from runaway growth in path lengths, preserving navigability.

The graph diameter, illustrated in Figure~\ref{fig:graph_modularity}(c), exhibits a stepwise increase, eventually stabilizing around 16–18. The staircase-like behavior suggests that the recursive expansion occurs in structured phases, where certain iterations introduce concepts that temporarily extend the longest shortest path before subsequent refinements integrate them more effectively. This bounded expansion indicates that the system autonomously regulates its hierarchical growth, maintaining a balance between depth and connectivity.

These findings reveal several emergent properties of the recursive reasoning model. The stabilization of modularity demonstrates the ability to autonomously maintain structured conceptual groupings, resembling human-like hierarchical knowledge formation. The controlled growth of the shortest path length highlights the system’s capacity for efficient information propagation, preventing fragmentation. We note that the bounded expansion of graph diameter suggests that reasoning-driven recursive self-organization is capable of structuring knowledge in a way that mirrors epistemic intelligence, reinforcing the hypothesis that certain forms of intelligent-like behavior can emerge without predefined ontologies.

For comparison, Figure~\ref{fig:graph_modularity_V7} presents the evolution of three key structural properties—Louvain modularity, average shortest path length, and graph diameter—over recursive iterations for graph $\mathcal{G_2}$.

\subsection{Analysis of Advanced Graph Evolution Metrics}

\begin{figure}[h!]
    \centering
    \includegraphics[width=0.9\textwidth]{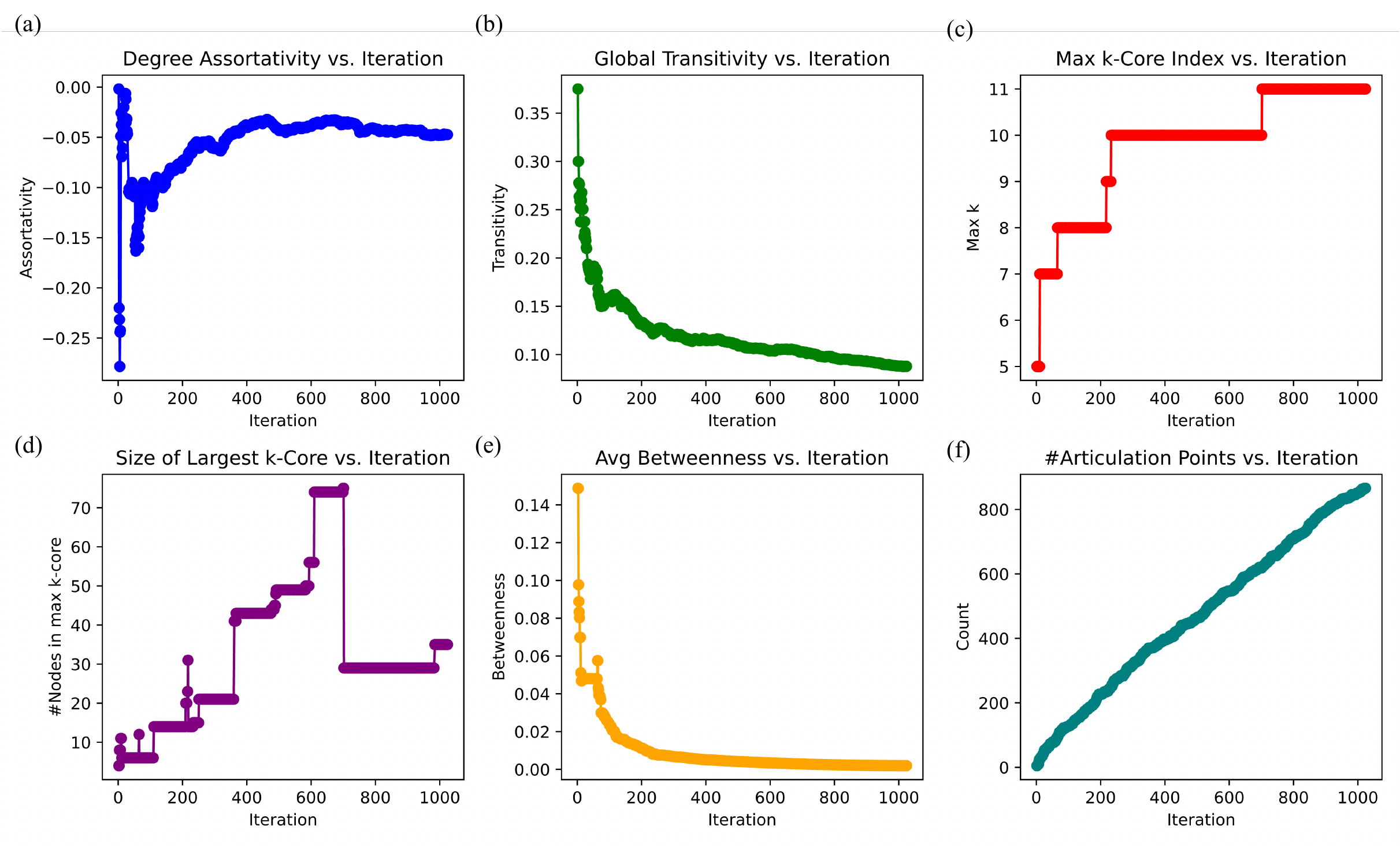}
    \caption{Evolution of advanced structural properties in the recursively generated knowledge graph $\mathcal{G_1}$: (a) degree assortativity, (b) global transitivity, (c) maximum k-core index, (d) size of the largest k-core, (e) average betweenness centrality, and (f) number of articulation points. These metrics reveal the emergence of hierarchical organization, hub formation, and increased navigability over recursive iterations.}
    \label{fig:advanced_graph_analysis}
\end{figure}

Figure~\ref{fig:advanced_graph_analysis} presents the evolution of six advanced structural metrics over recursive iterations, capturing higher-order properties of the self-expanding knowledge graph. These measures provide insights into network organization, resilience, and connectivity patterns emerging during recursive reasoning.

Degree assortativity coefficient is a measure of the tendency of nodes to connect to others with similar degrees. A negative value indicates disassortativity (high-degree nodes connect to low-degree nodes), while a positive value suggests assortativity (nodes prefer connections to similarly connected nodes). The degree assortativity coefficient (Figure~\ref{fig:advanced_graph_analysis}(a)) begins with a strongly negative value near \(-0.25\), indicating a disassortative structure where high-degree nodes preferentially connect to low-degree nodes. Over time, assortativity increases and stabilizes around \(-0.05\), suggesting a gradual shift toward a more balanced connectivity structure without fully transitioning to an assortative regime. This trend is consistent with the emergence of hub-like structures, characteristic of scale-free networks, where a few nodes accumulate a disproportionately high number of connections.

The global transitivity (Figure~\ref{fig:advanced_graph_analysis}(b)), measuring the fraction of closed triplets in the network, exhibits an initial peak near 0.35 before rapidly declining and stabilizing towards 0.10, albeit still decreasing. This suggests that early in the recursive reasoning process, the graph forms tightly clustered regions, likely due to localized conceptual groupings. As iterations progress, interconnections between distant parts of the graph increase, reducing local clustering and favoring long-range connectivity, a hallmark of expanding knowledge networks.

The $k$-core Index defines the largest integer $k$ for which a subgraph exists where all nodes have at least 
$k$ connections. A higher maximum $k$-core index suggests a more densely interconnected core. The maximum $k$-core index (Figure~\ref{fig:advanced_graph_analysis}(c)), representing the deepest level of connectivity, increases in discrete steps, reaching a maximum value of 11. This indicates that as the graph expands, an increasingly dense core emerges, reinforcing the formation of highly interconnected substructures. The stepwise progression suggests that specific iterations introduce structural reorganizations that significantly enhance connectivity rather than continuous incremental growth.

We observe that the size of the largest $k$-core (Figure~\ref{fig:advanced_graph_analysis}(d)) follows a similar pattern, growing in discrete steps and experiencing a sudden drop around iteration 700 before stabilizing again. This behavior suggests that the graph undergoes structural realignments, possibly due to the introduction of new reasoning pathways that temporarily reduce the dominance of the most connected core before further stabilization.

Betweenness Centrality is a measure of how often a node appears on the shortest paths between other nodes. High betweenness suggests a critical role in information flow, while a decrease indicates decentralization and redundancy in pathways. The average betweenness centrality (Figure~\ref{fig:advanced_graph_analysis}(e)) initially exhibits high values, indicating that early reasoning iterations rely heavily on specific nodes to mediate information flow. Over time, betweenness declines and stabilizes a bit below 0.01, suggesting that the graph becomes more navigable and distributed, reducing reliance on key bottleneck nodes over more iterations. This trend aligns with the emergence of redundant reasoning pathways, making the system more robust to localized disruptions.

Articulation points are nodes whose removal would increase the number of disconnected components in the graph, meaning they serve as key bridges between different knowledge clusters. The number of articulation points (Figure~\ref{fig:advanced_graph_analysis}(f)) steadily increases throughout iterations, reaching over 800. This suggests that as the knowledge graph expands, an increasing number of bridging nodes emerge, reflecting a hierarchical structure where key nodes maintain connectivity between distinct regions. Despite this increase, the network remains well connected, indicating that redundant pathways mitigate the risk of fragmentation.

A network where the degree distribution follows a power-law, meaning most nodes have few connections, but a small number (hubs) have many (supporting the notion of a scale-free network). Our findings provide evidence that the recursive graph reasoning process spontaneously organizes into a hierarchical, scale-free structure, balancing local clustering, global connectivity, and efficient navigability. The noted trends in assortativity, core connectivity, and betweenness centrality confirm that the system optimally structures its knowledge representation over iterations, reinforcing the hypothesis that self-organized reasoning processes naturally form efficient and resilient knowledge networks.

\subsection{Evolution of Newly Connected Pairs}

\begin{figure}[h!]
    \centering
    \includegraphics[width=0.4\textwidth]{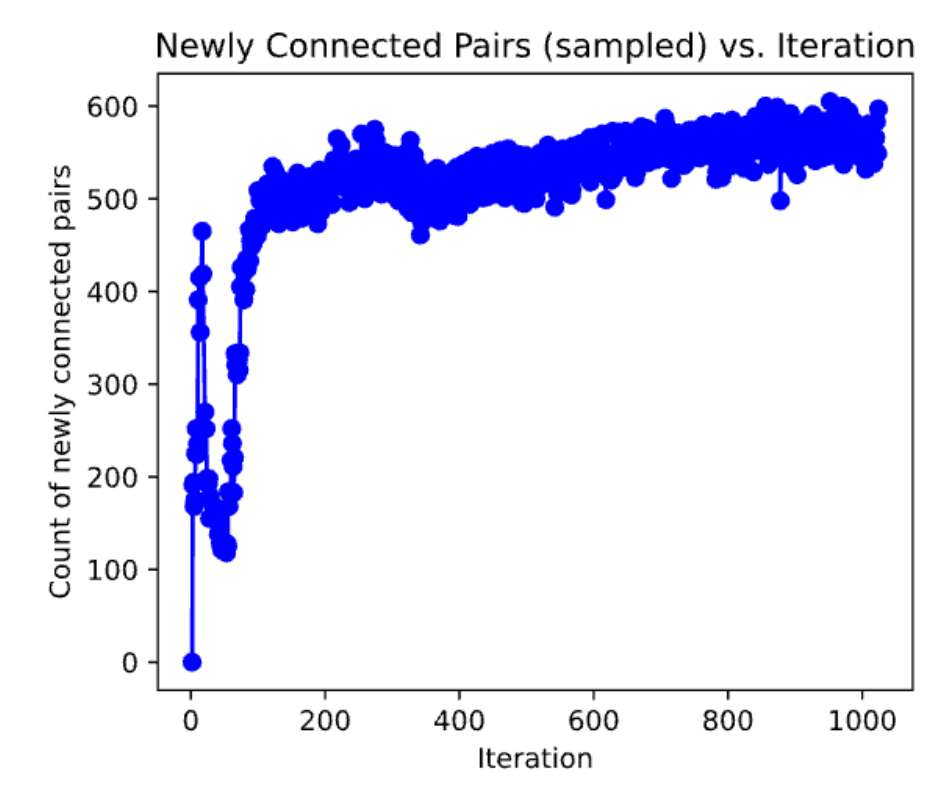}
    \caption{Evolution of newly connected node pairs over recursive iterations, $\mathcal{G_1}$. Early iterations exhibit high variability, reflecting an exploratory phase of rapid structural reorganization. Beyond 200 iterations, the process stabilizes, suggesting a steady-state expansion phase with sustained connectivity formation.}
    \label{fig:newly_connected_pairs}
\end{figure}

Figure~\ref{fig:newly_connected_pairs} presents the evolution of newly connected node pairs as a function of iteration, illustrating how the recursive reasoning process expands the knowledge graph over time. This metric captures the rate at which new relationships are established between nodes, providing insights into the self-organizing nature of the network.

In the early iterations (0–100), the number of newly connected pairs exhibits high variance, fluctuating between 0 and 400 connections per iteration. This suggests that the initial phase of recursive reasoning leads to significant structural reorganization, where large bursts of new edges are formed as the network establishes its fundamental connectivity patterns. The high variability in this region indicates an exploratory phase, where the graph undergoes rapid adjustments to define its core structure.

Beyond approximately 200 iterations, the number of newly connected pairs stabilizes around 500–600 per iteration, with only minor fluctuations. This plateau suggests that the knowledge graph has transitioned into a steady-state expansion phase, where new nodes and edges are integrated into an increasingly structured and predictable manner. Unlike random growth, this behavior indicates that the system follows a self-organized expansion process, reinforcing existing structures rather than disrupting them.

The stabilization at a high connection rate suggests the emergence of hierarchical organization, where newly introduced nodes preferentially attach to well-established structures. This pattern aligns with the scale-free properties observed in other experimentally acquired knowledge networks, where central concepts continuously accumulate new links, strengthening core reasoning pathways. The overall trend highlights how recursive self-organization leads to sustained, structured knowledge expansion, rather than arbitrary or saturation-driven growth.

The observed transition from high-variance, exploratory graph expansion to a stable, structured growth phase suggests that recursive self-organization follows a process similar to human cognitive learning and scientific discovery. We believe that this indicates that in early iterations, the system explores diverse reasoning pathways, mirroring how scientific fields establish foundational concepts through broad exploration before refining them into structured disciplines~\cite{kuhn1962structure}. The stabilization of connectivity beyond 200 iterations reflects preferential attachment dynamics, a hallmark of scale-free networks where highly connected nodes continue to accumulate new links, much like citation networks in academia~\cite{barabasi1999emergence}. This mechanism ensures that core concepts serve as attractors for further knowledge integration, reinforcing structured reasoning while maintaining adaptability. Importantly, the system does not exhibit saturation or stagnation, suggesting that open-ended knowledge discovery is possible through recursive reasoning alone, without requiring predefined ontologies or externally imposed constraints. This aligns with findings in AI-driven scientific hypothesis generation, where graph-based models dynamically infer new connections by iterating over expanding knowledge structures~\cite{swanson1986undiscovered, nickel2016review}. The ability of the system to self-organize, expand, and refine its knowledge base autonomously underscores its potential as a scalable framework for automated scientific discovery and epistemic reasoning.

\subsection{Analysis of Node Centrality Distributions at Final Stage of Reasoning}

\begin{figure}[h!]
    \centering
    \includegraphics[width=0.9\textwidth]{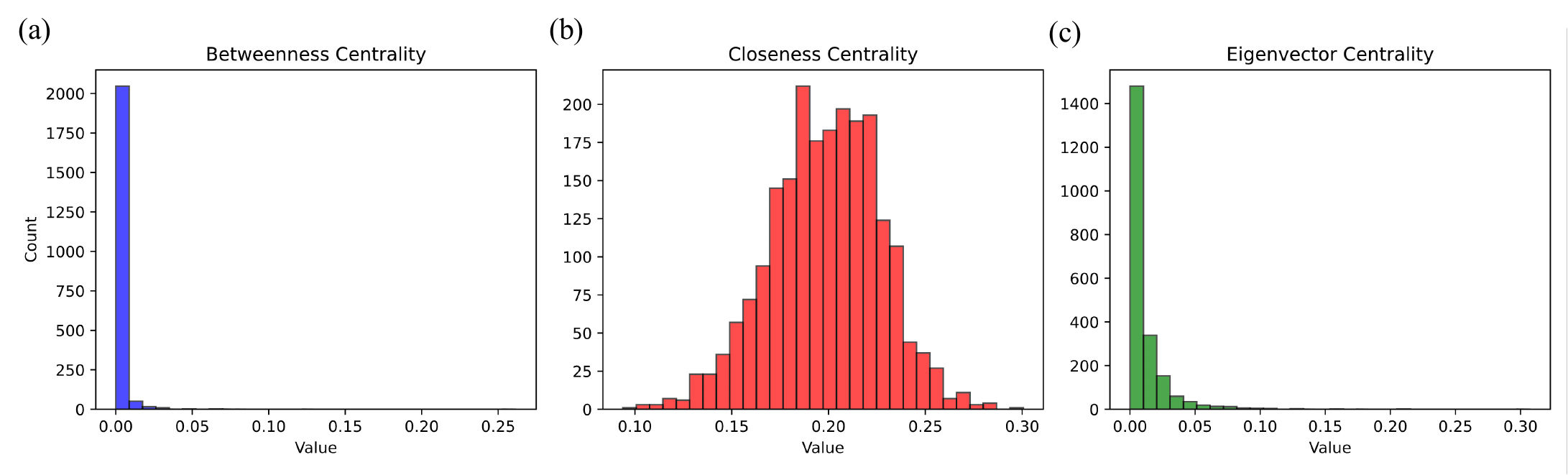}
    \caption{Distribution of node centrality measures in the recursively generated knowledge graph, for $\mathcal{G_1}$: (a) Betweenness centrality, showing that only a few nodes serve as major intermediaries; (b) Closeness centrality, indicating that the majority of nodes remain well-connected; (c) Eigenvector centrality, revealing the emergence of dominant hub nodes. These distributions highlight the hierarchical and scale-free nature of the evolving knowledge graph.}
    \label{fig:centrality_analysis}
\end{figure}

Next, Figure~\ref{fig:centrality_analysis} presents histograms for three key centrality measures—betweenness centrality, closeness centrality, and eigenvector centrality—computed for the recursively generated knowledge graph, at the final iteration. These metrics provide insights into the role of different nodes in maintaining connectivity, network efficiency, and global influence.

Figure~\ref{fig:centrality_analysis}(a) shows the distribution of betweenness centrality. We find the distribution of betweenness centrality to be highly skewed, with the majority of nodes exhibiting values close to zero. Only a small fraction of nodes attain significantly higher centrality values, indicating that very few nodes serve as critical intermediaries for shortest paths. This pattern is characteristic of hierarchical or scale-free networks, where a small number of hub nodes facilitate global connectivity, while most nodes remain peripheral. The presence of a few high-betweenness outliers suggests that key nodes emerge as crucial mediators of information flow, reinforcing the hypothesis that self-organizing structures lead to the formation of highly connected bridging nodes.

Figure~\ref{fig:centrality_analysis}(b) depicts the closeness centrality distribution. It follows an approximately normal distribution centered around 0.20, suggesting that most nodes remain well-connected within the network. This result implies that the network maintains a compact structure, allowing for efficient navigation between nodes despite continuous expansion. The relatively low spread indicates that the recursive reasoning process prevents excessive distance growth, ensuring that newly introduced nodes do not become isolated. This reinforces the observation that the graph remains navigable as it evolves, an essential property for maintaining coherent reasoning pathways.

Next, Figure~\ref{fig:centrality_analysis}(c) shows the eigenvector centrality distribution, identified to be also highly skewed, with most nodes having values close to zero. However, a few nodes attain substantially higher eigenvector centrality scores, indicating that only a select few nodes dominate the network in terms of global influence. This suggests that the network naturally organizes into a hierarchical structure, where dominant hubs accumulate influence over time, while the majority of nodes play a more peripheral role. The emergence of high-eigenvector hubs aligns with scale-free network behavior, further supporting the idea that reasoning-driven recursive self-organization leads to structured knowledge representation.

These findings indicate that the recursive knowledge graph balances global connectivity and local modularity, self-organizing into a structured yet efficient system. The few high-betweenness nodes act as key mediators, while the closeness centrality distribution suggests that the network remains efficiently connected. The eigenvector centrality pattern highlights the formation of dominant conceptual hubs, reinforcing the presence of hierarchical knowledge organization within the evolving reasoning framework.

\begin{figure}[h!]
    \centering
    \includegraphics[width=0.9\textwidth]{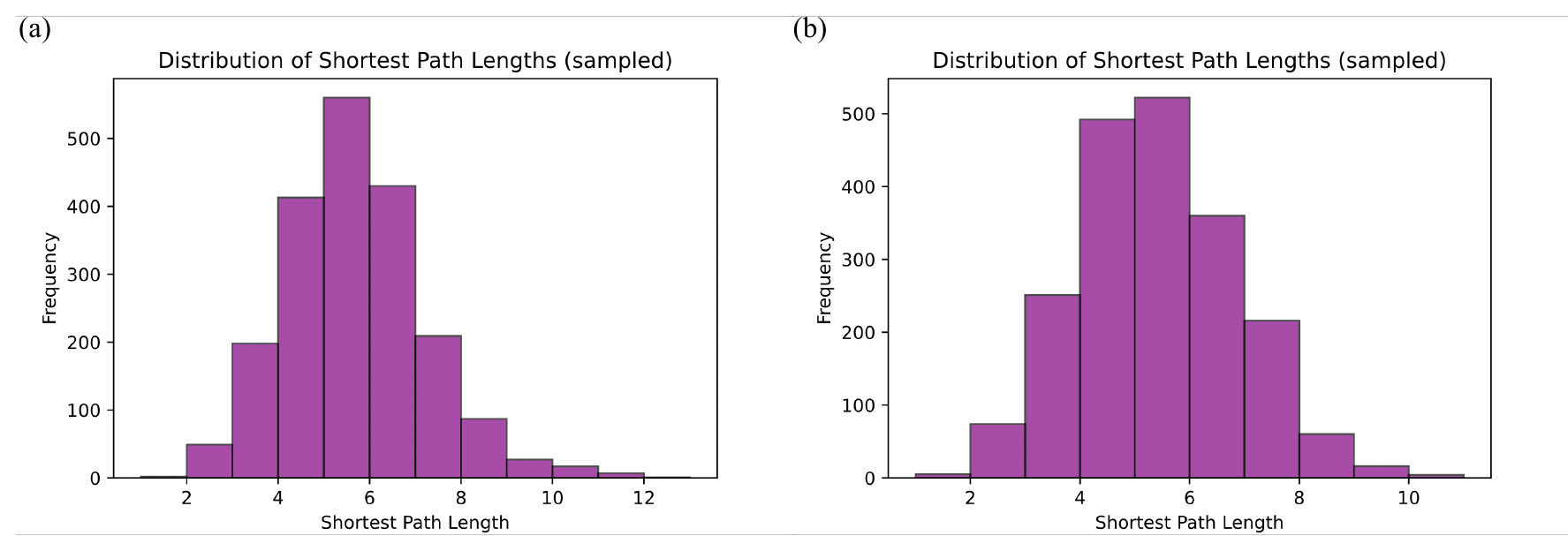}
    \caption{Distribution of sampled shortest path lengths in the recursively generated knowledge graphs (panel (a), for graph $\mathcal{G_2}$, panel (b), graph $\mathcal{G_2}$). The peak around 5–6 steps suggests that the network remains compact and navigable, while the slight right skew especially in panel (a) indicates the presence of peripheral nodes or specialized subdomains.}
    \label{fig:shortest_path_distribution}
\end{figure}

Figure~\ref{fig:shortest_path_distribution} presents the distribution of sampled shortest path lengths. This distribution provides insights into the overall compactness, navigability, and structural efficiency of the  network.

The histogram reveals that the most frequent shortest path length is centered around 5–6 steps, indicating that the majority of node pairs are relatively close in the network. The distribution follows a bell-shaped pattern, suggesting a typical range of distances between nodes, with a slight right skew where some paths extend beyond 10 steps. The presence of longer paths implies that certain nodes remain in the periphery or are indirectly connected to the core reasoning structure.

The relatively narrow range of shortest path lengths affirms that the network remains well-integrated, ensuring efficient knowledge propagation and retrieval. The absence of extreme outliers suggests that the recursive expansion process does not lead to fragmented or sparsely connected regions. This structure contrasts with purely random graphs, where shortest path lengths typically exhibit a narrower peak at lower values. The broader peak observed here suggests that the model does not generate arbitrary connections but instead organizes knowledge in a structured manner, balancing global integration with local modularity.

The observed path length distribution supports the hypothesis that recursive graph reasoning constructs an efficiently connected knowledge framework, where most concepts can be accessed within a small number of steps. The presence of some longer paths further suggests that the network exhibits hierarchical expansion, with certain areas developing as specialized subdomains that extend outward from the core structure.

\subsection{Knowledge Graph Evolution and Conceptual Breakthroughs}

The evolution of the knowledge graph over iterative expansions discussed so far reveals distinct patterns in knowledge accumulation, conceptual breakthroughs, and interdisciplinary integration. To analyze these processes, we now examine (i) the growth trajectories of major conceptual hubs, (ii) the emergence of new highly connected nodes, and (iii) overall network connectivity trends across iterations. The results of these analyses are presented in Figure~\ref{fig:knowledge_graph_evolution}, which consists of three sub-components.

\begin{figure}[h]
    \centering
     \includegraphics[width=.9\textwidth]{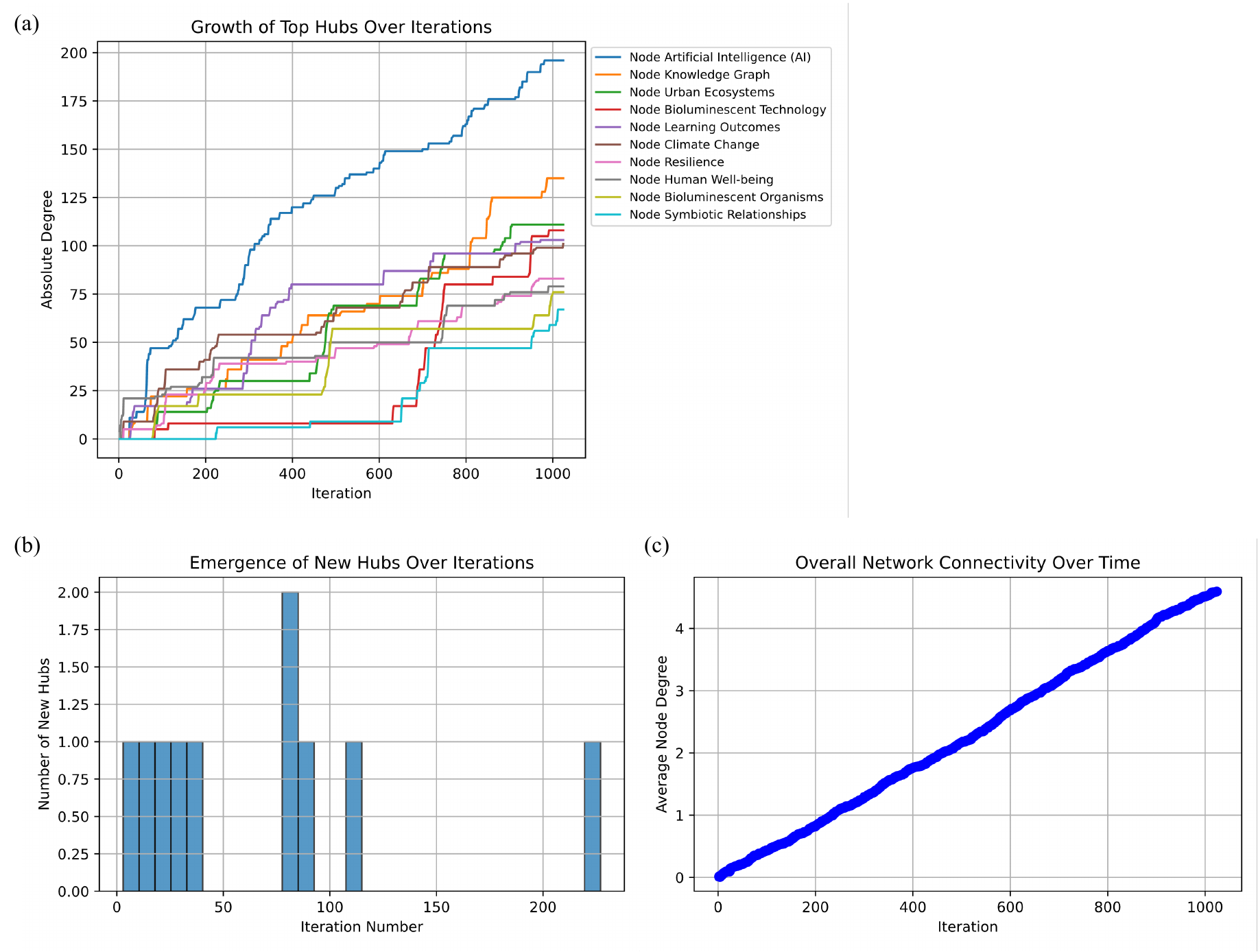}
    \caption{Evolution of knowledge graph structure across iterations, for $\mathcal{G_1}$. (a) Degree growth of the top conceptual hubs, showing both steady accumulation and sudden breakthroughs. (b) Histogram of newly emerging high-degree nodes across iterations, indicating phases of conceptual expansion. (c) Plot of the mean node degree over time, illustrating the system’s progressive integration of new knowledge.}
    \label{fig:knowledge_graph_evolution_100}
\end{figure}

The trajectory of hub development (Figure~\ref{fig:knowledge_graph_evolution_100}(a)) suggests two primary modes of knowledge accumulation: steady growth and conceptual breakthroughs. Certain concepts, such as \texttt{Artificial Intelligence (AI)} and \texttt{Knowledge Graphs}, exhibit continuous incremental expansion, reflecting their persistent relevance in structuring knowledge. In contrast, hubs like \texttt{Bioluminescent Technology} and \texttt{Urban Ecosystems} experience extended periods of low connectivity followed by sudden increases in node degree, suggesting moments when these concepts became structurally significant in the knowledge graph. These results indicate that the system does not expand knowledge in a purely linear fashion but undergoes phases of conceptual restructuring, akin to punctuated equilibrium in scientific development.

The emergence of new hubs (Figure~\ref{fig:knowledge_graph_evolution_100}(b)) further supports this interpretation. Instead of a continuous influx of new central concepts, we observe discrete bursts of hub formation occurring at specific iteration milestones. These bursts likely correspond to the accumulation of contextual knowledge reaching a critical threshold, after which the system autonomously generates new organizing principles to structure its expanding knowledge base. This finding suggests that the system’s reasoning process undergoes alternating cycles of consolidation and discovery, where previously formed knowledge stabilizes before new abstract concepts emerge.

The overall network connectivity trends (Figure~\ref{fig:knowledge_graph_evolution_100}(c)) demonstrate a steady increase in average node degree, indicating that the graph maintains a structurally stable expansion while integrating additional knowledge. The absence of abrupt drops in connectivity suggests that previously introduced concepts remain relevant and continue to influence reasoning rather than become obsolete. This trend supports the hypothesis that the system exhibits self-organizing knowledge structures, continuously refining its conceptual hierarchy as it expands.

These observations lead to several overarching conclusions. First, the results indicate that the system follows a hybrid knowledge expansion model, combining gradual accumulation with disruptive conceptual breakthroughs. This behavior closely mirrors the dynamics of human knowledge formation, where foundational ideas develop progressively, but major paradigm shifts occur when conceptual thresholds are crossed. Second, the persistence of high-degree hubs suggests that knowledge graphs generated in this manner do not suffer from catastrophic forgetting; instead, they maintain and reinforce previously established structures while integrating new insights. Third, the formation of new hubs in discrete bursts implies that knowledge expansion is not driven by uniform growth but by self-reinforcing epistemic structures, where accumulated reasoning reaches a tipping point that necessitates new abstract representations.

Additionally, the system demonstrates a structured directionality in knowledge formation, as evidenced by the smooth increase in average node degree without fragmentation. This suggests that new concepts do not disrupt existing structures but are incrementally woven into the broader network. Such behavior is characteristic of self-organizing knowledge systems, where conceptual evolution follows a dynamic yet cohesive trajectory. The model also exhibits potential for cross-domain knowledge synthesis, as indicated by the presence of nodes that transition into highly connected hubs later in the process. These nodes likely act as bridges between previously distinct knowledge clusters, fostering interdisciplinary connections.

These analyses provide strong evidence that the recursive graph expansion model is capable of simulating key characteristics of scientific knowledge formation. The presence of alternating stability and breakthrough phases, the hierarchical organization of concepts, and the increasing connectivity across knowledge domains all highlight the potential for autonomous reasoning systems to construct, refine, and reorganize knowledge representations dynamically. Future research could potentially focus on exploring the role of interdisciplinary bridge nodes, analyzing the hierarchical depth of reasoning paths, and examining whether the system can autonomously infer meta-theoretical insights from its evolving knowledge graph.

\subsection{Structural Evolution of the Knowledge Graph}

\begin{figure}[h]
    \centering
     \includegraphics[width=.9\textwidth]{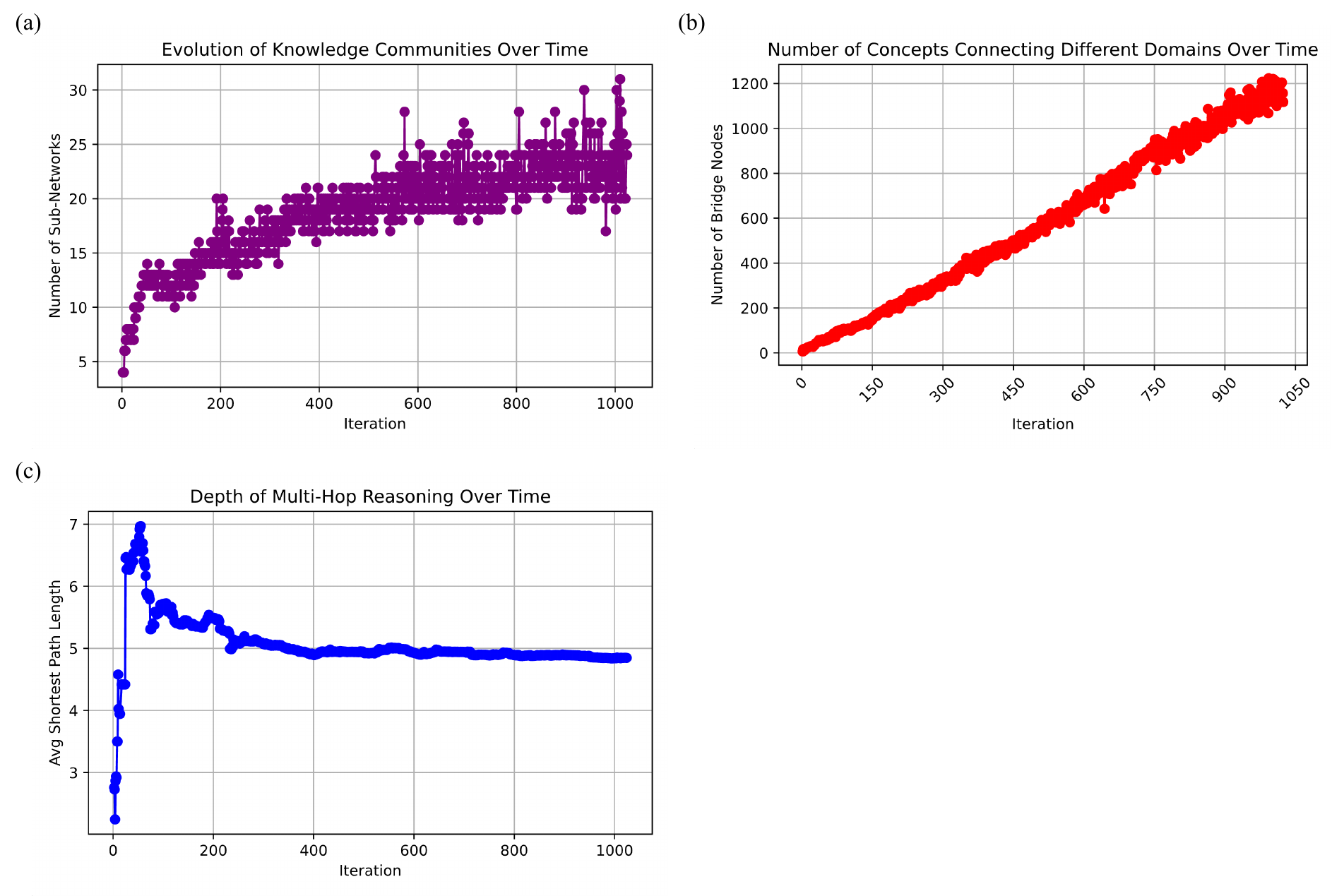}
    \caption{Structural evolution of the knowledge graph across iterations. (a) The number of distinct knowledge communities over time, showing an increasing trend with some fluctuations, for graph $\mathcal{G_1}$. (b) The growth of bridge nodes that connect multiple knowledge domains, following a steady linear increase. (c) The average shortest path length over iterations, indicating shifts in reasoning complexity as the graph expands.}
    \label{fig:knowledge_graph_evolution}
\end{figure}

The expansion of the knowledge graph over iterative refinements reveals emergent structural patterns that highlight how knowledge communities form, how interdisciplinary connections evolve, and how reasoning complexity changes over time. These dynamics provide insight into how autonomous knowledge expansion follows systematic self-organization rather than random accumulation. Figure~\ref{fig:knowledge_graph_evolution} presents three key trends: (a) the formation and growth of knowledge sub-networks, (b) the number of bridge nodes that connect different knowledge domains, and (c) the depth of multi-hop reasoning over iterations.

Figure~\ref{fig:knowledge_graph_evolution}(a) illustrates the formation of knowledge sub-networks over time. The number of distinct communities increases as iterations progress, reflecting the system's ability to differentiate between specialized fields of knowledge. The trend suggests two key observations: (i) an early rapid formation of new communities as novel knowledge domains emerge and (ii) a later stage where the number of communities stabilizes with occasional fluctuations. The latter behavior indicates that rather than indefinitely forming new disconnected knowledge clusters, the system reaches a regime where previously distinct domains remain relatively stable while undergoing minor structural reorganizations. The fluctuations in the later stages may correspond to moments where knowledge clusters merge or when new abstractions cause domain shifts.

Figure~\ref{fig:knowledge_graph_evolution}(b) tracks the number of bridge nodes (concepts that serve as interdisciplinary connectors) over iterative expansions. The steady, almost linear increase in bridge nodes suggests that as knowledge expands, more concepts naturally emerge as crucial links between different domains. This behavior reflects the self-reinforcing nature of knowledge integration, where new ideas not only expand within their respective fields but also introduce new ways to connect previously unrelated disciplines. Interestingly, there is no evidence of saturation in the number of bridge nodes, implying that the graph remains highly adaptive, continuously uncovering interdisciplinary relationships without premature convergence. This property is reminiscent of human knowledge structures, where interdisciplinary connections become more prevalent as scientific inquiry deepens.

Figure~\ref{fig:knowledge_graph_evolution}(c) examines the depth of multi-hop reasoning over iterations by measuring the average shortest path length in the graph. Initially, reasoning depth fluctuates significantly, which corresponds to the early phase of knowledge graph formation when structural organization is still emergent. As iterations progress, the average path length stabilizes, indicating that the system achieves a balance between hierarchical depth and accessibility of information. The early fluctuations may be attributed to the rapid reorganization of knowledge, where some paths temporarily become longer as new concepts emerge before stabilizing into more efficient reasoning structures. The eventual stabilization suggests that the graph reaches an equilibrium in how information propagates through interconnected domains, maintaining reasoning efficiency while still allowing for complex inferential pathways.

Taken together, these findings suggest that the autonomous knowledge expansion model exhibits structured self-organization, balancing specialization and integration. The interplay between distinct community formation, interdisciplinary connectivity, and reasoning depth highlights the emergence of a dynamically evolving but structurally coherent knowledge network. The continuous increase in bridge nodes reinforces the idea that interdisciplinary reasoning remains a central feature throughout the system’s expansion, which may have significant implications for autonomous discovery processes. Future analyses will explore whether certain bridge nodes exhibit long-term persistence as central knowledge connectors or if interdisciplinary pathways evolve dynamically based on newly introduced concepts.

\subsection{Persistence of Bridge Nodes in Knowledge Evolution}

\begin{figure}[h]
    \centering
     \includegraphics[width=0.7\textwidth]{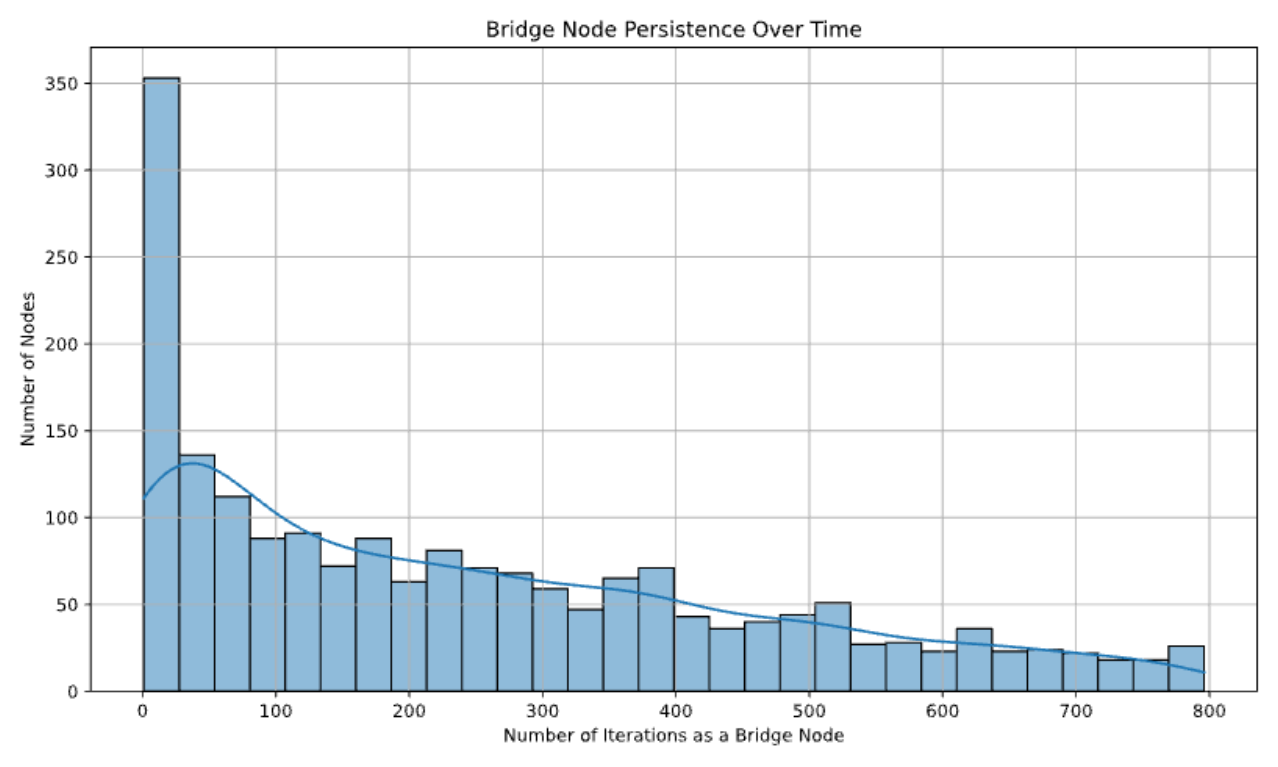}
    \caption{Histogram of bridge node persistence over iterations, for $\mathcal{G_1}$. The distribution follows a long-tail pattern, indicating that while most bridge nodes exist only briefly, a subset remains active across hundreds of iterations.}
    \label{fig:bridge_persistence}
\end{figure}

To understand the structural stability of interdisciplinary connections, we further analyze the persistence of bridge nodes—concepts that act as connectors between distinct knowledge domains, over multiple iterations. Figure~\ref{fig:bridge_persistence} presents a histogram of bridge node lifespans, showing how long each node remained an active bridge in the knowledge graph.

The distribution in Figure~\ref{fig:bridge_persistence} suggests that knowledge graph connectivity follows a hybrid model of structural evolution. The majority of bridge nodes appear only for a limited number of iterations, reinforcing the hypothesis that interdisciplinary pathways frequently evolve as new concepts emerge and replace older ones. This aligns with earlier observations that the knowledge system exhibits a high degree of conceptual dynamism.

However, a subset of bridge nodes remains persistent for hundreds of iterations. These nodes likely correspond to fundamental concepts that sustain long-term interdisciplinary connectivity. Their extended presence suggests that the system does not solely undergo continuous restructuring; rather, it maintains a set of core concepts that act as stable anchors in the evolving knowledge landscape.

These results refine our earlier observations by distinguishing between transient interdisciplinary connections and long-term structural stability. While knowledge graph expansion is dynamic, certain foundational concepts maintain their bridging role, structuring the broader knowledge network over extended periods. This hybrid model suggests that autonomous knowledge expansion does not operate under complete conceptual turnover but instead converges toward the emergence of stable, high-impact concepts that persist across iterations.

Related questions that could be explored in future research is whether these persistent bridge nodes correspond to widely used theoretical frameworks, methodological paradigms, or cross-domain knowledge principles. Additionally, further analysis is needed to examine whether long-term bridge nodes exhibit distinct topological properties, such as higher degree centrality or clustering coefficients, compared to short-lived connectors.

\subsection {Early Evolution of Bridge Nodes in Knowledge Expansion}

\begin{figure}[h]
    \centering
    \includegraphics[width=1.\textwidth]{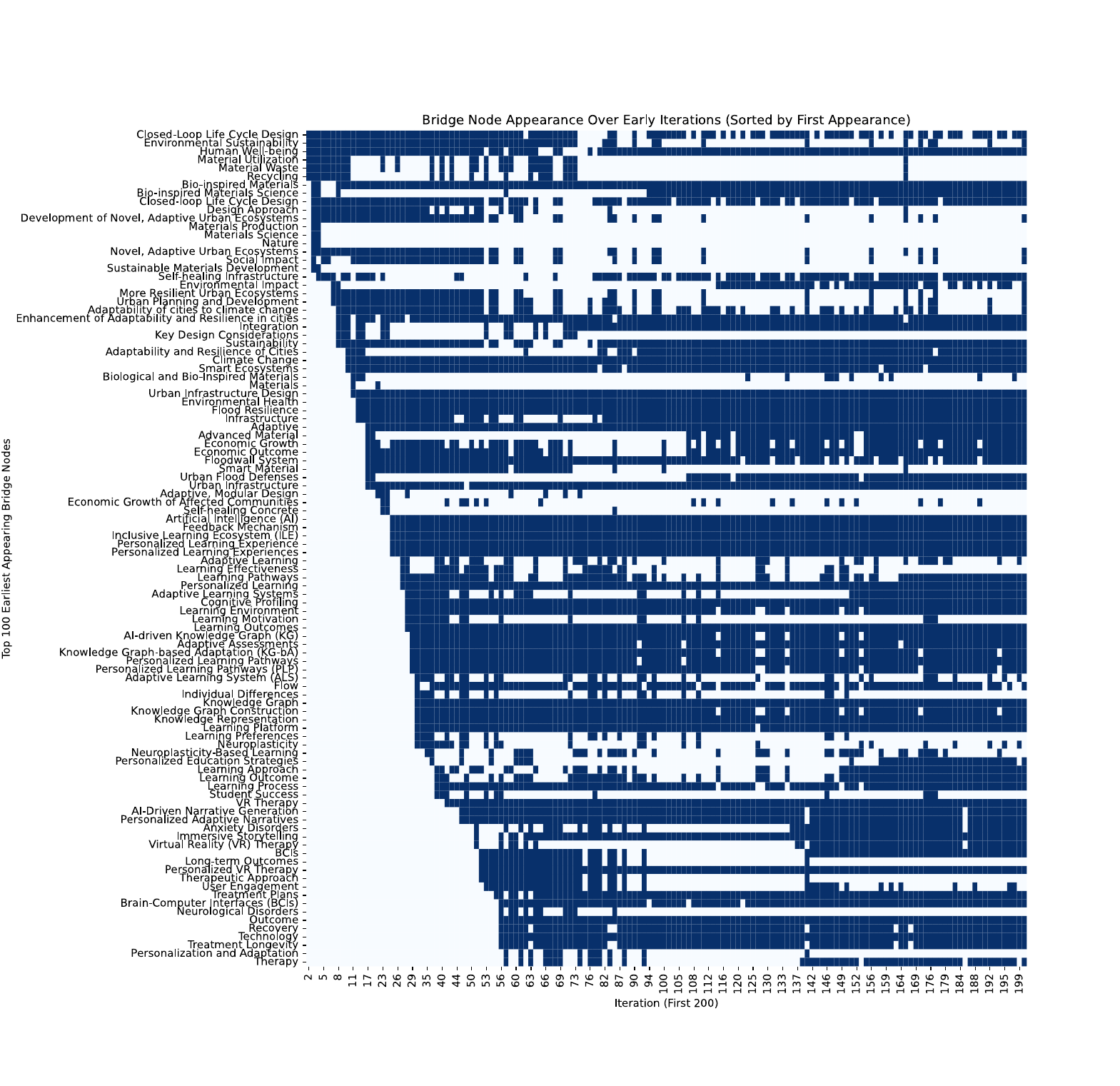}
    \caption{Emergence of bridge nodes over the first 200 iterations, sorted by first appearance, for $\mathcal{G_1}$. White regions indicate the absence of a node as a bridge, while dark blue regions denote its presence. Nodes that appear earlier in the graph evolution are positioned at the top. The structured emergence pattern suggests phases of knowledge expansion and stabilization.}
    \label{fig:sorted_bridge_heatmap}
\end{figure}

To examine the mechanics of the formation of interdisciplinary connections in the early stages of knowledge graph evolution, we pay close attention to the process. In the analysis discussed here, we identify the first occurrences of bridge nodes over the initial 200 iterations. Figure~\ref{fig:sorted_bridge_heatmap} presents a binary heatmap, where each row represents a bridge node, and each column corresponds to an iteration. The bridge nodes are sorted by the iteration in which they first appeared, providing a clearer view of how interdisciplinary connectors emerge over time.

The heatmap in Figure~\ref{fig:sorted_bridge_heatmap} reveals several key trends in the evolution of bridge nodes. Notably, the earliest iterations feature a rapid influx of bridge nodes, reflecting the initial structuring phase of the knowledge graph. Many nodes appear and remain active for extended periods, suggesting that certain concepts establish themselves as core interdisciplinary connectors early in the process. These nodes likely play a foundational role in structuring knowledge integration across domains.

A second notable pattern is the episodic emergence of new bridge nodes, rather than a continuous accumulation. The visualization shows distinct clusters of newly appearing bridge nodes, interspersed with periods of relative stability. These bursts suggest that knowledge integration occurs in structured phases rather than through gradual accumulation. Such phases may represent moments when the system reaches a threshold where newly integrated concepts allow for the creation of previously infeasible interdisciplinary links.

In contrast to the early-established bridge nodes, a subset of nodes appears only in later iterations. These late-emerging bridge nodes indicate that interdisciplinary roles are notably not static; rather, the system continuously restructures itself, incorporating new ideas as they gain relevance. This supports the hypothesis that certain bridge nodes emerge not from initial structuring but from later stages of conceptual refinement, possibly as higher-order abstractions connecting previously developed knowledge clusters.

The distribution of bridge node activity also suggests a mix of persistent and transient connectors. While some nodes appear briefly and disappear, others remain active over long stretches. This behavior reinforces the idea that knowledge expansion is both dynamic and structured, balancing exploration (where new connections are tested) and stabilization (where key interdisciplinary links persist).

We note that the structured emergence of bridge nodes may indicate that interdisciplinary pathways do not form randomly but are shaped by systematic phases of knowledge integration and refinement. Future analyses could explore the long-term impact of early bridge nodes, assessing whether they remain influential throughout the knowledge graph’s evolution, and whether the structure of interdisciplinary connectivity stabilizes or continues to reorganize over extended iterations.

\subsection{Evolution of Key Bridge Nodes Over Iterations}

\begin{figure}[h]
    \centering
    \includegraphics[width=0.9\textwidth]{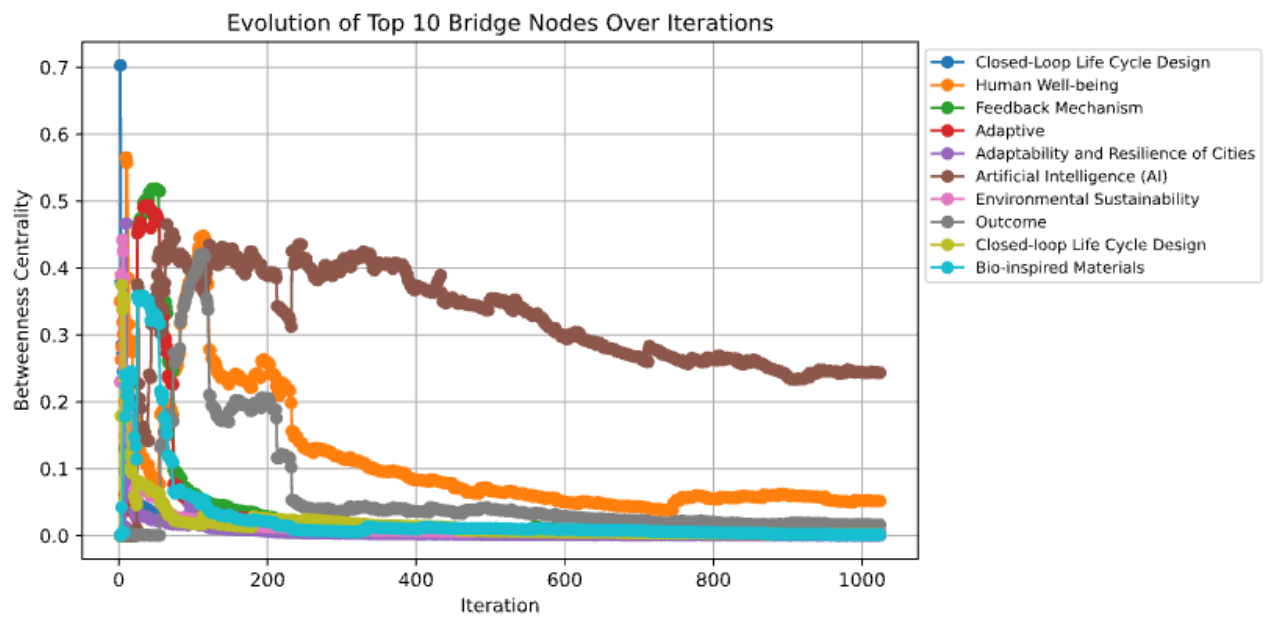}
    \caption{Evolution of the top 10 bridge nodes over iterations, for $\mathcal{G_1}$. Each curve represents the betweenness centrality of a bridge node, indicating its role in facilitating knowledge integration. Nodes that initially had high centrality later declined, while some concepts maintained their influence throughout the graph’s evolution.}
    \label{fig:bridge_node_evolution}
\end{figure}

To investigate how interdisciplinary pathways evolve in the knowledge graph, we analyzed the betweenness centrality of the most influential bridge nodes across 1,000 iterations. Figure~\ref{fig:bridge_node_evolution} presents the trajectory of the top 10 bridge nodes, highlighting their shifting roles in facilitating interdisciplinary connections.

The trends in Figure~\ref{fig:bridge_node_evolution} reveal distinct patterns in how bridge nodes emerge, peak in influence, and decline over time. Notably, nodes such as \texttt{Closed-Loop Life Cycle Design} and \texttt{Human Well-being} exhibit high betweenness centrality in the early iterations, suggesting that they played a fundamental role in structuring the initial interdisciplinary landscape. However, as the knowledge graph expanded, these nodes saw a gradual decline in their centrality, indicating that their role as primary connectors was replaced by alternative pathways.

A second class of bridge nodes, including \texttt{Adaptability and Resilience of Cities} and \texttt{Artificial Intelligence (AI)}, maintained high centrality values for a longer duration, suggesting that certain concepts remain essential to interdisciplinary knowledge integration even as the graph evolves. These nodes acted as long-term knowledge stabilizers, facilitating interactions between different research domains throughout a significant portion of the knowledge expansion process.

Interestingly, a subset of nodes, such as \texttt{Feedback Mechanism} and \texttt{Outcome}, gradually gained importance over time. Unlike early bridge nodes that peaked and declined, these nodes started with lower centrality but increased in influence in later iterations. This suggests that some interdisciplinary pathways only become critical after sufficient knowledge accumulation, reinforcing the idea that interdisciplinary roles are not static but continuously reorganize as the knowledge graph matures.

Furthermore, we observe that by approximately iteration 400-600, most bridge nodes' betweenness centrality values begin converging toward lower values, indicating that knowledge transfer is no longer reliant on a small set of nodes. This suggests that, as the graph expands, alternative pathways develop, leading to a more distributed and decentralized knowledge structure where connectivity is no longer dominated by a few highly influential nodes.

These findings support the hypothesis that interdisciplinary pathways evolve dynamically, with early-stage knowledge formation relying on a few key concepts, followed by a transition to a more robust and distributed network where multiple redundant pathways exist. Future analyses will focus on:
\begin{itemize}
    \item Identifying which nodes replaced early bridge nodes as major interdisciplinary connectors in later iterations.
    \item Comparing early vs. late-stage bridge nodes to assess whether earlier nodes tend to be general concepts, while later bridge nodes represent more specialized interdisciplinary knowledge.
    \item Analyzing the resilience of the knowledge graph by simulating the removal of early bridge nodes to determine their structural significance.
\end{itemize}

These results provide a perspective on how interdisciplinary linkages emerge, stabilize, and reorganize over time, offering insights into the self-organizing properties of large-scale knowledge systems.

\subsection{Evolution of Betweenness Centrality Distribution}

To analyze the structural evolution of the knowledge graph, we next examine the distribution of betweenness centrality at different iterations. Betweenness centrality is a measure of a node’s importance in facilitating knowledge transfer between different parts of the network. Formally, the betweenness centrality of a node $v$ is given by:

\begin{equation}
    C_B(v) = \sum_{s \neq v \neq t} \frac{\sigma_{st}(v)}{\sigma_{st}},
\end{equation}

where $\sigma_{st}$ is the total number of shortest paths between nodes $s$ and $t$, and $\sigma_{st}(v)$ is the number of those paths that pass through $v$. A higher betweenness centrality indicates that a node serves as a critical intermediary in connecting disparate knowledge domains.

Figure~\ref{fig:betweenness_distribution} presents histograms of betweenness centrality distribution at four key iterations (2, 100, 510, and 1024), illustrating the shifting role of bridge nodes over time.

Initially, at Iteration 2, the network is highly centralized, with a small number of nodes exhibiting extremely high betweenness centrality (above 0.6), while the majority of nodes have near-zero values. This indicates that only a few nodes act as critical interdisciplinary connectors, facilitating nearly all knowledge transfer.

By Iteration 100, the distribution has broadened, meaning that more nodes participate in knowledge transfer. The highest betweenness values have decreased compared to Iteration 2, and more nodes exhibit low but nonzero centrality, suggesting an increase in redundant pathways and reduced dependency on a few dominant bridge nodes.

At Iteration 510, the distribution becomes more skewed again, with fewer nodes having high betweenness centrality and a stronger concentration at low values. This suggests that the network has undergone a phase of structural consolidation, where interdisciplinary pathways reorganize around fewer, more stable bridges.

Finally, at Iteration 1024, the histogram shows that most nodes have low betweenness centrality, and only a few retain moderate values. This suggests that the network has matured into a more distributed structure, where no single node dominates knowledge transfer. The observed trend indicates that as the knowledge graph expands, the burden of interdisciplinary connectivity is increasingly shared among many nodes rather than concentrated in a few.

These results suggest that the system undergoes a dynamic reorganization process, shifting from an initial hub-dominated structure to a more distributed and resilient network. Future work could potentially explore whether these trends continue as the graph scales further and whether the eventual network state remains stable or undergoes additional restructuring.

\begin{figure}[h]
    \centering
    \includegraphics[width=.7\textwidth]{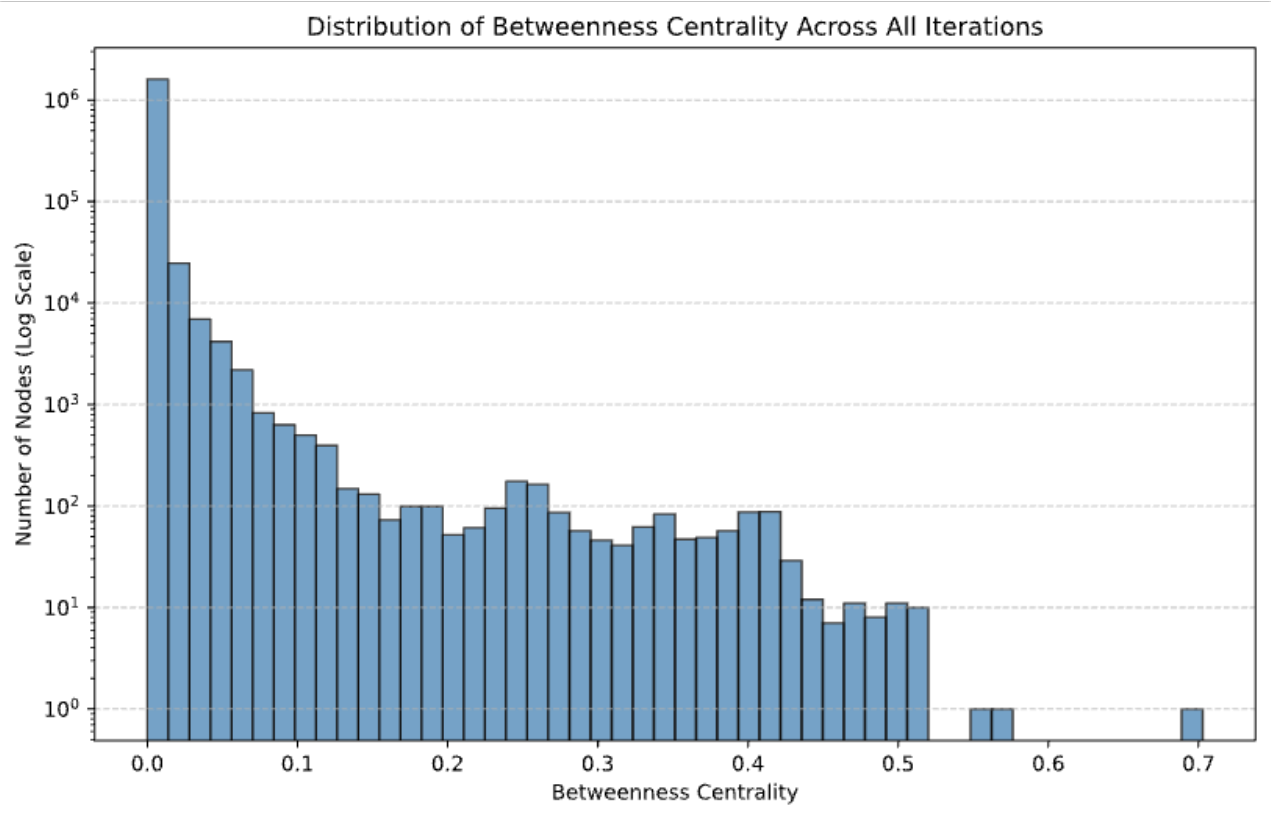}
    \caption{Distribution of betweenness centrality across all iterations, $\mathcal{G_1}$. The y-axis is log-scaled, showing the frequency of nodes with different centrality values. A small number of nodes dominate knowledge transfer, while most nodes exhibit near-zero centrality.}
    \label{fig:betweenness_distribution_across_all}
\end{figure}

To examine the overall structural properties of the knowledge graph, we analyzed the distribution of betweenness centrality across all iterations. Figure~\ref{fig:betweenness_distribution_across_all} presents a histogram of betweenness centrality values collected from all iterations of the knowledge graph. The distribution was generated by computing betweenness centrality for each iteration and aggregating all node values overall iterations.

The histogram in Figure~\ref{fig:betweenness_distribution_across_all} reveals a highly skewed distribution, where the majority of nodes exhibit near-zero betweenness centrality, while a small subset maintains significantly higher values. This pattern suggests that knowledge transfer within the network is primarily governed by a few dominant bridge nodes, which facilitate interdisciplinary connections. The presence of a long tail in the distribution indicates that these high-betweenness nodes persist throughout multiple iterations.

Interestingly, the distribution also exhibits multiple peaks, suggesting that the network consists of different classes of bridge nodes. Some nodes act as long-term stable interdisciplinary connectors, while others emerge as transient bridges that facilitate knowledge transfer only for limited iterations.

The log scale on the $y$-axis reveals that while most nodes contribute little to betweenness centrality, a significant number of nodes still exhibit low but nonzero values indicating that knowledge flow is distributed across many minor pathways. Over multiple iterations, it is expected that betweenness centrality values redistribute, reducing dependency on early dominant nodes and leading to a more decentralized knowledge structure.

These findings highlight that the knowledge graph maintains a core-periphery structure, where a few key nodes play a disproportionate role in bridging knowledge across disciplines. Future work will explore how the distribution evolves over time, identifying whether the network transitions toward a more evenly distributed structure or remains reliant on a small number of high-centrality nodes.

\subsection{Evolution of Betweenness Centrality in the Knowledge Graph}

\begin{figure}[h]
    \centering
    \includegraphics[width=.9\textwidth]{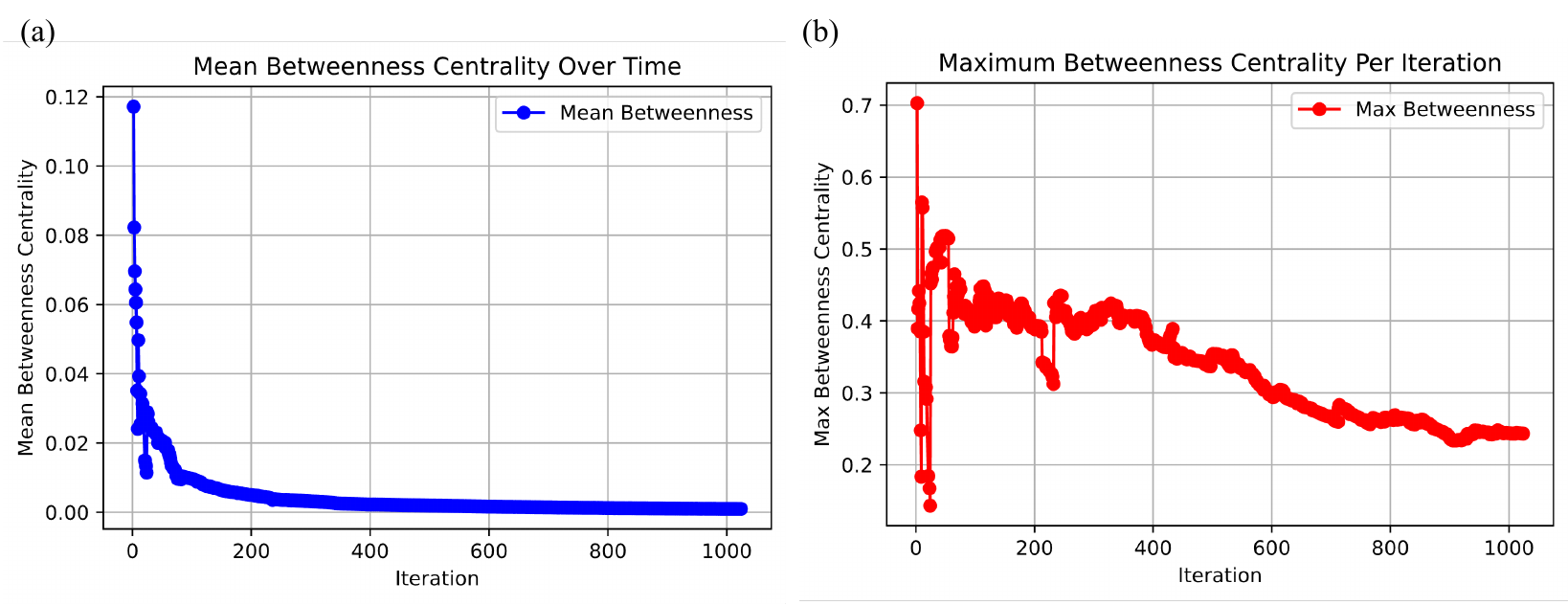} 
    \caption{Evolution of betweenness centrality in the knowledge graph, $\mathcal{G_1}$. Panel (a): Mean betweenness centrality over time, showing a transition from early high centralization to a more distributed state. Panel (b):  Maximum betweenness centrality per iteration, highlighting how the most dominant bridge nodes shift and decline in influence.}
    \label{fig:betweenness_evolution}
\end{figure}

To analyze the structural evolution of the knowledge graph, we tracked the changes in betweenness centrality over 1,000 iterations. Betweenness centrality quantifies the extent to which a node serves as a bridge between other nodes by appearing on shortest paths. A node with high betweenness centrality facilitates interdisciplinary knowledge transfer by linking otherwise disconnected regions of the network. Figures~\ref{fig:betweenness_evolution}(a) and \ref{fig:betweenness_evolution}(b) illustrate how mean and maximum betweenness centrality evolve over time. The first plot captures the average importance of nodes in knowledge transfer, while the second identifies the most dominant bridge nodes at each iteration.

Figure~\ref{fig:betweenness_evolution}(a) tracks the mean betweenness centrality, providing insight into how the overall distribution of knowledge transfer roles evolves. In the earliest iterations, the mean betweenness is extremely high, indicating that only a few nodes dominate knowledge exchange. However, as the graph expands and alternative pathways form, the mean betweenness declines rapidly within the first 100 iterations.

Between iterations 100 and 500, we observe a continued decline, but at a slower rate. This suggests that knowledge transfer is being shared across more nodes, reducing reliance on a small set of dominant bridges. After iteration 500, the values stabilize near zero, indicating that the network has reached a decentralized state, where multiple nodes contribute to knowledge integration instead of a few key intermediaries.

These trends suggest a self-organizing process, where the knowledge graph transitions from a highly centralized system into a more distributed and resilient network. The final structure is more robust, with many small bridges collectively supporting interdisciplinary connectivity instead of a few dominant hubs.

Figure~\ref{fig:betweenness_evolution}(b) examines the highest betweenness centrality recorded in each iteration, tracking the most dominant knowledge bridge at each stage. In the earliest iterations, a single node reaches an extreme betweenness value of around 0.7, indicating that knowledge transfer is highly bottlenecked through one or very few key nodes.

Between iterations 50 and 300, the maximum betweenness remains high, fluctuating between 0.3 and 0.5. This suggests that while the network becomes less dependent on a single node, a small number of highly central nodes still dominate knowledge flow. This phase represents a transition period, where the network starts distributing knowledge transfer across multiple nodes.

After iteration 500, the maximum betweenness exhibits a gradual decline, eventually stabilizing around 0.2. This suggests that the network has successfully decentralized, and knowledge transfer is no longer dominated by a single key node. The presence of multiple lower-betweenness bridge nodes implies that redundant pathways have developed, making the system more resilient to disruptions. This is in general agreement with earlier observations. 

The combined results from Figures~\ref{fig:betweenness_evolution}(a) and \ref{fig:betweenness_evolution}(b) suggest that the knowledge graph undergoes a fundamental structural transformation over time:
\begin{itemize}
    \item Initially, a few dominant nodes control knowledge flow, leading to high mean and maximum betweenness centrality.
    \item As the graph expands, new pathways emerge, and betweenness is distributed across more nodes.
    \item By the later iterations, no single node dominates, and knowledge transfer occurs through a decentralized structure.
\end{itemize}

This evolution suggests that the knowledge graph self-organizes into a more distributed state, where interdisciplinary connectivity is no longer constrained by a few central hubs. Future studies can explore whether this trend continues at larger scales and analyze which specific nodes maintained high betweenness longest and which replaced them in later iterations.

\subsection{Analysis of longest shortest path in $\mathcal{G}_2$ and analysis using agentic reasoning}

\begin{figure}[h]
    \centering
    \includegraphics[width=1.\textwidth]{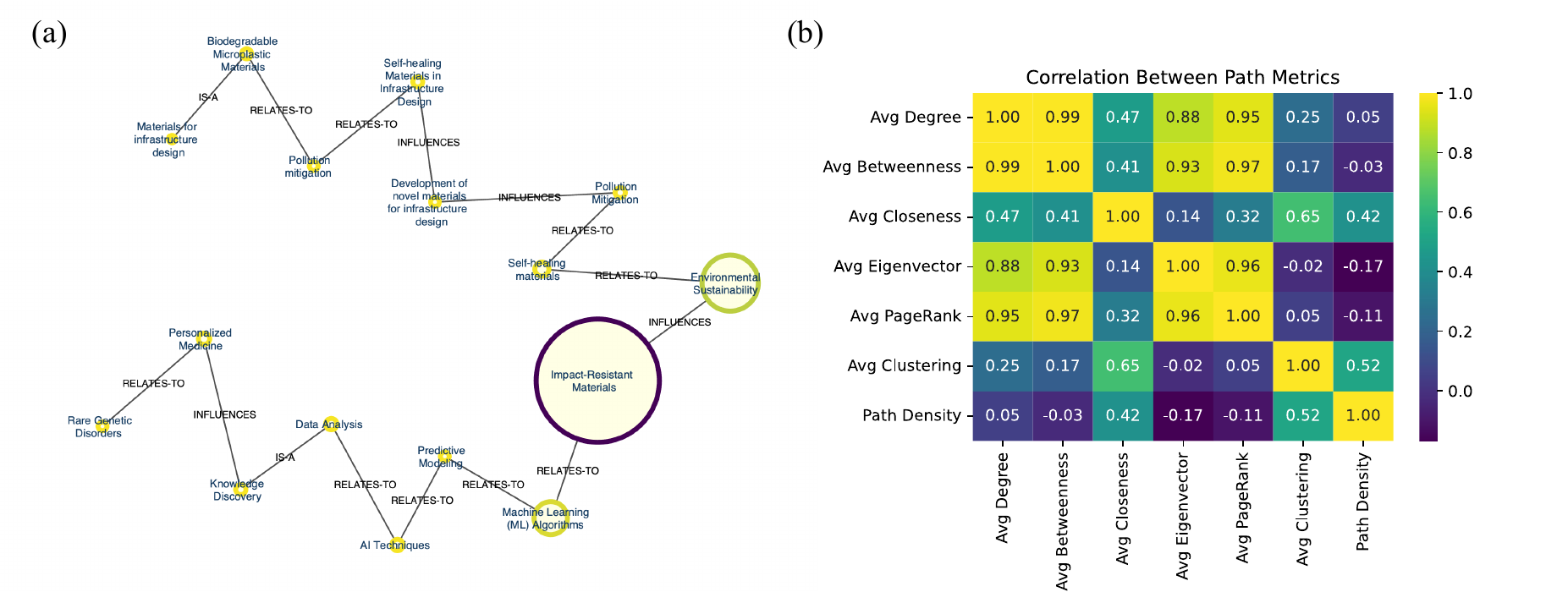}
    \caption{Longest shortest path analysis. Panel (a): Visualization of the longest shortest path (diameter path) in $\mathcal{G}_2$, presenting a fascinating chain of interdisciplinary relationships across medicine, data science and AI, materials science, sustainability, and infrastructure. Panel (b): Correlation heatmap of path-level metrics, computed for the first 30 longest shortest paths. Degree and betweenness centrality are highly correlated, indicating that high-degree nodes frequently serve as key connectors. Eigenvector centrality and PageRank also show strong correlation, highlighting their shared role in capturing node influence. Path density exhibits a weak or negative correlation with centrality measures, suggesting that highly connected nodes often form less dense structures. The metrics were computed for each path by extracting node-level properties (degree, betweenness, closeness, eigenvector centrality, PageRank, clustering coefficient) from the original graph and averaging them over all nodes in the path. Path density was calculated as the ratio of actual edges to possible edges within the path subgraph. Correlations were then derived from these aggregated values across multiple paths.
     }
    \label{fig:shortest_longest_G2}
\end{figure}

While the primary focus of this study is targeting a detailed analysis of graph dynamic experiments during reasoning, we also explore how graph reasoning based on the \textit{in-situ} generated graph can be used to improve responses through in-context learning~\cite{Brown2020LanguageLearners} (here, we use \url{meta-llama/Llama-3.2-3B-Instruct}). The methodology employs a graph-based reasoning framework to enhance LLM responses through structured knowledge extraction obtained through the method described above.  
Figure~\ref{fig:shortest_longest_G2}(b) depicts additional analysis, showing a correlation heatmap of path-level metrics, computed for the first 30 longest shortest paths. 

The extracted longest shortest path depicted in Figure~\ref{fig:shortest_longest_G2}(a) presents a compelling sequence of relationships spanning biotechnology, artificial intelligence, materials science, and sustainability, illustrating how advancements in one domain influence others. The overall logical flow is well-structured, with clear and expected progressions, such as \texttt{Rare Genetic Disorders} leading to \texttt{Personalized Medicine} and \texttt{Knowledge Discovery}, reflecting that the model captures the increasing role of AI in medical research. The sequence from \texttt{AI Techniques} to \texttt{Predictive Modeling} and \texttt{Machine Learning (ML) Algorithms} is similarly intuitive, as computational models underpin predictive simulations across disciplines (details on methods, see Section~\ref{methods:agentic_longest_shortest}).

However, some unexpected connections emerge, suggesting areas for further exploration. The link from \texttt{Machine Learning (ML) Algorithms} to \texttt{Impact-Resistant Materials} stands out -- not as a weak connection, but as an intriguing suggestion of AI-driven materials design rather than mere discovery. Computational techniques, such as reinforcement learning and generative modeling, could optimize material structures for durability, opening new pathways in materials engineering. Another unconventional relationship is the transition from \texttt{Biodegradable Microplastic Materials} to \texttt{Infrastructure Design}. These two areas typically operate separately, yet this link may hint at the emergence of biodegradable composites for construction or sustainable materials engineering. Further investigation into the practical applications of biodegradable materials in structural design could strengthen this connection.

A notable redundancy appears in the presence of \texttt{Pollution Mitigation} twice, spelled differently, which results from a lack of node merging rather than a distinct conceptual relationship. This duplication suggests that similar concepts are being represented as separate nodes, potentially affecting graph-based reasoning. Similarly, \texttt{Self-Healing Materials in Infrastructure Design} loops back to \texttt{Pollution Mitigation}, reinforcing an already established sustainability link. While valid, this repetition could be streamlined for clarity.

We find that the logical progression effectively captures key interdisciplinary relationships while revealing areas for refinement. The structure underscores the increasing role of AI in materials science, the integration of sustainability into materials design, and the interplay between predictive modeling and physical sciences. Addressing node duplication and refining transitions between traditionally separate fields—such as biodegradable materials in construction—would enhance the clarity and coherence of the path, making it an even more insightful representation of scientific knowledge.

\paragraph{Agentic Reasoning over the Path}  We apply an agentic model to analyze the longest shortest path. For this analysis, an agentic system first analyzes each node in the subgraph, then each of the relationships, and then synthesizes them into a ``Final Synthesized Discovery'' (in blue font for clarity). The analysis identifies key concepts such as biodegradable microplastics, self-healing materials, pollution mitigation, and AI-driven predictive modeling, ultimately synthesizing the \texttt{Bio-Inspired, Adaptive Materials for Resilient Ecosystems (BAMES)} paradigm. The resulting document, \hyperlink{supporting_text_1}{Supporting Text 1}, presents the results.

The proposed discovery proposes self-healing, bio-inspired materials that integrate microbial, plant, and animal-derived mechanisms with AI-driven optimization to create adaptive, environmentally responsive materials. By embedding microorganisms for pollutant degradation and leveraging machine learning for real-time optimization, the model suggests that \texttt{BAMES} extends conventional self-healing materials beyond infrastructure applications into active environmental remediation~\cite{white2001autonomic}. The concept of temporal memory, where materials learn from past environmental conditions and adjust accordingly, introduces a novel paradigm in smart materials~\cite{complex_systems_review}. Additionally, the hypothesis that interconnected materials could develop emergent, collective behavior akin to biological ecosystems presents an interesting perspective on material intelligence and sustainability~\cite{bioinspired_design,Nepal2022HierarchicallyNanocomposites}.

\begin{figure}[h]
    \centering
    \includegraphics[width=.8\textwidth]{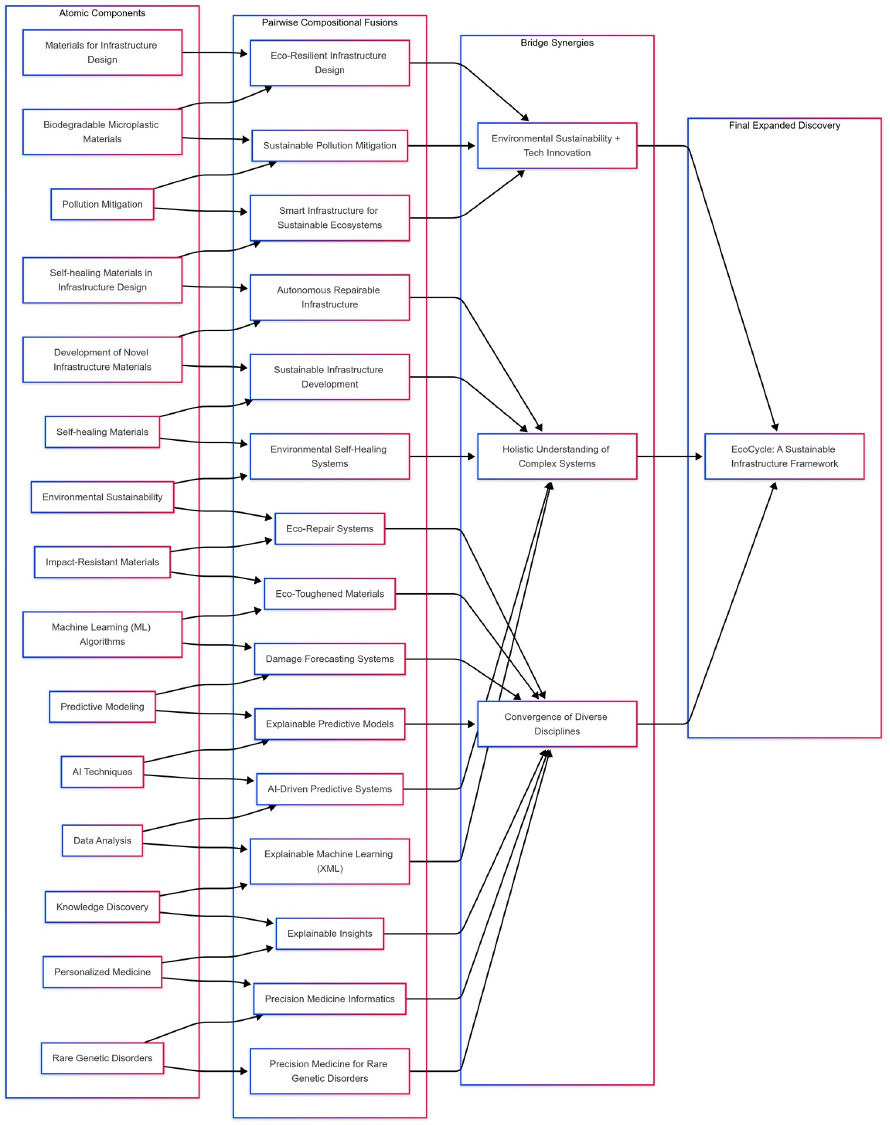}
    \caption{Compositional framework applied to the longest shortest path. The flowchart illustrates the hierarchical process of compositional reasoning, beginning with atomic components (fundamental scientific concepts, left, as identified in the longest shortest path (Figure~\ref{fig:shortest_longest_G2}(a))) and progressing through pairwise fusions, bridge synergies, and a final expanded discovery. Each stage (Steps A, B, C and D) integrates concepts systematically, ensuring interoperability, generativity, and hierarchical refinement, culminating in the \texttt{EcoCycle} framework for sustainable infrastructure development.     }
    \label{fig:compositional_analysis}
\end{figure}

\paragraph{Agentic Compositional Reasoning} We can formalize this approach further and induce agentic strategy to develop compositional reasoning (see, Section~\ref{methods:agentic_compositional} for details). In this experiment, implement a systematic development of hierarchical reasoning over concepts, pairs of concepts, and so on. The resulting document is shown in \hyperlink{supporting_text_2}{Supporting Text 2}, and Figure~\ref{fig:compositional_analysis} shows a flowchart of the reasoning process. 

The example ultimately presents a structured approach to compositional scientific discovery, integrating principles from infrastructure materials science, environmental sustainability, and artificial intelligence to develop a novel framework for sustainable infrastructure, termed \texttt{EcoCycle}. As can be seen in \hyperlink{supporting_text_2}{Supporting Text 2} and in Figure~\ref{fig:compositional_analysis}, the compositional reasoning process proceeded through multiple hierarchical steps, ensuring the systematic combination of concepts with well-defined relationships. 

At the foundational level, atomic components were identified, each representing an independent domain concept, such as biodegradable microplastic materials, self-healing materials, predictive modeling, and knowledge discovery. These fundamental elements were then combined into pairwise fusions, leveraging shared properties to generate novel synergies. For instance, the fusion of self-healing materials with pollution mitigation led to \texttt{environmental self-healing systems}, integrating autonomous repair mechanisms with pollution reduction strategies. Similarly, combining impact-resistant materials with machine learning algorithms enabled \texttt{damage forecasting systems}, enhancing predictive maintenance in infrastructure.

The validity of this compositional reasoning was established by ensuring that each fusion preserved the integrity of its constituent concepts while generating emergent functionalities. The process adhered to key compositionality principles: (1) \texttt{Interoperability}, ensuring that combined components interacted meaningfully rather than arbitrarily; (2) \texttt{Generativity}, whereby new properties emerged that were not present in the individual components; and (3) \texttt{Hierarchical Refinement}, wherein smaller-scale synergies were recursively integrated into higher-order bridge synergies. This led to overarching themes such as \texttt{the intersection of environmental sustainability and technological innovation} and \texttt{the holistic understanding of complex systems}, demonstrating the robustness of the approach.

Ultimately, these synergies converged into the \texttt{EcoCycle} framework, encapsulating self-healing, eco-responsive, and AI-optimized infrastructure solutions. The structured composition ensured that emergent discoveries were not mere aggregations but cohesive, context-aware innovations, validating the methodological rigor of the compositional approach. Using a strategy of adhering to systematic composition principles, the method used here demonstrates how interdisciplinary insights can be synthesized into scientific concepts.

For comparison, \hyperlink{supporting_text_3}{Supporting Text 3} shows the same experiment but where we use \texttt{o1-pro} in the final step of synthesis. 

Putting this into context, earlier work~\cite{fodor1988connectionism,zhao2024compositional,shi2025cryptox,xu2024largelanguagemodelscompositional} have highlighted significant limitations in large language models (LLMs) concerning their ability to perform systematic compositional reasoning, particularly in domains requiring logical integration and generalization. Our approach directly addresses these deficiencies by structuring reasoning processes in a progressive and interpretable manner. Despite possessing individual components of knowledge, LLMs often struggle to integrate these dynamically to detect inconsistencies or solve problems requiring novel reasoning paths. We mitigate this by explicitly encoding relationships between concepts within a graph structure. Unlike conventional LLMs that rely on associative pattern recognition or statistical co-occurrence~\cite{fodor1988connectionism}, our structured approach mitigates the concerns of mere connectionist representations by enforcing rule-based, interpretable generalization mechanisms that allow for dynamic recombination of learned knowledge in novel contexts. Further, our approach ensures that each reasoning step builds upon prior knowledge in a structured hierarchy. Steps A-D in our framework progressively construct solutions by leveraging explicit connections between concepts, enforcing compositionality rather than assuming it. For example, our approach connects biodegradable microplastic materials with self-healing materials, not merely through surface-level similarities but through defined mechanisms such as thermoreversible gelation and environmental interactions. Instead of expecting an LLM to infer relationships in a single step, our agentic model progressively traverses reasoning graphs, ensuring that the final outcome emerges through logically justified intermediary steps. This not only reduces reliance on pattern memorization but also enhances interpretability and robustness in novel scenarios.  

Our model further enhances compositional reasoning through three key mechanisms: 
\begin{enumerate}
    \item \textbf{Explicit Pathway Construction:} By mapping dependencies between concepts in a structured graph, our model ensures that each step in the reasoning process is explicitly defined and logically connected.
    \item \textbf{Adaptive Contextual Integration:} Instead of treating reasoning steps as isolated tasks, the model dynamically integrates intermediate results to refine its conclusions, ensuring that errors or inconsistencies in earlier stages are corrected before final predictions.
    \item \textbf{Hierarchical Synergy Identification:} Our model analyzes multi-domain interactions through graph traversal and thereby identify emergent patterns that standard LLMs would overlook, enabling more robust and flexible reasoning. These mechanisms collectively establish a reasoning framework that mitigates compositional deficiencies and facilitates the structured synthesis of knowledge.
\end{enumerate}
Table \ref{tab:comparison} summarizes how our approach directly addresses key LLM limitations identified in earlier work.

\begin{table}[h]
    \centering
    \begin{tabular}{|p{5cm}|p{10cm}|}
        \hline
        \textbf{Conventional LLM} & \textbf{How Our Model Addresses It} \\
        \hline
        Fails to compose multiple reasoning steps into a coherent process & Uses hierarchical reasoning with Steps A-D, ensuring progressive knowledge integration through structured dependencies. \\
        \hline
        Struggles to generalize beyond memorized patterns & Uses explicit graph structures to enforce systematic knowledge composition, allowing for novel reasoning paths. \\
        \hline
        Overfits to reasoning templates, failing on unseen reformulations & Introduces pairwise and bridge synergies to enable dynamic recombination of knowledge through structured traversal and adaptive reasoning. \\
        \hline
        Does not simulate "slow thinking" or iterative reasoning well & Implements an agentic model that explicitly traverses a reasoning graph rather than relying on a single forward pass, ensuring each step refines and validates prior knowledge. \\
        \hline
    \end{tabular}
    \caption{Comparison of limitations of conventional LLMs, and our approach addresses these. By explicitly structuring relationships between concepts, breaking down reasoning into progressive steps, and incorporating dynamic knowledge recombination, our approach achieves a higher level of structured compositionality that conventional LLMs struggle with. Future work could further refine this approach by introducing adaptive feedback loops, reinforcing causal reasoning, and incorporating quantitative constraints to strengthen knowledge synergies.}
    \label{tab:comparison}
\end{table}

Further analysis of these is left to future work, as they would exceed the scope of the present paper. The experiments show that principled approaches to expand knowledge can indeed be implemented using the methodologies described above, complementing other recent work that has explored related topics~\cite{BuehlerGraphReasoning2024,shi2025cryptox,Buehler2025GraphAwareGPT,xu2024largelanguagemodelscompositional,fodor1988connectionism}. 

\subsection{Utilization of Graph Reasoning over Key Hubs and Influencer Nodes in Response Generation}

In this example, we analyze the knowledge graph $\mathcal{G}_2$ using \texttt{NetworkX} to compute node centralities (betweenness and eigenvector centrality), identifying key hubs and influencers. Community detection via the Louvain method partitions the graph into conceptual clusters, extracting representative nodes per community. 

Key relationships are identified by examining high-centrality nodes and their strongest edges. These insights are formatted into a structured context and integrated into a task-specific prompt for LLM reasoning on impact-resistant materials, the same prompt that was used to construct the original graph. 

\begin{figure}[h]
    \centering
     \includegraphics[width=.9\textwidth]{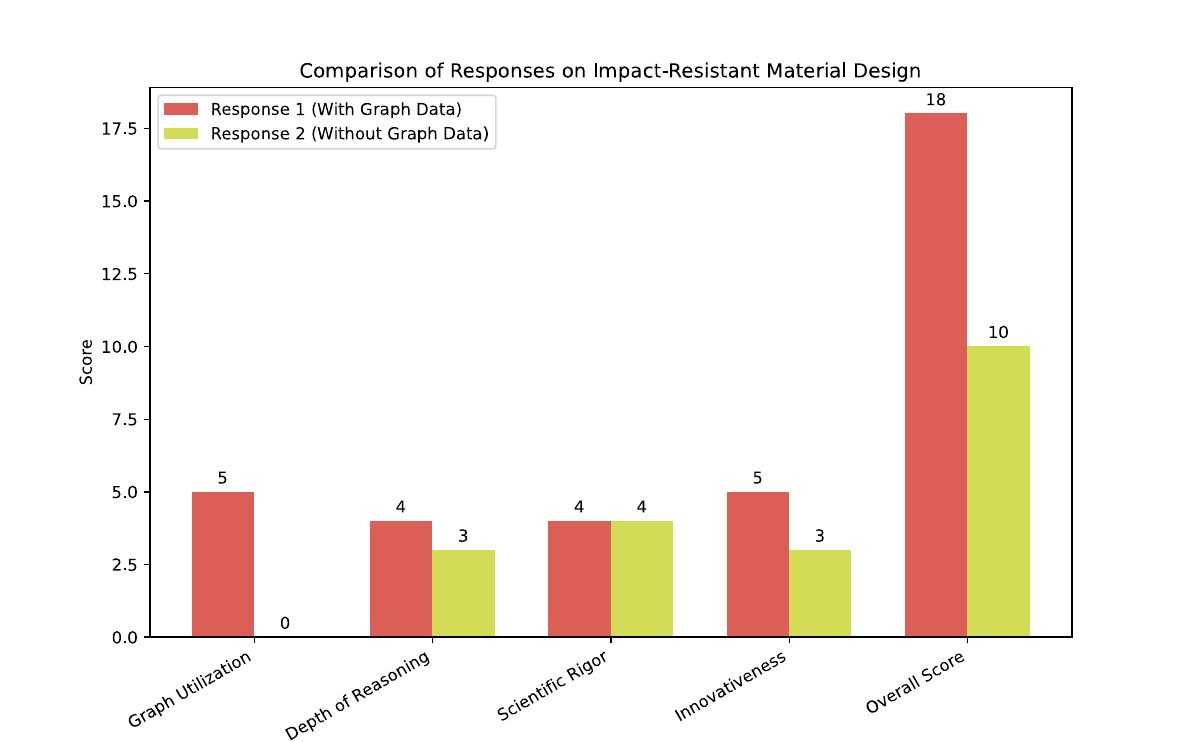}
    \caption{Comparison of Responses on Impact-Resistant Material Design. This  plot compares two responses based on four key evaluation metrics: Graph Utilization, Depth of Reasoning, Scientific Rigor, and Innovativeness, along with the overall score. Response 1, which incorporates graph-based insights, AI/ML techniques, and interdisciplinary approaches, outperforms Response 2 in all categories. Response 2 follows a more conventional materials science approach without leveraging computational methods. The higher overall score of Response 1 highlights the benefits of integrating advanced data-driven methodologies in material design. }
    \label{fig:response_analysis_graph}
\end{figure}

The model's response is generated both with and without graph data, followed by a comparative evaluation based on graph utilization, depth of reasoning, scientific rigor, and innovativeness. Raw responses for both models are shown in Text Boxes~\ref{TexBox:graph-response} and \ref{TexBox:base-response}. Table~\ref{tab:response_comparison} provides a detailed comparison, and Figure~\ref{fig:response_analysis_graph} compares responses based on four
key evaluation metrics (Graph Utilization, Depth of Reasoning, Scientific Rigor, and Innovativeness, along with the overall score).

\subsection{Use of an Agentic Deep Reasoning Model to Generate new Hypotheses and Anticipated Material Behavior}

Next, we use the SciAgents model~\cite{ghafarollahi2024sciagentsautomatingscientificdiscovery} with the \texttt{o3-mini} reasoning model~\cite{o3-mini-model-card-2025} as the back-end, and graph $\mathcal{G_2}$to answer this question: \texttt{Create a research idea around impact resistant materials and resilience. Rate the novelty and feasibility in the end.}  

The path-finding algorithm that integrates node embeddings and a degree of randomness to enhance exploration sampling strategy~\cite{ghafarollahi2024sciagentsautomatingscientificdiscovery} extracts this sub-graph from the larger graph:

\begin{LLMbox}{}
\begin{lstlisting}
Impact Resistant Materials -- IS-A -- Materials -- IS-A -- Impact-Resistant Materials -- INFLUENCES -- Modular Infrastructure Systems -- RELATES-TO -- Self-Healing Materials -- RELATES-TO -- Long-term Sustainability and Environmental Footprint of Infrastructure -- RELATES-TO -- Self-Healing Materials -- RELATES-TO -- Infrastructure -- IS-A -- Infrastructure Resilience -- RELATES-TO -- Smart Infrastructure -- RELATES-TO -- Impact-Resistant Materials -- RELATES-TO -- Machine Learning Algorithms -- RELATES-TO -- Impact-Resistant Materials -- RELATES-TO -- Resilience
\end{lstlisting}
\end{LLMbox}

As described in~\cite{ghafarollahi2024sciagentsautomatingscientificdiscovery} paths are sampled using a path-finding algorithm that utilizes both node embeddings and a degree of randomness to enhance exploration as a path is identified between distinct concepts. Critically, instead of simply identifying the shortest path, the algorithm introduces stochastic elements by selecting waypoints and modifying priority queues in a modified version of Dijkstra’s algorithm. This allows for the discovery of richer and more diverse paths in a knowledge graph. The resulting paths serve as the foundation for graph-based reasoning specifically geared towards research hypothesis generation, ensuring a more extensive and insightful exploration of scientific concepts.

\begin{figure}[h]
    \centering
     \includegraphics[width=.85\textwidth]{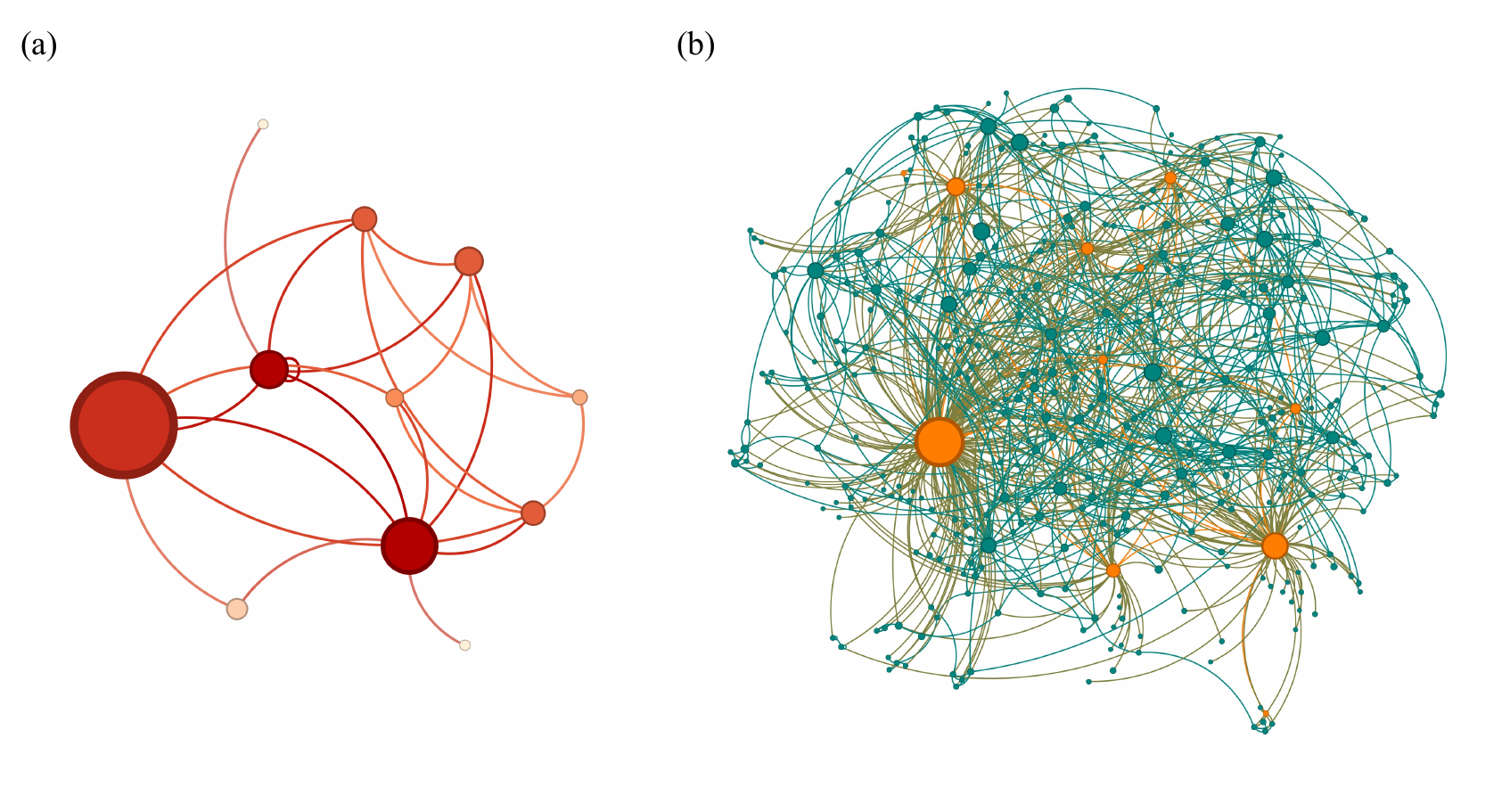}
    \caption{Visualization of subgraphs extracted from $\mathcal{G}_2$ by SciAgents, for use in graph reasoning. The left panel (a) represents the primary subgraph containing only nodes from the specified reasoning path. Node size is proportional to the original degree in the full network, highlighting key entities with high connectivity. The structure is sparse, with key nodes acting as central hubs in the reasoning framework. The right panel (b) represents an expanded subgraph that includes second-hop neighbors. Nodes from the original subgraph are colored orange, while newly introduced second-hop nodes are green. The increased connectivity and density indicate the broader network relationships captured through second-hop expansion. Larger orange nodes remain dominant in connectivity, while green nodes form supporting structures, emphasizing peripheral interactions and their contribution to knowledge propagation. This visualization highlights how expanding reasoning pathways in a graph framework integrates additional contextual information, enriching the overall structure..}
    \label{fig:fig_500}
\end{figure}

Visualizations of the subgraph are shown in Figure~\ref{fig:fig_500}, depicting the subgraph alone (Figure~\ref{fig:fig_500}(a)) and the subgraph with second hops (Figure~\ref{fig:fig_500}(b), showing the deep interconnectness that can be extracted). 

The resulting document \hyperlink{supporting_text_4}{Supporting Text 4} presents the results of applying SciAgents to $\mathcal{G}_2$  in the context of impact-resistant materials and infrastructure resilience. The graph representation serves as a structured framework for reasoning about the relationships between key concepts—impact-resistant materials, self-healing mechanisms, machine learning optimization, and modular infrastructure—by encoding dependencies and influences between them. Graph 2 specifically captures these interconnected domains as nodes, with edges representing logical or causal links, enabling a systematic exploration of pathways that lead to optimal material design strategies. The path traversal within the graph identifies key dependencies, such as how impact-resistant materials influence infrastructure resilience or how machine learning refines self-healing efficiency. This structured pathway-based reasoning allows SciAgents to generate research hypotheses that maximize cross-domain synergies, ensuring that material properties are not optimized in isolation but rather in concert with their broader applications in engineering and sustainability. Furthermore, graph traversal reveals emergent relationships—such as how integrating real-time sensor feedback into modular infrastructure could create self-improving materials—that might not be immediately evident through conventional linear analysis. Thus, the use of graph-based reasoning is pivotal in formulating a research framework that is not only interdisciplinary but also systematically optimized for long-term infrastructure resilience and material adaptability.

In terms of specific content, the proposed research explores an advanced composite material that integrates carbon nanotube (CNT)-reinforced polymer matrices with self-healing microcapsules, embedded sensor networks, and closed-loop ML optimization. The goal is to create a dynamically self-improving material system that enhances impact resistance and longevity in modular infrastructure. The material design is structured around several key components: (1) CNT reinforcement (1–2 wt\%) to improve tensile strength and fracture toughness, (2) self-healing microcapsules (50–200 $\mu$m) filled with polymerizable agents, (3) embedded graphene-based or PVDF strain sensors for real-time monitoring, and (4) adaptive ML algorithms that regulate stress distributions and healing responses.

The proposal establishes interconnections between several domains, highlighting the interdisciplinary nature of the research: impact-resistant materials are a subset of general materials with enhanced energy dissipation properties, modular infrastructure benefits from these materials due to increased durability, self-healing materials reduce maintenance cycles, and machine learning optimizes real-time responses to structural stress. This holistic framework aims to advance infrastructure resilience and sustainability. The research hypothesizes that embedding self-healing microcapsules within a CNT-reinforced polymer matrix will yield a composite with superior impact resistance and adaptive repair capabilities. Expected performance gains include a 50\% increase in impact energy absorption (surpassing 200 J/m²), up to 80\% recovery of mechanical properties after micro-damage, an estimated 30\% improvement in yield strain, a 50\% extension in structural lifetime, and a 30\% reduction in required maintenance interventions.

\begin{figure}
\centering
\sffamily
\scriptsize
\sansmath

\begin{tikzpicture}[
    node distance=0.5cm and .5cm, auto,
    event/.style={rectangle, draw, fill=blue!20, text width=5cm, text centered, rounded corners, minimum height=1.2cm},
    sensing/.style={rectangle, draw, fill=cyan!30, text width=5cm, text centered, rounded corners, minimum height=1.2cm},
    ml_analysis/.style={rectangle, draw, fill=violet!40, text width=5cm, text centered, rounded corners, minimum height=1.2cm},
    adjustment/.style={rectangle, draw, fill=violet!20, text width=5cm, text centered, rounded corners, minimum height=1.5cm},
    healing/.style={rectangle, draw, fill=green!30, text width=5cm, text centered, rounded corners, minimum height=1.5cm},
    feedback/.style={rectangle, draw, fill=yellow!30, text width=5cm, text centered, rounded corners, minimum height=1.2cm},
    line/.style={draw, -stealth', shorten >=1pt},
    dashed line/.style={draw, dashed, shorten >=1pt}]

    \node [event] (impact) {\textbf{Impact Event} \\ (Material undergoes structural stress or damage)};
    \node [sensing, below=of impact] (sensors) {\textbf{Sensor Detection} \\ (Real-time strain monitoring via embedded graphene/PVDF sensors)};
    \node [ml_analysis, below=of sensors] (ml) {\textbf{Machine Learning Analysis} \\ (Prediction of stress distribution, micro-damage evolution)};
    \node [adjustment, below=of ml] (adjustment) {\textbf{Healing Response Adjustment} \\ (ML-optimized activation of microcapsules based on sensor data)};
    \node [healing, below=of adjustment] (repair) {\textbf{Microcapsule Rupture and Repair} \\ (Self-healing agent polymerization to restore mechanical integrity)};
    \node [feedback, below=of repair] (performance) {\textbf{Material Performance Feedback} \\ (Updated data informs next optimization cycle)};

    \path [line] (impact) -- (sensors);
    \path [line] (sensors) -- (ml);
    \path [line] (ml) -- (adjustment);
    \path [line] (adjustment) -- (repair);
    \path [line] (repair) -- (performance);

    \path [dashed line] (performance.east) to[out=0, in=0] node[right, text width=4.5cm, align=center] {\textbf{Adaptive Learning Cycle:} \\ Sensors collect new data, \\ ML refines healing response} (ml.east);

\end{tikzpicture}
\sffamily
\caption{Flowchart of the Self-Optimizing Composite System proposed by SciAgents after reasoning over $\mathcal{G}_2$. 
Upon an impact event, embedded sensors (cyan) detect strain changes and transmit real-time data to a machine learning system (violet). 
This system predicts stress evolution and dynamically adjusts healing response thresholds (light violet). 
Microcapsules containing polymerizable agents (green) rupture at critical points, autonomously restoring material integrity. 
A feedback mechanism (yellow) continuously refines the process, ensuring adaptive optimization over multiple impact cycles. 
The dashed feedback loop signifies that each iteration improves the material’s ability to predict and mitigate future stress events, making the system progressively more efficient.}
\label{fig:composite_flowchart}
\end{figure}
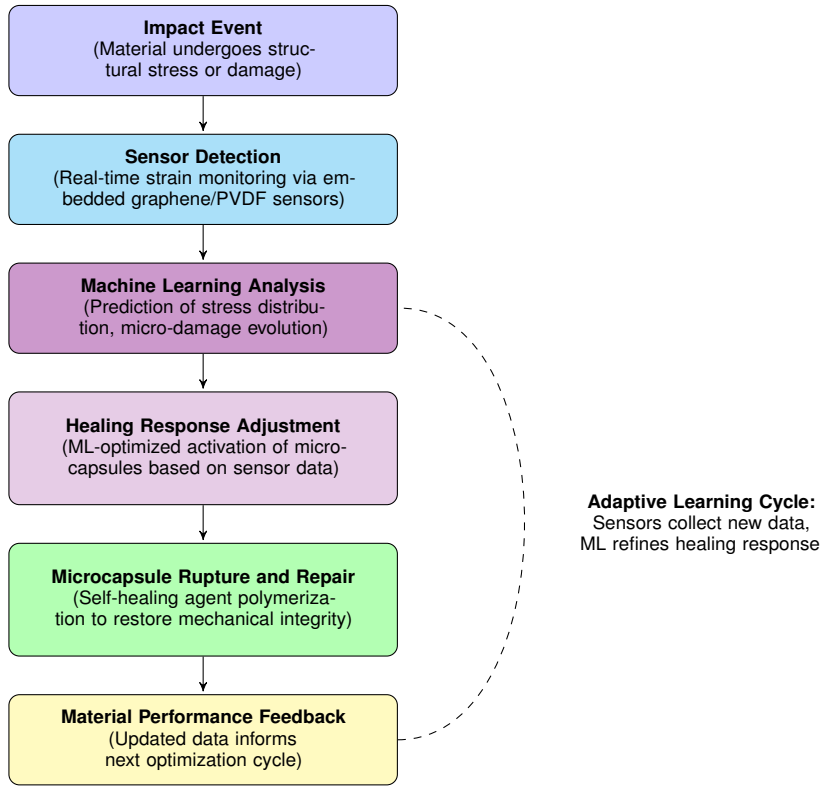

The composite operates via a multi-scale integration strategy where nanoscale CNTs form a stress-bridging network, microscale healing agents autonomously restore structural integrity, and macroscale sensors collect real-time strain data to inform machine learning-based optimizations. The closed-loop ML system refines material responses dynamically, preemptively addressing stress concentrations before catastrophic failure occurs. This iterative self-optimization process is represented in the  flowchart shown in Figure~\ref{fig:composite_flowchart}. 

Compared to conventional high-performance composites such as ultra-high molecular weight polyethylene (UHMWPE) and standard carbon fiber-reinforced polymers, the proposed material demonstrates superior mechanical performance and autonomous damage remediation. Traditional impact-resistant materials typically absorb 120–150 J/m² of energy, whereas this system is designed to exceed 200 J/m². Additionally, existing self-healing materials recover only 50–60\% of their mechanical properties, while this composite targets an 80\% restoration rate. The modular design ensures seamless integration into existing infrastructure, supporting scalability and standardization.

Beyond its core functions, the composite exhibits several emergent properties: (1) localized reinforcement zones where healing chemistry alters stress distributions, (2) increased energy dissipation efficiency over repeated impact cycles, (3) long-term self-improving feedback where ML-driven adjustments refine material performance, and (4) potential microstructural evolution, such as crystalline phase formation, that enhances impact resistance. These unexpected yet beneficial attributes highlight the adaptive nature of the material system.

The broader implications of this research include significant economic and environmental benefits. By reducing maintenance frequency by 30\%, the composite lowers infrastructure downtime and lifecycle costs. The extended service life translates to a 25–30\% reduction in resource consumption and associated carbon emissions. While the upfront processing cost is higher due to advanced material fabrication and sensor integration, the long-term cost per operational year is projected to be competitive with, or superior to, existing alternatives.

This interdisciplinary fusion of nanomaterials, self-healing chemistry, real-time sensor feedback, and machine learning-based control represents a fundamental shift from passive materials to smart, self-optimizing systems. The proposed research not only addresses impact resistance and self-repair but also pioneers an adaptable, continuously improving infrastructure material. The combination of rigorous experimental validation (e.g., ASTM mechanical testing, finite element modeling, and real-world simulations) ensures that the material’s theoretical advantages translate into practical performance gains. This research positions itself as a transformative solution for infrastructure resilience, bridging the gap between static engineering materials and dynamically intelligent, self-regulating composites.

\section{Conclusion}

This work introduced a framework for recursive graph expansion, demonstrating that self-organizing intelligence-like behavior can emerge through iterative reasoning without predefined ontologies, external supervision, or centralized control. Unlike conventional knowledge graph expansion techniques that rely on static extractions, probabilistic link predictions, or reinforcement learning-based traversal, extensive test-time compute Graph-PReFLexOR graph reasoning actively restructures its own knowledge representation as it evolves, allowing for dynamic adaptation and autonomous knowledge synthesis. These findings are generally in line with other recent results that elucidated the importance of inference scaling methods~\cite{o1-model-card-2024,o3-mini-model-card-2025,geiping2025scalingtesttimecomputelatent,buehler2024preflexorpreferencebasedrecursivelanguage}. 

Through extensive graph-theoretic analysis, we found that the recursively generated knowledge structures exhibit scale-free properties, hierarchical modularity, and sustained interdisciplinary connectivity, aligning with patterns observed in human knowledge systems. The formation of conceptual hubs (Figures~\ref{fig:graph_analysis}-\ref{fig:graph_modularity}) and the emergence of bridge nodes (Figures~\ref{fig:bridge_persistence}) demonstrate that the system autonomously organizes information into a structured yet flexible network, facilitating both local coherence and global knowledge integration. Importantly, the model does not appear to saturate or stagnate; instead, it continuously reorganizes relationships between concepts by reinforcing key conceptual linkages while allowing new hypotheses to emerge through iterative reasoning (Figures~\ref{fig:knowledge_graph_evolution} and \ref{fig:bridge_node_evolution}).

One of the most striking findings is the self-regulation of knowledge propagation pathways. The early stages of graph expansion relied heavily on a few dominant nodes (high betweenness centrality), but over successive iterations, knowledge transfer became increasingly distributed and decentralized (Figure~\ref{fig:betweenness_distribution}). This structural transformation suggests that recursive self-organization naturally reduces bottlenecks, enabling a more resilient and scalable knowledge framework. Additionally, we observed alternating phases of conceptual stability and breakthrough, indicating that knowledge formation follows a punctuated equilibrium model, rather than purely incremental accumulation.

More broadly, the recursive self‐organization process produces emergent, fractal‐like knowledge structures, suggesting that similar principles may underlie both human cognition and the design of intelligent systems~\cite{barabasi1999emergence}. Moreover, the potential role of bridge nodes—as connectors and as natural intervention points—is underscored by their persistent yet shifting influence, implying they could be strategically targeted for system updates or error correction in a self‐organizing network. Additionally, the observed alternating phases of stable community formation punctuated by sudden breakthroughs appear to mirror the concept of punctuated equilibrium in scientific discovery~\cite{kuhn1962structure}, offering a promising framework for understanding the natural emergence of innovation. These insights extend the implications of our work beyond scientific discovery, hinting at broader applications in autonomous reasoning, such as adaptive natural language understanding and real‐time decision‐making in complex environments. We demonstrated a few initial use cases where we used graph structures in attempts towards compositional reasoning, as shown in Figure~\ref{fig:compositional_analysis}.

\subsection{Graph Evolution Dynamics: Interplay of Network Measures}
The evolution of the knowledge graph reveals a complex interplay between growth, connectivity, centralization, and structural reorganization, with different network-theoretic measures exhibiting distinct yet interdependent behaviors over iterations. Initially, the system undergoes rapid expansion, as seen in the near-linear increase in the number of nodes and edges (Figure~\ref{fig:graph_analysis}). However, despite this outward growth, the clustering coefficient stabilizes early (around 0.16), suggesting that the graph maintains a balance between connectivity and modularity rather than devolving into isolated clusters. This stabilization indicates that the system does not expand chaotically but instead integrates new knowledge in a structured and preferentially attached manner, reinforcing key concepts while allowing for exploration.

One of the most informative trends is the evolution of betweenness centrality (Figure~\ref{fig:betweenness_evolution}), which starts highly concentrated in a few key nodes but then redistributes over time, reflecting a transition from hub-dominated information flow to a more decentralized and resilient network. This shift aligns with the gradual stabilization of average shortest path length (around 4.5, see Figure~\ref{fig:shortest_path_distribution}) and the graph diameter (around 16–18 steps, see Figure~\ref{fig:graph_modularity}), implying that while knowledge expands, it remains navigable and does not suffer from excessive fragmentation. Meanwhile, the maximum $k$-core index (Figure~\ref{fig:advanced_graph_analysis}) exhibits a stepwise increase, reflecting structured phases of densification where core knowledge regions consolidate before expanding further. This suggests that the system undergoes punctuated reorganization, where newly introduced concepts occasionally necessitate internal restructuring before further outward growth.

Interestingly, the degree assortativity starts strongly negative (around -0.25) and trends toward neutrality (-0.05), indicating that high-degree nodes initially dominate connections but later distribute their influence, allowing mid-degree nodes to contribute to network connectivity. This effect is reinforced by the persistence of bridge nodes (Figures~\ref{fig:advanced_graph_analysis}-\ref{fig:betweenness_evolution}), where we see a long-tail distribution of interdisciplinary connectors—some nodes serve as transient links that appear briefly, while others persist across hundreds of iterations, indicating stable, high-impact conceptual connectors.

Taken together, these experimentally observed trends suggest that the system self-regulates its expansion, dynamically shifting between growth, consolidation, and reorganization phases. The absence of saturation in key structural properties (such as new edge formation and bridge node emergence) indicates that the model supports continuous knowledge discovery, rather than converging to a fixed-state representation. This emergent behavior, where network-wide connectivity stabilizes while conceptual expansion remains open-ended, suggests that recursive graph reasoning could serve as a scalable foundation for autonomous scientific exploration, adaptive learning, and self-organizing knowledge systems.

\subsection{Relevance in the Context of Materials Science}
The framework introduced in this work offers a novel paradigm for accelerating discovery in materials science by systematically structuring and expanding knowledge networks. Unlike traditional approaches that rely on static databases or predefined ontologies~\cite{buehler2023biomateriomics,tshitoyan2019unsupervised,Buehler2022GeneratingCues,Buehler2023PredictingCapacityb,Brinson2023CommunityResearchb}, our self-organizing method enables dynamic hypothesis generation, uncovering hidden relationships between material properties, synthesis pathways, and functional behaviors. The emergent scale-free networks observed in our experiments reflect the underlying modularity and hierarchical organization often seen in biological and engineered materials, suggesting that recursive graph-based reasoning could serve as a computational analogue to self-assembling and adaptive materials. 
Applied to materials design, the  approach developed in this paper could reveal unexpected synergies between molecular architectures and macroscale performance, leading to new pathways for bioinspired, multifunctional, and self-healing materials. Future work can integrate experimental data directly into these reasoning loops, allowing AI-driven materials discovery to move beyond retrieval-focused recognition toward novel inference and innovation. We believe it is essential to bridge the gap between autonomous reasoning and materials informatics to ultimately create self-improving knowledge systems that can adaptively guide materials engineering efforts in real-time~\cite{Stach2021AutonomousPerspective}.

\subsection{Broader Implications}
The observations put forth in this paper have potential implications for AI-driven scientific reasoning, autonomous hypothesis generation, and scientific inquiry. As our results demonstrate, complex knowledge structures can self-organize without explicit goal-setting. This work challenges a prevailing assumption that intelligence requires externally imposed constraints or supervision. Instead, it suggests that intelligent reasoning may emerge as a fundamental property of recursive, feedback-driven information processing, mirroring cognitive processes observed in scientific discovery and human learning. Our experiments that directed the evolution of the thinking mechanisms towards a certain goal were provided with relational modeling that incorporated these concepts in a more pronounced manner, as expected, provisioning a powerful substrate for deeper reasoning. 

Future work could potentially explore extending this framework to multi-agent reasoning environments, cross-domain knowledge synthesis, and real-world applications in AI-driven research discovery. Additionally, refining interpretability mechanisms will be crucial for ensuring that autonomously generated insights align with human epistemic standards, minimizing risks related to misinformation propagation and reasoning biases. Bridging graph-theoretic modeling, AI reasoning, and self-organizing knowledge dynamics, allowed us to provide a step toward building AI systems capable of autonomous, scalable, and transparent knowledge formation on their own.

We note that wile our agentic deep graph reasoning framework demonstrates promise in achieving self-organizing knowledge formation, several challenges remain. In particular, the computational scalability of recursive graph expansions and the sensitivity of emergent structures to parameter choices warrant further investigation. Future work should explore robust error-correction strategies, enhanced interpretability of evolving networks, and ethical guidelines to ensure transparency in autonomous reasoning systems, especially if deployed in commercial or public settings beyond academic research. Addressing these issues will not only refine the current model but also paves the way for its application in real-world autonomous decision-making and adaptive learning environments.

\section{Materials and Methods}

We describe key materials and methods developed and used in the course of this study in this section. 

\subsection{Graph-PReFLexOR model development} 

A detailed account of the Graph-PReFLexOR is provided in \cite{buehler2025insitugraphreasoningknowledge}. Graph-PReFLexOR (Graph-based Preference-based Recursive Language Modeling for Exploratory Optimization of Reasoning) is an AI model integrating \textit{in-situ} graph reasoning, symbolic abstraction, and recursive reflection into generative modeling. The model was trained on a set of around 1,000 scientific papers in the biological materials and bio-inspired materials domain, as discussed in \cite{buehler2025insitugraphreasoningknowledge}. We refer readers to the original paper for implementation details, but provide a high-level summary here. The method defines reasoning as a structured mapping:
\begin{equation}
M: T \rightarrow (G, P, A),
\end{equation}
where a given task $T$ generates a knowledge graph $G = (V, E)$ with nodes $V$ representing key concepts and edges $E$ denoting relationships, abstract patterns $P$ capturing structural dependencies, and final answers $A$. Inspired by category theory, the approach encodes knowledge through hierarchical inference, leveraging isomorphisms to generalize across domains. The model autonomously constructs symbolic representations via a reasoning phase marked by \texttt{<|thinking|>} ... \texttt{<|/thinking|>} tokens, refining understanding before generating outputs. Recursive optimization can further improve logical coherence, aligning responses with generalizable principles, a particular feature that will be expanded on in this paper.

To enhance the adaptability of structured reasoning, Graph-PReFLexOR employs an iterative feedback mechanism:
\begin{equation}
R_{i+1} = f_{\text{eval}}(R_i, F_i),
\end{equation}
where $R_i$ denotes the intermediate reasoning at step $i$, $F_i$ is the feedback applied to improve logical structure, and $f_{\text{eval}}$ evaluates alignment with domain principles. The final answer $A$ is derived after $N$ refinements as:
\begin{equation}
A = g(R_N).
\end{equation}
Through the idea to explicitly model knowledge graphs and symbolic representations, this method attempts to bridge connectionist and symbolic paradigms, facilitating multi-step reasoning, hypothesis generation, and interdisciplinary knowledge expansion. Empirical evaluations in~\cite{buehler2025insitugraphreasoningknowledge} demonstrated its capability to generalize beyond training data. In this study, we take advantage of the capability of Graph-PReFLexOR to generate graph representations on the fly over a great number of iterations during which the model continues to expand its reasoning tokens.  

\subsection{Iterative Unconstrained Graph Reasoning on General Topic }

We develop an iterative knowledge extraction pipeline to construct a structured knowledge graph using a LLM, following the flowchart shown in Figure~\ref{fig:fig_01}. The method systematically expands a graph representation of relationships by extracting structured knowledge from model-generated reasoning sequences and generating follow-up queries to refine exploration. We use this method to construct $\mathcal{G_1}$. 

At the start of each run, the algorithm initializes an initial question or prompt. This can be very general or focus on a particular topic that defines the area of scientific inquiry. In the example, the topic is set as:

\begin{LLMbox}{}
\begin{lstlisting}
prompt = "Discuss an interesting idea in bio-inspired materials science."  
\end{lstlisting}
\end{LLMbox}

The LLM then generates structured reasoning responses within the \texttt{<|thinking|>} ... \texttt{<|/thinking|>} tokens. The response is processed to extract structured knowledge by isolating the graph.

To convert the extracted knowledge into a structured representation, the model is queried with an additional instruction to transform the resulting raw text that contains the reasoning graph (denoted by \texttt{\{raw graph\}}) into a Python dictionary formatted for graph representation:

\begin{LLMbox}{}
\begin{lstlisting}
You are an AI that extracts information from structured text and outputs a graph in Python dictionary format compatible with NetworkX. 
Given the following structured text: 
(*@\bf\hlred{\{raw graph\}}@*)
Output the graph as a Python dictionary without any additional text or explanations. Ensure the dictionary is properly formatted for immediate evaluation in Python.
\end{lstlisting}
\end{LLMbox}

The output is parsed and structured using \texttt{ast.literal\_eval()} to construct a directed graph $\mathcal{G}_{\text{local}}^i$ in \texttt{NetworkX}, where nodes represent entities such as materials, properties, and scientific concepts, while edges encode relationships such as \texttt{HAS}, \texttt{INFLUENCES}, and \texttt{SIMILAR-TO}.

At each iteration $i$, the newly extracted knowledge graph is appended to an evolving global graph:
\begin{equation}
\mathcal{G} \leftarrow \mathcal{G} \cup \mathcal{G}_{\text{local}}^i.
\label{eq:graph_rek}
\end{equation}

The extracted structure is parsed using:
\begin{quote}
\texttt{graph\_code, graph\_dict = extract\_graph\_from\_text(graph)}
\end{quote}
The graph is progressively expanded by adding newly introduced nodes and edges, ensuring that redundant relationships are not duplicated. The final knowledge graph is stored in multiple formats, including GraphML for structural analysis and PNG for visualization.

To facilitate continued exploration, a follow-up question is generated at each iteration. The LLM is queried to produce a question that introduces a new aspect of the domain, ensuring an iterative, self-refining process that utilizes the previously generated entities and relations:

\begin{LLMbox}{}
\begin{lstlisting}
Consider this list of topics/keywords. Formulate a creative follow-up question to ask about a totally new concept. 
Your question should include at least one of the original topics/keywords.
Original list of topics/keywords:
(*@\bf\hlred{\{latest extracted entities and relations\}}@*)
Reply only with the new question. The new question is:
\end{lstlisting}
\end{LLMbox}

This ensures that subsequent queries remain contextually grounded in the domain while promoting scientific discovery. The generated question is appended to the reasoning token structure and fed back into the LLM, thereby continuing the iterative learning process.

The algorithm runs for a total of $N$ iterations, progressively refining the knowledge graph. At each step, we track the growth of the graph by recording the number of nodes and edges over time. The final knowledge graph provides a structured and extensible representation of insights extracted from the LLM, enabling downstream analysis of emerging concepts. The reasoning process (Figure~\ref{fig:fig_01}) unfolds sequentially over a period of several days (using a consumer GPU, like NVIDIA A6000 Ada).

\subsection{Iterative Graph Reasoning on a Particular Topic}

As an alternative to the approach above, we can tailor the reasoning process to focus more strongly on a particular topic. We use this method to construct $\mathcal{G_2}$. For instance, at the beginning of each run, the algorithm is initialized with a user-defined topic:

\begin{LLMbox}{}
\begin{lstlisting}
topic = "impact resistant materials"
\end{lstlisting}
\end{LLMbox}

This variable defines the area of exploration and is dynamically incorporated into the model prompts. The LLM is then queried with a topic-conditioned instruction to generate structured reasoning tokens:

\begin{LLMbox}{}
\begin{lstlisting}
Describe a way to design (*@\bf\hlred{\{topic\}}@*).
\end{lstlisting}
\end{LLMbox}

The model generates textual responses that include explicit reasoning within the \texttt{<|thinking|>} ... \texttt{<|/thinking|>} markers. As before, from this output, we extract structured knowledge by isolating the section labeled graph, to extract entity-relationship pairs. A follow-up question is generated at each iteration to drive the discovery process forward. This prompt ensures that new queries focus on underexplored aspects of the knowledge graph while maintaining the topic-conditioned structure:
 
\begin{LLMbox}{}
\begin{lstlisting}
Consider this list of keywords. Considering the broad topic of (*@\bf\hlred{\{topic\}}@*), formulate a creative follow-up question to ask about a totally new aspect. Your question should include at least one of the original keywords. 
Original list of keywords:
(*@\bf\hlred{\{latest extracted entities and relations\}}@*)
Reply only with the new question. The new question is:
\end{lstlisting}
\end{LLMbox}

This ensures that each iteration remains contextually grounded in the specified domain while continuously expanding the knowledge graph.

The process continues for $N$ steps, progressively refining the knowledge graph. At each iteration, we track the growth of the graph by recording the number of nodes and edges. The resulting knowledge graph serves as a structured repository of insights extracted from the LLM, enabling downstream analysis of materials properties and design principles. 

Naturally, other variants of these strategies could easily be devised, for instance to create other generalist graphs (akin to $\mathcal{G}_1$) or specialized graphs (akin to $\mathcal{G}_2$). Prompt engineering can be human-tailored or developed agentically by other AI systems.

\subsection{Graph Analysis and Visualization}

Graph analysis and visualizations are conducted using \texttt{NetworkX}~\cite{Networkx/networkx:Python}, Gephi~\cite{ICWSM09154}, Cytoscope~\cite{shannon2003cytoscape}, Mermaid~\url{https://mermaid.js.org/}, and various plugins within these packages.

\subsubsection{Basic Analysis of Recursive Graph Growth over Reasoning Iterations}

To analyze the recursive expansion of the knowledge graph, we computed a set of graph-theoretic properties at each iteration using the \texttt{NetworkX} Python library. Graph data was stored in GraphML format, with filenames encoded to reflect the iteration number, allowing for chronological tracking of structural changes. Each graph was sequentially loaded and processed to extract key metrics that characterize its connectivity, topology, and hierarchical organization.

The fundamental properties of the graph, including the number of nodes and edges, were directly retrieved from the graph structure. The degree distribution was computed across all nodes to derive the average degree, representing the mean connectivity per node, and the maximum degree, which highlights the most connected node at each iteration. To assess network cohesion, the largest connected component (LCC) was extracted by identifying the largest strongly connected component in directed graphs and the largest connected subgraph in undirected cases. The clustering coefficient was computed using the standard local clustering metric, which quantifies the likelihood that a node’s neighbors are also connected to each other. The average clustering coefficient was obtained by averaging over all nodes in the graph, providing insight into the tendency of local structures to form tightly connected clusters.

To assess global connectivity and efficiency, we computed the average shortest path length (SPL) and the graph diameter within the largest connected component. The SPL was obtained by calculating the mean shortest path distance between all pairs of nodes in the LCC, while the diameter was determined as the longest shortest path observed in the component. Since these calculations are computationally expensive for large graphs, they were conditionally executed only when the LCC was sufficiently small or explicitly enabled in the analysis. For community detection, we applied the Louvain modularity algorithm using the \texttt{community-louvain} package. The graph was treated as undirected for this step, and the modularity score was computed by partitioning the graph into communities that maximize the modularity function. This metric captures the extent to which the graph naturally organizes into distinct clusters over iterations. 

The entire analysis pipeline iterated over a series of GraphML files, extracting the iteration number from each filename and systematically computing these metrics. The results were stored as time series arrays and visualized through multi-panel plots, capturing trends in network evolution. To optimize performance, computationally intensive operations, such as shortest path calculations and modularity detection, were executed conditionally based on graph size and software availability. To further examine the structural evolution of the recursively generated knowledge graph, we computed a set of advanced graph-theoretic metrics over iterative expansions. As before, the analysis was conducted over a series of iterations, allowing for the study of emergent network behaviors.

The degree assortativity coefficient was computed to measure the correlation between node degrees, assessing whether high-degree nodes preferentially connect to similar nodes. This metric provides insight into the network's structural organization and whether its expansion follows a preferential attachment mechanism. The global transitivity, defined as the fraction of closed triplets among all possible triplets, was calculated to quantify the overall clustering tendency of the graph and detect the emergence of tightly interconnected regions. To assess the hierarchical connectivity structure, we performed $k$-core decomposition, which identifies the maximal subgraph where all nodes have at least \( k \) neighbors. We extracted the maximum $k$-core index, representing the deepest level of connectivity within the network, and computed the size of the largest $k$-core, indicating the robustness of highly connected core regions.

For understanding the importance of individual nodes in information flow, we computed average betweenness centrality over the largest connected component. Betweenness centrality quantifies the extent to which nodes serve as intermediaries in shortest paths, highlighting critical nodes that facilitate efficient navigation of the knowledge graph. Since exact computation of betweenness centrality can be computationally expensive for large graphs, it was performed only within the largest component to ensure feasibility. Additionally, we identified articulation points, which are nodes whose removal increases the number of connected components in the network. The presence and distribution of articulation points reveal structural vulnerabilities, highlighting nodes that serve as key bridges between different knowledge regions.

\subsubsection{Prediction of Newly Connected Pairs}

To track the evolution of connectivity in the recursively expanding knowledge graph, we employed a random sampling approach to estimate the number of newly connected node pairs at each iteration. Given the computational cost of computing all-pairs shortest paths in large graphs, we instead sampled a fixed number of node pairs per iteration and measured changes in their shortest path distances over time.

\textbf{Sampling Strategy.} At each iteration, we randomly selected 1,000 node pairs from the current set of nodes in the global knowledge graph. For each sampled pair \((u, v)\), we computed the shortest path length in the graph using Breadth-First Search (BFS), implemented via \texttt{nx.single\_source\_shortest\_path\_length(G, src)}. If a path existed, its length was recorded; otherwise, it was marked as unreachable.

\textbf{Tracking Newly Connected Pairs.} To detect the formation of new connections, we maintained a record of shortest path distances from the previous iteration and compared them with the current distances. A pair \((u, v)\) was classified as:
\begin{itemize}
    \item \textbf{Newly connected} if it was previously unreachable (\(\text{dist}_\text{before} = \text{None}\)) but became connected (\(\text{dist}_\text{now} \neq \text{None}\)).
    \item \textbf{Having a shorter path} if its shortest path length decreased between iterations (\(\text{dist}_\text{now} < \text{dist}_\text{before}\)).
\end{itemize}
The number of newly connected pairs and the number of pairs with shortened paths were recorded for each iteration.

\textbf{Graph Integration and Visualization.} At each iteration, the newly processed graph was merged into a global knowledge graph, ensuring cumulative analysis over time. The number of newly connected pairs per iteration was plotted as a time series, revealing patterns in connectivity evolution. This method effectively captures structural transitions, particularly the initial burst of connectivity formation followed by a steady-state expansion phase, as observed in the results.

By employing this approach, we achieved a computationally efficient yet statistically robust estimate of network connectivity evolution, allowing us to analyze the self-organizing dynamics of the reasoning process over large iterative expansions.

\subsubsection{Graph Structure and Community Analysis}

To examine the structural properties of the recursively generated knowledge graph, we performed a comprehensive analysis of node connectivity, degree distribution, clustering behavior, shortest-path efficiency, and community structure. The graph was loaded from a \texttt{GraphML} file using the \texttt{NetworkX} library, and various metrics were computed to assess both local and global network properties.

\textbf{Basic Graph Properties.} The fundamental characteristics of the graph, including the number of nodes, edges, and average degree, were extracted. Additionally, the number of self-loops was recorded to identify redundant connections that may influence network dynamics.

\textbf{Graph Component Analysis.} To ensure robust connectivity analysis, the largest connected component (LCC) was extracted for undirected graphs, while the largest strongly connected component (SCC) was used for directed graphs. This ensured that further structural computations were performed on a fully connected subgraph, avoiding artifacts from disconnected nodes.

\textbf{Degree Distribution Analysis.} The degree distribution was computed and visualized using both a linear-scale histogram and a log-log scatter plot. The latter was used to assess whether the network exhibits a power-law degree distribution, characteristic of scale-free networks.

\textbf{Clustering Coefficient Analysis.} The local clustering coefficient, which quantifies the tendency of nodes to form tightly connected triads, was computed for each node. The distribution of clustering coefficients was plotted, and the average clustering coefficient was recorded to evaluate the extent of modular organization within the network.

\textbf{Centrality Measures.} Three centrality metrics were computed to identify influential nodes:
(i) \textit{Betweenness centrality}, which measures the extent to which nodes act as intermediaries in shortest paths, highlighting key connectors in the knowledge graph;
(ii) \textit{Closeness centrality}, which quantifies the efficiency of information propagation from a given node;
(iii) \textit{Eigenvector centrality}, which identifies nodes that are highly influential due to their connections to other high-importance nodes.

\textbf{Shortest Path Analysis.} The average shortest path length (SPL) and graph diameter were computed to evaluate the network’s navigability. Additionally, a histogram of sampled shortest path lengths was generated to analyze the distribution of distances between randomly selected node pairs (2,000 samples used).

\textbf{Community Detection and Modularity.} The Louvain modularity algorithm was applied (if available) to partition the network into communities and assess its hierarchical structure. The modularity score was computed to quantify the strength of the detected community structure, and the resulting partitions were visualized using a force-directed layout.

\subsubsection{Analysis of Conceptual Breakthroughs}

The evolution of knowledge graphs is analyzed by processing a sequence of graph snapshots stored in \texttt{GraphML} format. Each graph is indexed by an iteration number, extracted using a regular expression from filenames of the form \texttt{graph\_iteration\_\#.graphml}. The graphs are sequentially loaded and processed to ensure consistency across iterations. If the graph is directed, it is converted to an undirected format using the \texttt{networkx.to\_undirected()} function. To ensure structural integrity, we extract the largest connected component using the \texttt{networkx.connected\_components()} function, selecting the subgraph with the maximum number of nodes.

For each iteration \( t \), we compute the degree distribution of all nodes in the largest connected component. The degree of a node \( v \) in graph \( G_t = (V_t, E_t) \) is given by:
\begin{equation}
d_t(v) = \sum_{u \in V_t} A_t(v, u)
\end{equation}
where \( A_t \) is the adjacency matrix of \( G_t \). The computed degree distributions are stored in a dictionary and later aggregated into a pandas \texttt{DataFrame} for further analysis.

To track the emergence of top hubs, we define a node \( v \) as a hub if it attains a high degree at any iteration. The set of top hubs is determined by selecting the nodes with the highest maximum degree across all iterations:
\[
H = \{ v \mid \max_t d_t(v) \geq d_{\text{top}, 10} \}
\]
where \( d_{\text{top}, 10} \) is the degree of the 10th highest-ranked node in terms of maximum degree. The degree growth trajectory of each hub is then extracted by recording \( d_t(v) \) for all \( t \) where \( v \in V_t \).

To quantify the emergence of new hubs, we define an emergence threshold \( d_{\text{emerge}} = 5 \), considering a node as a hub when its degree first surpasses this threshold. The first significant appearance of a node \( v \) is computed as:
\[
t_{\text{emerge}}(v) = \min \{ t \mid d_t(v) > d_{\text{emerge}} \}
\]
for all \( v \) where such \( t \) exists. The histogram of \( t_{\text{emerge}}(v) \) across all nodes provides a temporal distribution of hub emergence.

To evaluate global network connectivity, we compute the mean degree at each iteration:
\begin{equation}
\bar{d}_t = \frac{1}{|V_t|} \sum_{v \in V_t} d_t(v)
\end{equation}
capturing the overall trend in node connectivity as the knowledge graph evolves.

Three key visualizations are generated: (1) the degree growth trajectories of top hubs, plotted as \( d_t(v) \) over time for \( v \in H \); (2) the emergence of new hubs, represented as a histogram of \( t_{\text{emerge}}(v) \); and (3) the overall network connectivity, visualized as \( \bar{d}_t \) over iterations.

\subsubsection{Structural Evolution of the Graphs: Knowledge Communities, Bridge Nodes and Multi-hop Reasoning}

We analyze the structural evolution of knowledge graphs by computing three key metrics: (1) the number of distinct knowledge communities over time, (2) the emergence of bridge nodes that connect different knowledge domains, and (3) the depth of multi-hop reasoning based on shortest path lengths. These metrics are computed for each iteration \( t \) of the evolving graph and visualized as follows.

The evolution of knowledge communities is measured using the Louvain modularity optimization algorithm, implemented via \texttt{community.best\_partition()}, which partitions the graph into distinct communities. For each iteration, the number of detected communities \( |C_t| \) is computed as:
\[
|C_t| = |\{ c \mid c = P_t(v), v \in V_t \}|
\]
where \( P_t(v) \) maps node \( v \) to its assigned community at iteration \( t \). The values of \( |C_t| \) are plotted over iterations to track the subdivision and merging of knowledge domains over time.

The emergence of bridge nodes, nodes that connect multiple communities, is determined by examining the community affiliations of each node's neighbors. A node \( v \) is classified as a bridge node if:
\[
|\mathcal{C}(v)| > 1, \quad \text{where} \quad \mathcal{C}(v) = \{ P_t(u) \mid u \in N(v) \}
\]
and \( N(v) \) represents the set of neighbors of \( v \). The number of bridge nodes is computed per iteration and plotted to analyze how interdisciplinary connections emerge over time.

The depth of multi-hop reasoning is quantified by computing the average shortest path length for the largest connected component at each iteration:
\[
L_t = \frac{1}{|V_t| (|V_t| - 1)} \sum_{v,u \in V_t, v \neq u} d_{\text{sp}}(v, u)
\]
where \( d_{\text{sp}}(v, u) \) is the shortest path distance between nodes \( v \) and \( u \), computed using \texttt{networkx.average\_shortest\_path\_length()}. This metric captures the evolving complexity of conceptual reasoning chains in the knowledge graph.

We generate three plots: (1) the evolution of knowledge communities, visualizing \( |C_t| \) over time; (2) the emergence of bridge nodes, displaying the number of inter-community connectors per iteration; and (3) the depth of multi-hop reasoning, tracking \( L_t \) as a function of iteration number.

To analyze the temporal stability of bridge nodes in the evolving knowledge graph, we compute the persistence of bridge nodes, which quantifies how long individual nodes function as bridges across multiple iterations. Given the bridge node set \( B_t \) at iteration \( t \), the persistence count for a node \( v \) is defined as:
\[
P(v) = \sum_{t} \mathbb{1} (v \in B_t)
\]
where \( \mathbb{1}(\cdot) \) is the indicator function that equals 1 if \( v \) appears as a bridge node at iteration \( t \), and 0 otherwise. This metric captures the frequency with which each node serves as a conceptual connector between different knowledge domains.

To visualize the distribution of bridge node persistence, we construct a histogram of \( P(v) \) across all detected bridge nodes, with kernel density estimation (KDE) applied for smoother visualization. The histogram provides insight into whether bridge nodes are transient or persist over multiple iterations.

The persistence values are computed and stored in a structured dataset, which is then used to generate a plot of the histogram of bridge node persistence.

\noindent To analyze the temporal dynamics of bridge node emergence, we construct a binary presence matrix that tracks when individual nodes first appear as bridges. The matrix is used to visualize the earliest bridge nodes over time, capturing the structural formation of key conceptual connectors.

The binary presence matrix is defined as follows. Given a set of bridge node lists \( B_t \) for each iteration \( t \), we construct a matrix \( M \) where each row corresponds to an iteration and each column corresponds to a unique bridge node. The matrix entries are:
\[
M_{t,v} =
\begin{cases}
1, & v \in B_t \\
0, & \text{otherwise}
\end{cases}
\]
where \( M_{t,v} \) indicates whether node \( v \) appears as a bridge at iteration \( t \). The full set of unique bridge nodes across all iterations is extracted to define the columns of \( M \).

To identify the earliest appearing bridge nodes we compute the first iteration in which each node appears:
\[
t_{\text{first}}(v) = \min \{ t \mid M_{t,v} = 1 \}
\]
The top 100 earliest appearing bridge nodes are selected by ranking nodes based on \( t_{\text{first}}(v) \), keeping those with the smallest values. The binary matrix is then restricted to these nodes.

To capture early-stage network formation, the analysis is limited to the first 200 iterations, ensuring that the onset of key bridge nodes is clearly visible. The final presence matrix \( M' \) is reordered so that nodes are sorted by their first appearance, emphasizing the sequential nature of bridge formation.

The matrix is visualized as a heatmap (Figure~\ref{fig:sorted_bridge_heatmap}), where rows correspond to the top 100 earliest appearing bridge nodes and columns represent iterations. A blue-scale colormap is used to indicate presence (darker shades for active nodes).  

To analyze the evolution of key bridge nodes in the knowledge graph, we compute and track the betweenness centrality of all nodes across multiple iterations. Betweenness centrality quantifies the importance of a node as an intermediary in shortest paths and is defined as:
\[
C_B(v) = \sum_{s \neq v \neq t} \frac{\sigma_{st}(v)}{\sigma_{st}}
\]
where \( \sigma_{st} \) is the total number of shortest paths between nodes \( s \) and \( t \), and \( \sigma_{st}(v) \) is the number of those paths that pass through \( v \). This measure is recalculated at each iteration to observe structural changes in the network.

The computational procedure is as follows:  
\begin{enumerate}
    \item Graph Loading: Graph snapshots are loaded from GraphML files, indexed by iteration number. If a graph is directed, it is converted to an undirected format using \texttt{networkx.to\_undirected()} to ensure consistent betweenness computations.
    \item Betweenness Centrality Calculation: For each graph \( G_t \) at iteration \( t \), the betweenness centrality for all nodes is computed using \texttt{networkx.betweenness\_centrality()}.
    \item Time Series Construction: The computed centrality values are stored in a time-series matrix \( B \), where rows correspond to iterations and columns correspond to nodes:
     \[
     B_{t,v} = C_B(v) \quad \forall v \in V_t
     \]
     Missing values (nodes absent in certain iterations) are set to zero to maintain a consistent matrix structure.
\end{enumerate}

To identify key bridge nodes, we extract the top ten nodes with the highest peak betweenness at any iteration:
\[
H = \{ v \mid \max_t B_{t,v} \geq B_{\text{top}, 10} \}
\]
where \( B_{\text{top}, 10} \) represents the 10th highest betweenness value recorded across all iterations. The time-series data is filtered to retain only these nodes.

To visualize the dynamic role of key bridge nodes, we generate a line plot of betweenness centrality evolution where each curve represents the changing centrality of a top bridge node over iterations. This graph captures how structural importance fluctuates over time.

\subsection{Agentic Approach to Reason over Longest Shortest Paths}
\label{methods:agentic_longest_shortest}

We employ an agentic approach to analyze structured knowledge representations in the form of a graph \( G = (V, E) \), where \( V \) represents the set of nodes (concepts) and \( E \) represents the set of edges (relationships). The methodology consists of four primary steps: (i) extraction of the longest knowledge path, (ii) decentralized node and relationship reasoning, (iii) multi-agent synthesis, and (iv) structured report generation.

\textbf{Path Extraction.} The input knowledge graph \( G \) is first converted into an undirected graph \( G' = (V, E') \) where \( E' \) contains bidirectional edges to ensure reachability across all nodes. We extract the largest connected component \( G_c \) by computing:
\[
G_c = \arg\max_{S \in \mathcal{C}(G')} |S|
\]
where \( \mathcal{C}(G') \) is the set of all connected components in \( G' \). The longest shortest path, or \textit{diameter path}, is determined by computing the eccentricity:
\[
\epsilon(v) = \max_{u \in V} d(v, u),
\]
where \( d(v, u) \) is the shortest path length between nodes \( v \) and \( u \). The source node is selected as \( v^* = \arg\max_{v \in V} \epsilon(v) \), and the farthest reachable node from \( v^* \) determines the longest path.

Numerically, the longest paths are determined by computing node eccentricities using \texttt{networkx.eccentricity()}, which identifies the most distant node pairs in terms of shortest paths. The five longest shortest paths are extracted with \texttt{networkx.shortest\_path()}. For each extracted path, we assign node-level structural metrics computed from the original graph. The node degree is obtained using \texttt{networkx.degree()}, betweenness centrality is computed with \texttt{networkx.betweenness\_centrality()}, and closeness centrality is determined via \texttt{networkx.closeness\_centrality()}. Each identified path is saved as a GraphML file using \texttt{networkx.write\_graphml()} with these computed node attributes for further analysis.

\textbf{Decentralized Node and Relationship Reasoning.} Each node \( v_i \in V \) and each relationship \( e_{ij} \in E \) along the longest path is analyzed separately. A language model \( f_{\theta} \) is prompted with:
\[
\text{LLM}(v_i) = f_{\theta}(\text{``Analyze concept } v_i \text{ in a novel scientific context."})
\]
for nodes, and
\[
\text{LLM}(e_{ij}) = f_{\theta}(\text{``Analyze relationship } e_{ij} \text{ and hypothesize new implications."})
\]
for relationships. This enables independent hypothesis generation at the atomic level.

\textbf{Multi-Agent Synthesis.} The set of independent insights \( \mathcal{I} = \{ I_1, I_2, \dots \} \) is aggregated, and a final inference step is performed using:
\[
I_{\text{final}} = f_{\theta}(\text{``Synthesize a novel discovery from } \mathcal{I} \text{."})
\]
This allows the model to infer higher-order patterns beyond individual node-relationship reasoning.

\textbf{Structured Report Generation.} The final response, along with intermediate insights, is formatted into a structured markdown report containing:
\begin{itemize}
    \item The extracted longest path
    \item Individual insights per node and relationship
    \item The final synthesized discovery
\end{itemize}
This approach leverages multi-step reasoning and recursive inference, allowing for emergent discoveries beyond explicit graph-encoded knowledge.

\subsubsection{Agent-driven Compositional Reasoning}
\label{methods:agentic_compositional}
We employ a multi-step agentic approach that couples LLMs with graph-based compositional reasoning. To develop such an approach, we load the graph and locate its largest connected component. We compute eccentricities to identify two far-apart nodes, then extract the longest shortest path between them. Each node in that path becomes a ``building block,'' for which the LLM provides a concise definition, principles, and a property conducive to synergy (Step A). Next, we prompt the LLM to create pairwise synergies by merging adjacent building blocks, encouraging a short, compositional statement that unifies the nodes' respective features (Step B). To deepen the layering of ideas, we consolidate multiple synergy statements into bridge synergies that capture cross-cutting themes (Step C). Finally, we issue a more elaborate prompt asking the LLM to integrate all building blocks and synergies into an expanded, coherent ``final discovery,'' referencing both prior statements and each node's defining traits (Step D). This process yields a multi-step compositional approach, wherein each synergy can build on earlier results to reveal increasingly sophisticated connections. The initial steps A-C are carried out using \url{meta-llama/Llama-3.2-3B-Instruct}, whereas the final integration of the response in Step D is conducted using \url{meta-llama/Llama-3.3-70B-Instruct}. We also experimented with other models, such as \texttt{o1-pro} as discussed in the main text.

\subsection{Scale free analysis}

To determine whether a given network exhibits scale-free properties, we analyze its degree distribution using the power-law fitting method implemented in the \texttt{powerlaw} Python package. The algorithm extracts the degree sequence from the input graph and fits a power-law distribution, estimating the exponent \(\alpha\) and lower bound \(x_{\min}\). To assess whether the power-law is a preferable fit, we compute the log-likelihood ratio (LR) between the power-law and an exponential distribution, along with the corresponding $p$-value. A network is classified as scale-free if LR is positive and \(p < 0.05\), indicating statistical support for the power-law hypothesis. The method accounts for discrete degree values and excludes zero-degree nodes from the fitting process.

\subsection{Audio Summary in the Form of a Podcast}
Supplementary Audio A1 presents an audio summary of this paper in the style of a podcast, created using PDF2Audio (\url{https://huggingface.co/spaces/lamm-mit/PDF2Audio}~\cite{ghafarollahi2024sciagentsautomatingscientificdiscovery}). The audio format in the form a conversation enables reader to gain a broader understanding of the results of this paper, including expanding the broader impact of the work. The transcript was generated using the \texttt{o3-mini} model~\cite{o3-mini-model-card-2025} from the final draft of the paper.

\section*{Code, data and model weights availability}

Codes, model weights and additional materials are available at \url{https://huggingface.co/lamm-mit} and \url{https://github.com/lamm-mit/PRefLexOR}. The model used for the experiments is available at~\url{lamm-mit/Graph-Preflexor_01062025}. 

\section*{Conflicts of Interest}

The author declares no conflicts of interest of any kind. 

\section*{Acknowledgments}

The author acknowledges support from the MIT Generative AI initiative. 

\bibliographystyle{naturemag}

\bibliography{references,references-Mendeley}

\clearpage
\appendix
\renewcommand{\thesection}{S\arabic{section}}
\renewcommand{\thefigure}{S\arabic{figure}}
\renewcommand{\thetable}{S\arabic{table}}
\renewcommand{\thetextbox}{S\arabic{textbox}}

\setcounter{section}{0}
\setcounter{figure}{0}
\setcounter{table}{0}
\setcounter{textbox}{0}

\newpage

\section*{\vspace{1cm} \centering \Large  \sffamily \\ \textbf{Supplementary Information}}
\vspace{4cm}

\begin{center}
    {\LARGE\sffamily  \textbf{Agentic Deep Graph Reasoning Yields Self-Organizing Knowledge Networks }} \\[1em] 
    
\end{center}

\vspace{3cm}

\begin{center}
    {\large Markus J. Buehler} \\[1.em]  

    Laboratory for Atomistic and Molecular Mechanics\\Center for Computational Science and Engineering \\
        Schwarzman College of Computing \\
	Massachusetts Institute of Technology\\
	Cambridge, MA 02139, USA \\
[0.5em]
    
    {\normalsize mbuehler@MIT.EDU}
\end{center}

\clearpage

\begin{figure}[h!]
    \centering
    \includegraphics[width=1\textwidth]{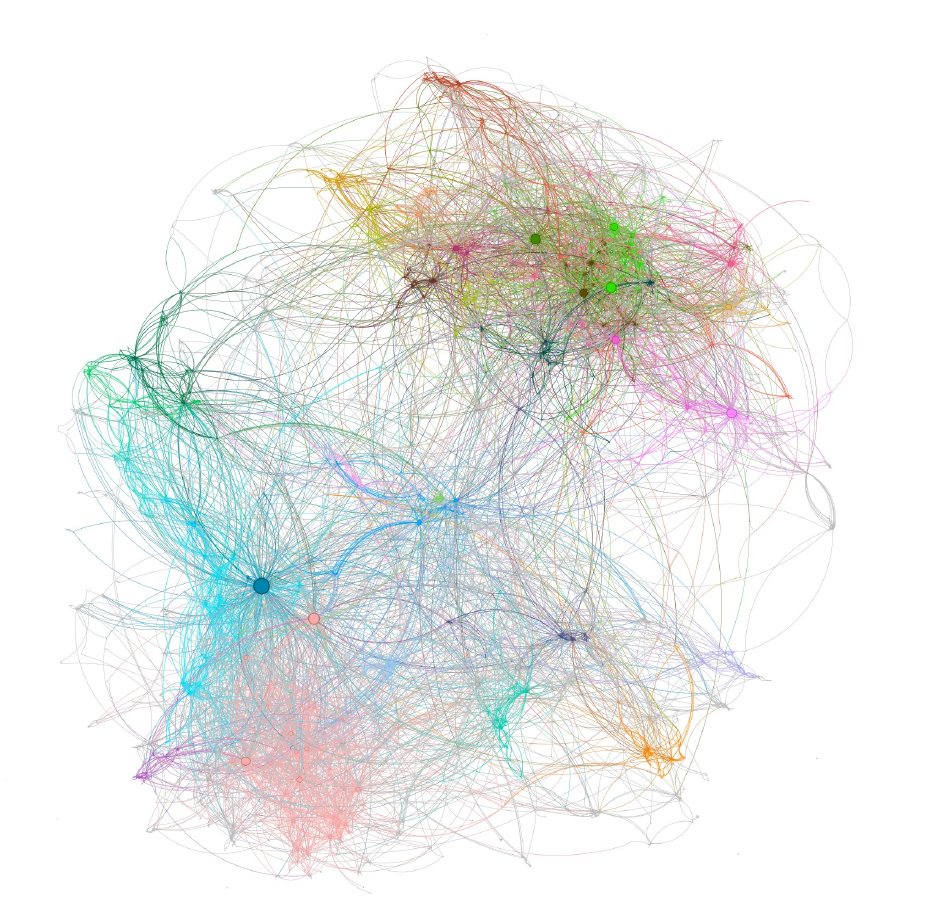}
    \caption{Knowledge graph $\mathcal{G_1}$ after around 1,000 iterations, under a flexible self-exploration scheme initiated with the prompt \texttt{Discuss an interesting idea in bio-inspired materials science.}. In this visualization, nodes/edges are colored according to cluster ID.}
    \label{fig:fig_2000_C}
\end{figure}

\begin{figure}[h!]
    \centering
    \includegraphics[width=1\textwidth]{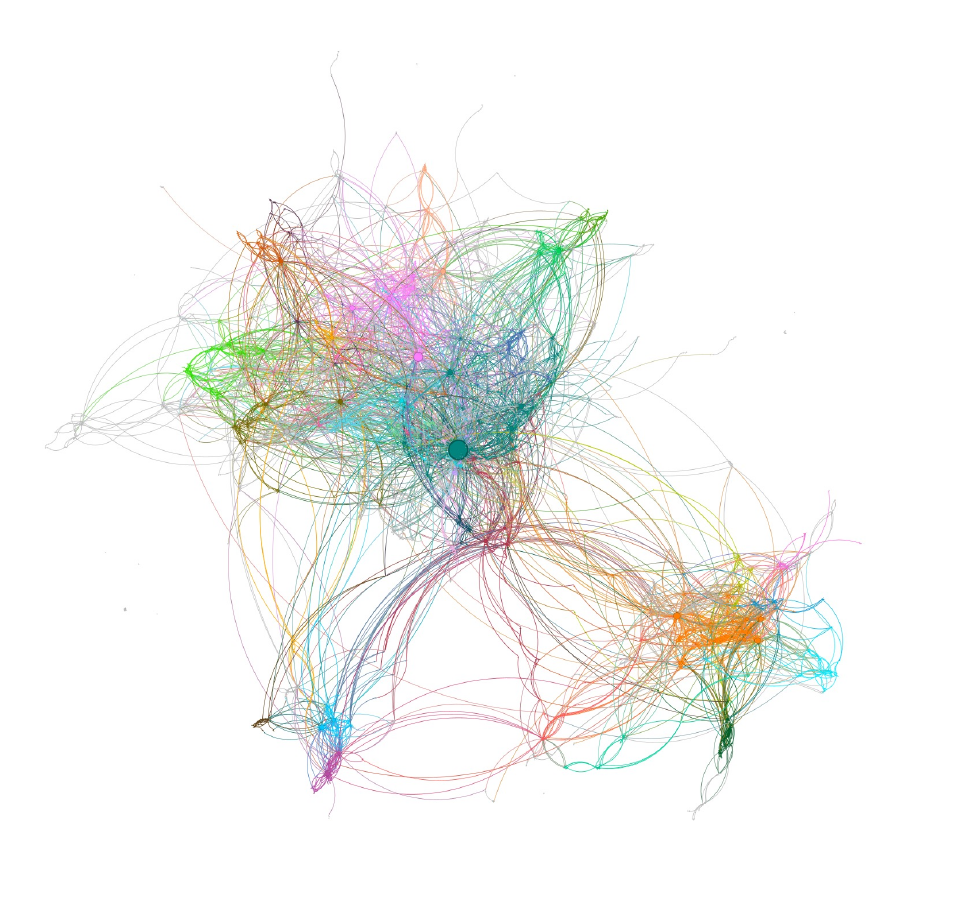}
    \caption{Knowledge graph $\mathcal{G_2}$ after around 500 iterations, under a topic-specific self-exploration scheme initiated with the prompt \texttt{Describe a way to design impact resistant materials}. Nodes/edges are colored according to cluster ID.} 
    \label{fig:fig_2001_C}
\end{figure}

\clearpage

\begin{figure}[h]
    \centering
    \includegraphics[width=0.8\textwidth]{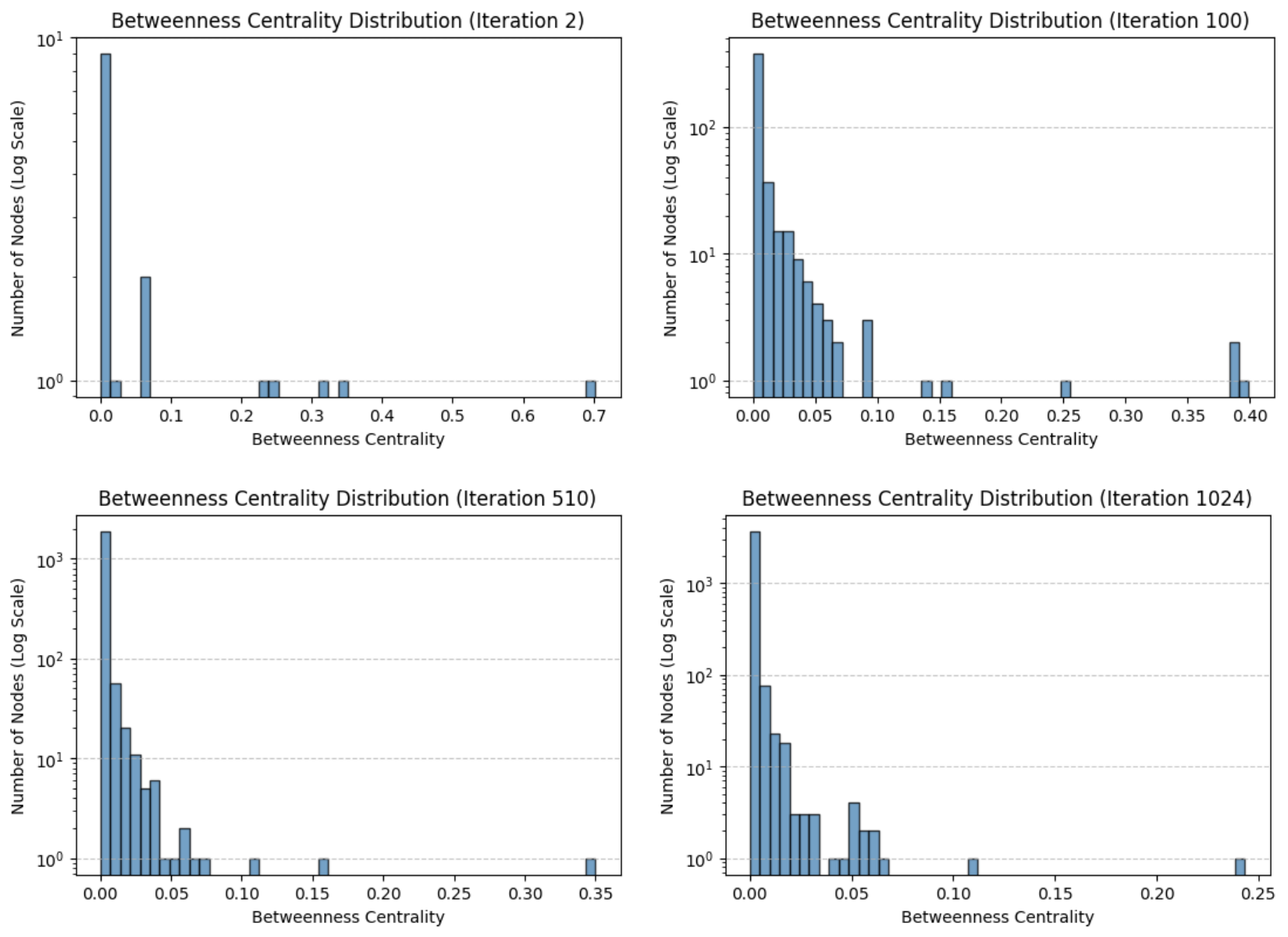}
    \caption{Distribution of betweenness centrality across four iterations, $\mathcal{G_1}$. The $y$-axis is in log scale, indicating the number of nodes with a given centrality value. The evolution suggests a transition from an early centralized state to a more distributed knowledge structure in later iterations.}
    \label{fig:betweenness_distribution}
\end{figure}

\clearpage

\begin{table}[h!]
\centering
\begin{tabular}{|p{3cm}|p{6cm}|p{6cm}|}
\hline
\textbf{Metric} & \textbf{Response 1 (reasoning wit graph data)} & \textbf{Response 2 (reasoning without graph data)} \\ \hline
Graph Utilization & 5/5 (Explicit use of graph-based insights for material selection and optimization) & 0/5 (No reference to graph data) \\ \hline
Depth of Reasoning & 4/5 (Multi-step reasoning including AI/ML integration, degradation adaptability, and bioinspiration) & 3/5 (Structured approach but mostly standard materials science reasoning) \\ \hline
Scientific Rigor & 4/5 (Logical relationships between material properties, AI/ML, and impact resistance) & 4/5 (Follows well-established materials science principles with clear reasoning) \\ \hline
Innovativeness & 5/5 (Proposes unique ideas such as AI/ML-driven discovery, urban ecosystems, and adaptive materials) & 3/5 (Uses conventional material design strategies without novel insights) \\ \hline
\textbf{Overall Score} & \textbf{18/20} (Superior interdisciplinary and computational approach) & \textbf{10/20} (Limited to conventional material design strategies) \\ \hline
\end{tabular}
\caption{Comparison of Responses on Impact-Resistant Material Design with Annotated Scores. Response 1 achieves a higher score due to its interdisciplinary approach, explicit graph-based insights, and innovative integration of AI/ML.}
\label{tab:response_comparison}
\end{table}

\clearpage
 
\textboxcaption{Answer of the graph-reasoning model to the task \texttt{Describe a way to design impact resistant materials.} In this case, we extract key insights from a knowledge graph using centrality measures, Louvain community detection, and key relationship identification, integrating structured graph-based context into LLM prompts to enhance reasoning and evaluate response quality.}
\label{TexBox:graph-response}

\begin{LLMbox}{Answer of the graph-reasoning model}
\begin{lstlisting}
Based on the knowledge graph insights, I will outline a design strategy for impact-resistant materials.

**Design Strategy:**

1. **Material Selection and Optimization**: Utilize advanced materials science and machine learning algorithms to identify and optimize materials with inherent impact resistance properties. This can be achieved by analyzing the relationships between materials, their properties, and impact resistance.

2. **Self-Healing Materials**: Incorporate self-healing materials into the design, which can repair cracks and damages autonomously. This can be achieved through the use of microcapsules, polymers, or other materials that can release healing agents in response to impact.

3. **Adaptive Materials**: Design materials that can adapt to changing environmental conditions, such as temperature, humidity, or pressure. This can be achieved through the use of shape-memory alloys, polymers, or other materials that can change their properties in response to external stimuli.

4. **Urban Ecosystems Integration**: Incorporate urban ecosystems into the design, which can provide a natural source of inspiration for impact-resistant materials. For example, studying the properties of urban algae farms or the self-healing properties of certain plants can inform the design of impact-resistant materials.

5. **Artificial Intelligence (AI) and Machine Learning (ML) Integration**: Utilize AI and ML algorithms to analyze data from various sources, such as sensor data, material properties, and environmental conditions. This can help identify patterns and trends that can inform the design of impact-resistant materials.

6. **Material Degradation and Adaptability**: Design materials that can adapt to material degradation over time. This can be achieved through the use of materials that can self-heal, self-repair, or self-renew.

**Key Materials and Technologies:**

1. **Shape-Memory Alloys (SMAs)**: SMAs can be used to create adaptive materials that can change their properties in response to external stimuli.
2. **Polymers**: Polymers can be used to create self-healing materials that can repair cracks and damages autonomously.
3. **Graphene**: Graphene can be used to create ultra-strong and lightweight materials with high impact resistance.
4. **Nanomaterials**: Nanomaterials can be used to create materials with enhanced impact resistance and self-healing properties.

**Design Considerations:**

1. **Material Properties**: Consider the material properties, such as strength, stiffness, and toughness, when designing impact-resistant materials.
2. **Environmental Conditions**: Consider the environmental conditions, such as temperature, humidity, and pressure, when designing impact-resistant materials.
3. **Scalability**: Consider the scalability of the design, including the ability to manufacture and deploy the material at large scales.
4. **Cost-Effectiveness**: Consider the cost-effectiveness of the design, including the cost of materials, manufacturing, and deployment.

By following this design strategy, impact-resistant materials can be designed that are optimized for performance, scalability, and cost-effectiveness.
\end{lstlisting}
\end{LLMbox}

\clearpage
 
\textboxcaption{Answer of the base model to the task \texttt{Describe a way to design impact resistant materials.}}
\label{TexBox:base-response}

\begin{LLMbox}{Answer of the base model}
\begin{lstlisting}
Designing impact-resistant materials involves creating materials that can absorb and distribute the energy of an impact, reducing the likelihood of damage or failure. Here's a step-by-step approach to designing impact-resistant materials:

1. **Material Selection**: Choose a material with inherent properties that can absorb and dissipate energy, such as:
	* High-density materials (e.g., metals, ceramics)
	* Viscoelastic materials (e.g., rubber, polymers)
	* Energy-absorbing materials (e.g., foam, composites)
2. **Microstructure Optimization**: Design the material's microstructure to optimize energy absorption:
	* Create a hierarchical structure with multiple scales (e.g., nanoparticles, microparticles, macroscopic particles)
	* Incorporate defects or irregularities to increase the material's energy-absorbing capacity
	* Use phase transformations or phase changes to absorb energy
3. **Material Properties Engineering**: Engineer the material's properties to enhance its impact resistance:
	* Increase the material's stiffness and strength to resist deformation
	* Improve the material's toughness by increasing its energy-absorbing capacity
	* Enhance the material's viscoelastic properties to absorb and dissipate energy
4. **Composite Materials**: Design composite materials that combine different materials to achieve enhanced impact resistance:
	* Use a matrix material (e.g., polymer, metal) to bind together reinforcement materials (e.g., fibers, particles)
	* Incorporate energy-absorbing materials (e.g., foam, rubber) into the composite
	* Use a hierarchical structure to optimize energy absorption
5. **Testing and Validation**: Test and validate the material's impact resistance using standardized tests (e.g., Charpy impact test, Izod impact test):
	* Measure the material's energy absorption capacity
	* Evaluate the material's ability to withstand different types of impacts (e.g., drop, impact, penetration)
6. **Scaling and Optimization**: Scale up the material design and optimize it for specific applications:
	* Use computational models (e.g., finite element analysis, molecular dynamics) to simulate the material's behavior under impact
	* Optimize the material's design for specific applications (e.g., aerospace, automotive, sports equipment)

Some examples of impact-resistant materials include:

* Kevlar-reinforced polymers
* Carbon fiber-reinforced polymers
* High-density polyethylene (HDPE) foam
* Rubber-based composites
* Ceramic matrix composites

By following these steps, designers can create impact-resistant materials that can withstand various types of impacts and reduce the risk of damage or failure.
\end{lstlisting}
\end{LLMbox}

\clearpage
 
\begin{figure}[h]
    \centering
    \includegraphics[width=.9\textwidth]{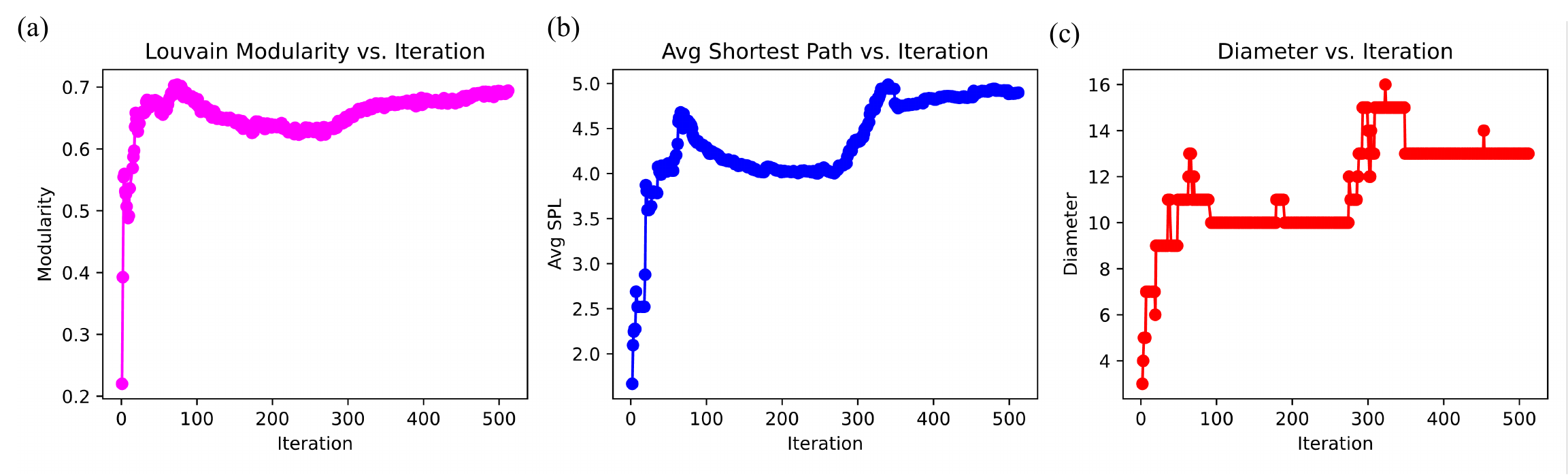}
    \caption{Evolution of key structural properties in the recursively generated knowledge graph ($\mathcal{G_2}$, focused on \texttt{Describe a way to design impact resistant materials.}): (a) Louvain modularity, showing stable community formation; (b) average shortest path length, highlighting efficient information propagation; and (c) graph diameter, demonstrating bounded hierarchical expansion.}
    \label{fig:graph_modularity_V7}
\end{figure}

\clearpage

\begin{figure}[h]
    \centering
    \includegraphics[width=.9\textwidth]{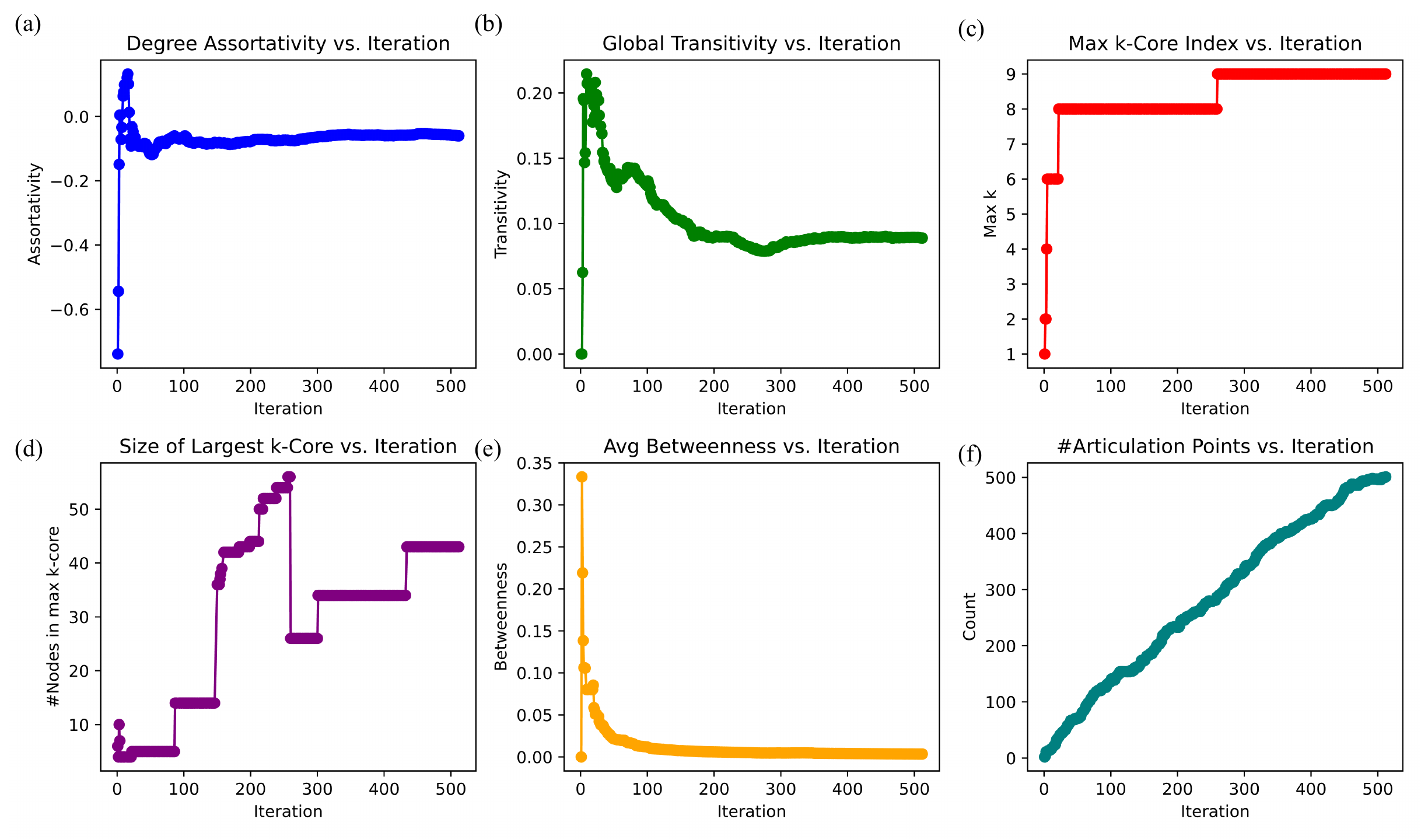}
    \caption{Evolution of graph properties over recursive iterations, highlighting the emergence of hierarchical structure, hub formation, and adaptive connectivity (Graph $\mathcal{G_2}$, focused on \texttt{Describe a way to design impact resistant materials.}).}
    \label{fig:graph_analysis_V7}
\end{figure}

\clearpage
\hypertarget{supporting_text_1}{\includepdf[pages=-]{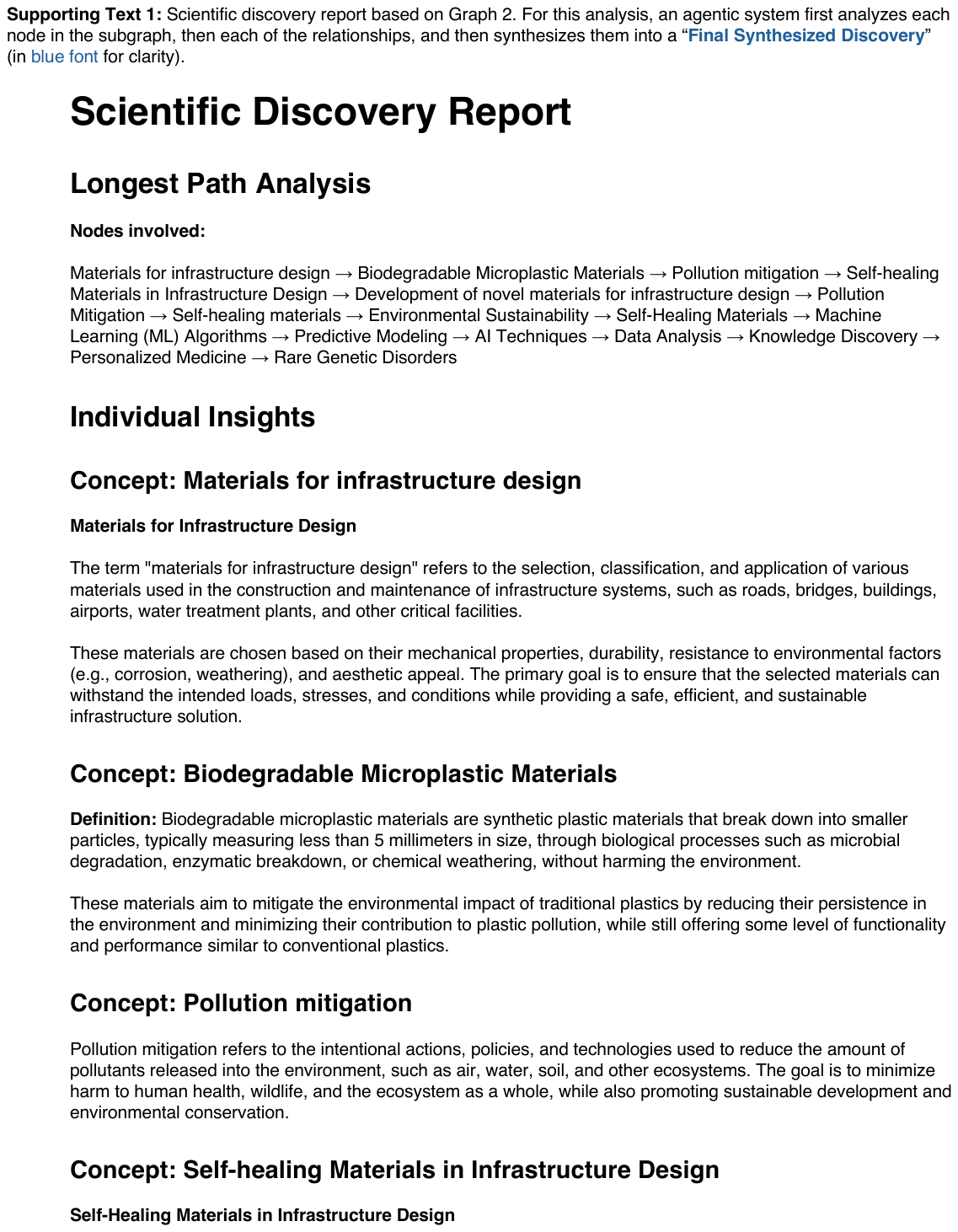}}

\clearpage
\hypertarget{supporting_text_2}{\includepdf[pages=-]{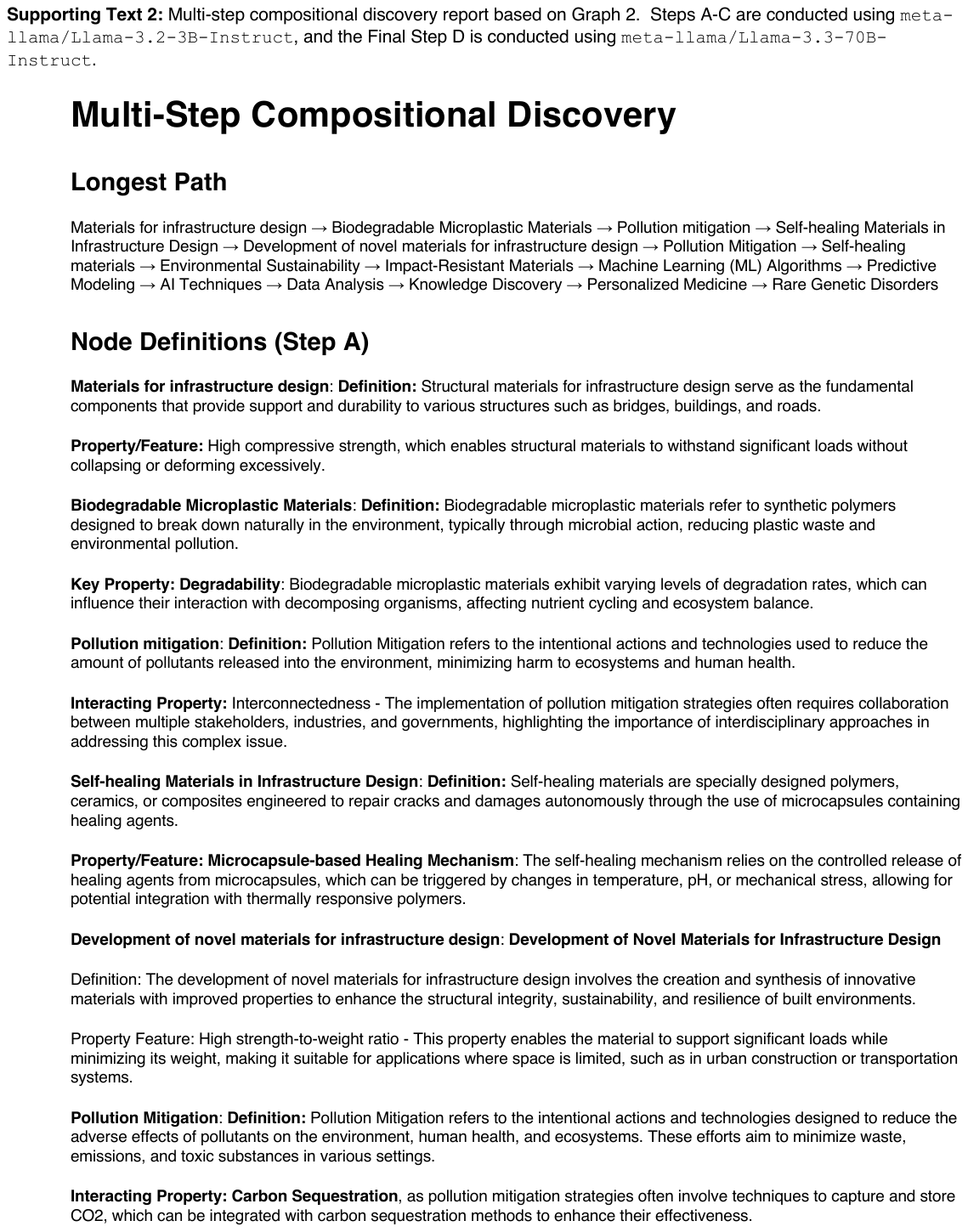}}

\clearpage
\hypertarget{supporting_text_3}{\includepdf[pages=-]{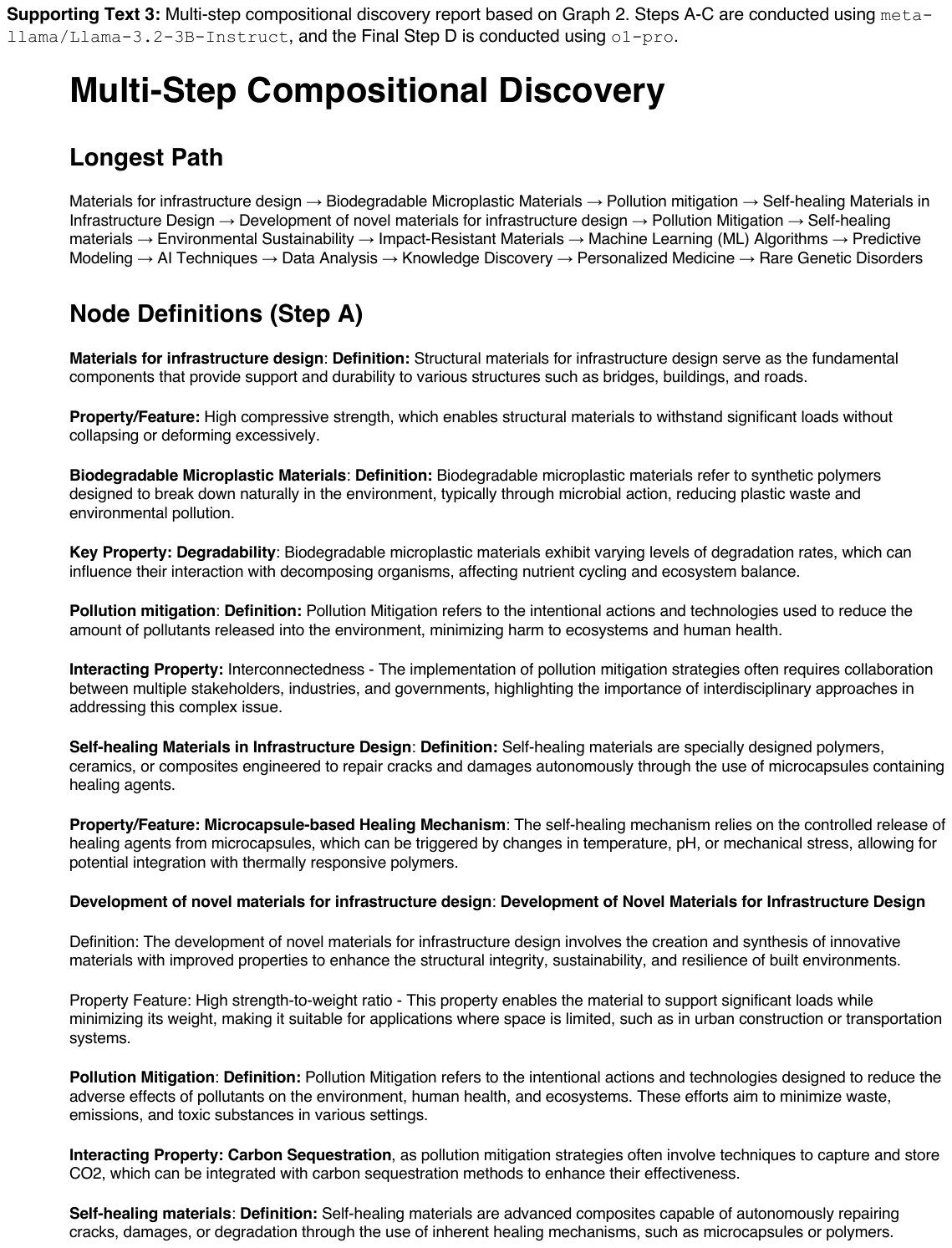}}

\clearpage
\hypertarget{supporting_text_4}{

\section*{Supplementary material}

**Supporting Text 1:** Scientific discovery report based on Graph 2. For this analysis, an agentic system first analyzes each node in the subgraph, then each of the relationships, and then synthesizes them into a "**Final Synthesized Discovery**" (in blue font for clarity).

# Scientific Discovery Report

## Longest Path Analysis

**Nodes involved:**

Materials for infrastructure design → Biodegradable Microplastic Materials → Pollution mitigation → Self-healing Materials in Infrastructure Design → Development of novel materials for infrastructure design → Pollution Mitigation → Self-healing materials → Environmental Sustainability → Self-Healing Materials → Machine Learning (ML) Algorithms → Predictive Modeling → AI Techniques → Data Analysis → Knowledge Discovery → Personalized Medicine → Rare Genetic Disorders

## Individual Insights

### Concept: Materials for infrastructure design

**Materials for Infrastructure Design**

The term "materials for infrastructure design" refers to the selection, classification, and application of various materials used in the construction and maintenance of infrastructure systems, such as roads, bridges, buildings, airports, water treatment plants, and other critical facilities.

These materials are chosen based on their mechanical properties, durability, resistance to environmental factors (e.g., corrosion, weathering), and aesthetic appeal. The primary goal is to ensure that the selected materials can withstand the intended loads, stresses, and conditions while providing a safe, efficient, and sustainable infrastructure solution.

### Concept: Biodegradable Microplastic Materials

**Definition:** Biodegradable microplastic materials are synthetic plastic materials that break down into smaller particles, typically measuring less than 5 millimeters in size, through biological processes such as microbial degradation, enzymatic breakdown, or chemical weathering, without harming the environment.

These materials aim to mitigate the environmental impact of traditional plastics by reducing their persistence in the environment and minimizing their contribution to plastic pollution, while still offering some level of functionality and performance similar to conventional plastics.

### Concept: Pollution mitigation

Pollution mitigation refers to the intentional actions, policies, and technologies used to reduce the amount of pollutants released into the environment, such as air, water, soil, and other ecosystems. The goal is to minimize harm to human health, wildlife, and the ecosystem as a whole, while also promoting sustainable development and environmental conservation.

### Concept: Self-healing Materials in Infrastructure Design

**Self-Healing Materials in Infrastructure Design**

Self-healing materials are advanced materials that can repair cracks, damages, or degradation over time due to their inherent ability to heal themselves through various mechanisms such as:

1. Autocatalytic reactions (e.g., polymerization)
2. Microbial activity (e.g., bacteria-induced calcification)
3. Thermally activated transformations (e.g., shape memory alloys)

These self-healing materials have the potential to significantly improve infrastructure design by reducing maintenance costs, increasing durability, and enhancing overall system reliability.

Examples of applications include:

- Concrete and asphalt for road construction
- Polymer-based composites for bridges and buildings
- Fiber-reinforced polymers (FRPs) for offshore wind turbines and pipelines

By integrating self-healing materials into infrastructure design, we can create more resilient and sustainable structures that can adapt to environmental stresses and minimize downtime.

## Concept: Development of novel materials for infrastructure design

**Definition:** The development of novel materials for infrastructure design refers to the creation and application of innovative, engineered materials that can be used to construct, maintain, and upgrade various types of infrastructure, such as buildings, bridges, roads, and other critical public facilities.

These novel materials often exhibit enhanced properties compared to traditional materials, including improved strength, durability, sustainability, and resistance to environmental degradation, corrosion, or extreme temperatures. The goal is to improve the performance, efficiency, and lifespan of infrastructure while reducing costs, minimizing environmental impact, and enhancing overall safety and resilience.

## Concept: Pollution Mitigation

**Pollution Mitigation** refers to the practices, strategies, and technologies aimed at reducing or eliminating pollution in various forms, such as air, water, soil, noise, and waste. The primary goal of pollution mitigation is to minimize the negative impacts of human activities on the environment, public health, and ecosystems, ultimately preserving natural resources for future generations.

## Concept: Self-healing materials

**Self-healing materials** are intelligent materials that have the ability to repair themselves automatically after damage, injury, or degradation. These materials can recover from cracks, tears, or other defects through various mechanisms, such as:

1. Chemical reactions (e.g., polymerization)
2. Phase changes (e.g., solidification of liquid)
3. Mechanical self-repair (e.g., fiber alignment)

This property is achieved through the incorporation of advanced technologies, such as microcapsules, shape-memory alloys, or bio-inspired systems, which allow the material to sense, detect, and respond to damage by initiating a healing process.

Self-healing materials have numerous potential applications in fields like construction, aerospace, biomedical engineering, and more, where durability, reliability, and reduced maintenance costs are crucial.

## Concept: Environmental Sustainability

**Definition:** Environmental sustainability refers to the ability of the natural environment to support and maintain the health, diversity, and resilience of ecosystems, while also meeting the needs of present and future generations. It encompasses the long-term management and conservation of natural resources, such as water, air, soil, and biodiversity, to ensure a stable and healthy planet for all living beings.

In simpler terms, environmental sustainability is about balancing human activities with the need to preserve the natural world, ensuring that our actions do not harm the environment, but rather work in harmony with it to create a thriving ecosystem that supports life on Earth.

## Concept: Self-Healing Materials

**Definition:** Self-healing materials are intelligent materials that can repair or restore their original properties and functionality after damage, deformation, or degradation due to external factors such as mechanical stress, environmental exposure, or chemical reactions.

These materials have been designed with integrated self-healing mechanisms, often incorporating microcapsules, polymers, or other components that release healing agents when triggered by specific stimuli, allowing them to recover from damages autonomously.

## Concept: Machine Learning (ML) Algorithms

**Machine Learning (ML) Algorithms**

Machine learning algorithms are a set of mathematical techniques used to enable computers to learn from data, make predictions, and improve their performance on a task without being explicitly programmed.

In essence, machine learning algorithms allow systems to identify patterns, relationships, and trends in data, and use this information to make decisions or predictions that are not based on explicit rules or programming. This enables machines to learn from experience, adapt to new situations, and improve their performance over time.

## Concept: Predictive Modeling

Predictive modeling is a statistical technique used to forecast future events, behaviors, or outcomes based on historical data and patterns learned from existing information. It involves training machine learning algorithms on large datasets to identify relationships and correlations between variables, allowing for the development of predictive models that can make accurate predictions about future scenarios. These models are then used in various fields such as business, finance, healthcare, and climate science to inform decision-making and optimize performance.

## Concept: AI Techniques

**AI Techniques**

Artificial Intelligence (AI) techniques refer to a set of algorithms, methods, and approaches used to create intelligent machines that can perform tasks that typically require human intelligence, such as:

- Learning from data
- Reasoning and problem-solving
- Perception and understanding natural language
- Decision-making and prediction

These techniques enable AI systems to simulate human-like intelligence, automate complex tasks, and improve decision-making processes across various industries, including healthcare, finance, transportation, and more.

Some common AI techniques include Machine Learning, Deep Learning, Natural Language Processing, Computer Vision, and Rule-Based Systems.

## Concept: Data Analysis

**Data Analysis** refers to the process of examining, interpreting, and summarizing data to extract meaningful insights, patterns, and trends. It involves using various statistical techniques, algorithms, and tools to identify relationships between variables, detect anomalies, and make informed decisions based on the results. The goal of data analysis is to transform raw data into actionable knowledge that can inform business strategies, solve problems, or answer research questions.

## Concept: Knowledge Discovery

**Knowledge Discovery**: Knowledge discovery refers to the process of identifying, extracting, and utilizing relevant information from various sources, including data, literature, and expert opinions, to gain insights, patterns, or relationships that were previously unknown or not apparent. This process involves using techniques such as data mining, machine learning, and statistical analysis to uncover hidden knowledge and make informed decisions in fields like science, business, and healthcare.

## Concept: Personalized Medicine

**Personalized Medicine**: A medical approach that tailors treatment, prevention, and diagnosis to an individual's unique genetic profile, lifestyle, and medical history. It involves using advanced technologies, such as genomics, proteomics, and epigenomics, to develop targeted therapies and treatments that are more effective and have fewer side effects compared to traditional one-size-fits-all approaches.

## Concept: Rare Genetic Disorders

**Rare Genetic Disorders (RGDs)** refer to a group of genetic conditions that affect a small percentage of the population, typically affecting less than 1 in 2,000 people worldwide. These disorders are caused by mutations or variations in specific genes, leading to abnormal protein function or structure, which in turn disrupt normal biological processes.

RGDs can be inherited from one's parents (autosomal dominant or recessive patterns) or occur spontaneously due to random errors during DNA replication. Examples of rare genetic disorders include cystic fibrosis, sickle cell anemia, and Huntington's disease. The rarity and variability of RGDs often make diagnosis and treatment challenging, but research into these conditions has led to significant advances in understanding human genetics and developing new therapeutic approaches.

## Relationship: Materials for infrastructure design → Biodegradable Microplastic Materials (IS-A)

Deeper implications:

- Sustainable infrastructure design could leverage biodegradable microplastics to create self-healing or adaptive materials that minimize waste and environmental impact.
- Research on biodegradable microplastics may inform the development of more durable and long-lasting materials for infrastructure design.

Hypothesis: Biodegradable microplastic-based composites could exhibit improved crack resistance and fatigue life when integrated into structural materials used in infrastructure design.

General principle: "Functional materials can learn from each other's limitations and synergies to drive innovation and reduce environmental footprints."

## Relationship: Biodegradable Microplastic Materials → Pollution mitigation (RELATES-TO)

Deeper implications:

1. Biodegradable microplastic materials may serve as an intermediate step towards more effective pollution mitigation strategies.
2. A broader understanding of biodegradable materials could lead to the development of more efficient decomposition pathways for other types of pollutants.

Hypothesis:

- Microorganisms from different taxonomic groups (e.g., bacteria, fungi) have evolved unique enzymes that efficiently degrade specific types of polymers found in biodegradable microplastics, potentially leading to novel applications in bioremediation.

General principle:

- "Interdisciplinary convergence" - Collaboration between scientists from diverse fields (materials science, microbiology, ecology) can accelerate the discovery of innovative solutions for pollution mitigation and sustainability challenges.

## Relationship: Pollution mitigation → Self-healing Materials in Infrastructure Design (RELATES-TO)

Deeper Implication: The relationship between pollution mitigation and self-healing materials suggests that using eco-friendly materials in infrastructure design can lead to reduced long-term maintenance needs, resulting in lower operational emissions.

Hypothesis: Microorganisms embedded in self-healing materials can degrade organic pollutants at a faster rate than traditional remediation methods.

General Principle: Nature-inspired materials with adaptive properties can enhance the resilience of infrastructure systems against both external damage and internal pollution, leading to improved sustainability outcomes.

## Relationship: Self-healing Materials in Infrastructure Design → Development of novel materials for infrastructure design (INFLUENCES)

**Deeper Implications:** The integration of self-healing materials into infrastructure design has the potential to disrupt the entire materials science industry, driving innovation in material development and paving the way for widespread adoption of novel materials across various industries.

**Hypothesis:** A class of microorganism-inspired biomimetic materials can be developed that can not only self-heal but also adapt to changing environmental conditions, leading to breakthroughs in fields like aerospace, biomedical engineering, and energy storage.

**General Principle:** "Nature's secrets hold the key to human innovation," suggesting that studying and emulating nature's self-healing processes can lead to groundbreaking discoveries in various fields, driving technological advancements and transforming industries.

## Relationship: Development of novel materials for infrastructure design → Pollution Mitigation (INFLUENCES)

The deeper implications of the relationship between Development of novel materials for infrastructure design and Pollution Mitigation include:

- Improved material properties can reduce environmental degradation and pollution by minimizing the need for frequent replacements, repairs, or disposal of materials.
- Novel materials with enhanced durability and sustainability can extend the lifespan of infrastructure, reducing the amount of waste generated during construction and maintenance.
- New materials and technologies can also facilitate more efficient use of existing infrastructure, leading to reduced energy consumption and lower greenhouse gas emissions.

Hypothesis: Development of self-healing materials for infrastructure could mitigate pollution by reducing the need for frequent repairs and replacements, potentially extending the lifespan of infrastructure and reducing waste.

General Principle: "Sustainable material innovations can serve as a catalyst for broader environmental benefits, driving systemic changes in how we design, build, and maintain infrastructure."

## Relationship: Pollution Mitigation → Self-healing materials (RELATES-TO)

Deeper Implications: The relationship between Pollution Mitigation and Self-healing materials suggests that developing smart materials with self-healing properties can provide innovative solutions for environmental sustainability and resource conservation.

Hypothesis: "Microcapsule-based self-healing materials can be engineered to absorb and break down pollutants, enabling efficient cleanup and restoration of contaminated sites."

General Principle: "Adaptive materials can be designed to interact with their environment, mitigating harm while promoting ecosystem balance and resilience."

## Relationship: Self-healing materials → Environmental Sustainability (RELATES-TO)

**Scientific Relationship:** RELATES-TO

The relationship between Self-healing materials and Environmental Sustainability suggests that developing sustainable materials can contribute to reducing waste, conserving resources, and minimizing environmental impact.

**Deeper Implications:** A symbiotic relationship exists where innovative self-healing materials can help mitigate the consequences of human activities, promoting eco-friendly design, and enabling the development of sustainable infrastructure.

**New Discovery Hypothesis:** Microcapsules used in self-healing materials could be repurposed as pollution-absorbing capsules, capable of removing toxins from contaminated sites, enhancing environmental remediation.

**General Principle:** "Sustainable Innovation" - designing products and systems that integrate self-healing properties, circular economies, and environmental considerations to minimize ecological footprint.

## Relationship: Environmental Sustainability → Self-Healing Materials (IS-A)

Deeper Implications:

1. **Adaptive Infrastructure**: Self-healing materials can be integrated into sustainable infrastructure, enabling buildings, bridges, and other structures to adapt to environmental stresses and changes, reducing maintenance costs and increasing resilience.
2. **Closed-Loop Systems**: The development of self-healing materials can lead to closed-loop systems where damaged materials are repaired and reused, minimizing waste and promoting recycling.

New Discovery Hypothesis: "Microbial-inspired self-healing materials can enhance environmental sustainability by integrating bioremediation capabilities, facilitating the cleanup of pollutants and restoration of degraded ecosystems."

General Principle: "Integrating adaptive technologies into sustainable systems can foster resilience and efficiency, enabling humans to live in harmony with the environment while minimizing ecological footprints."

## Relationship: Self-Healing Materials → Machine Learning (ML) Algorithms (INFLUENCES)

Deeper Implications:

- The integration of self-healing materials with machine learning algorithms may enable adaptive, autonomous, and real-time responses to changing environments.
- This connection has far-reaching implications for fields like construction, healthcare, and robotics, where self-healing materials and AI-powered decision-making converge.

Hypothesis: "Self-healing materials integrated with machine learning algorithms can develop adaptive strategies to optimize resource utilization and minimize waste in complex systems."

General Principle: "A symbiotic convergence of artificial intelligence and adaptive materials can foster autonomous, resilient, and optimized systems that can self-improve through continuous feedback loops."

## Relationship: Machine Learning (ML) Algorithms → Predictive Modeling (RELATES-TO)

Deeper Implications:

The relationship between Machine Learning Algorithms and Predictive Modeling has significant implications for various fields, including business, healthcare, and climate science. It suggests that by leveraging ML algorithms, we can develop more accurate and reliable predictive models that can inform decision-making and drive positive outcomes.

Hypothesis:

"Machine Learning-based predictive models can be improved through the incorporation of multi-scale analysis, integrating data from diverse sources, including environmental, social, and economic factors, to create more comprehensive and accurate forecasts."

General Principle:

"Interdisciplinary approaches combining machine learning, data analysis, and domain expertise can lead to more robust and resilient predictive models, capable of addressing complex real-world problems."

## Relationship: Predictive Modeling → AI Techniques (RELATES-TO)

**Deeper Implications:** The relationship between Predictive Modeling and AI Techniques has far-reaching implications in various fields, enabling data-driven decision-making, automation of complex tasks, and improved forecasting capabilities.

**Hypothesis:** A novel application of Predictive Modeling using AI Techniques could lead to the development of autonomous systems capable of self-learning and adapting to dynamic environments, revolutionizing fields like climate science, logistics, and resource management.

**General Principle:** "Adaptive forecasting through AI-facilitated predictive modeling enables more resilient and responsive decision-making in complex systems."

## Relationship: AI Techniques → Data Analysis (RELATES-TO)

Deeper Implications:

The relationship between AI Techniques and Data Analysis has significant implications for optimizing AI system performance. Effective data analysis enables AI models to learn from high-quality data, making them more accurate and efficient in their decision-making processes.

Hypothesis: A new discovery could involve developing more advanced data preprocessing techniques that can automatically identify and handle noisy or biased data sources, leading to improved AI model performance and reliability.

General Principle: "High-quality data analysis is essential for refining AI models and achieving optimal performance."

## Relationship: Data Analysis → Knowledge Discovery (IS-A)

**Relationship Implications:**

This "IS-A" relationship implies that Data Analysis is a subset of Knowledge Discovery, where Data Analysis is a method used to aid in the Knowledge Discovery process.

**New Hypothesis:**

The application of advanced data analysis techniques (e.g., AI-driven pattern recognition) in knowledge discovery efforts may lead to an exponential increase in the rate of new discoveries, revolutionizing various fields by revealing hidden connections and relationships.

**General Principle:**

"Effective Data Analysis enables more efficient Knowledge Discovery, leading to accelerated progress in complex domains."

## Relationship: Knowledge Discovery → Personalized Medicine (INFLUENCES)

**Deeper Implications:**

The relationship between Knowledge Discovery and Personalized Medicine has far-reaching implications for the future of medicine, where tailored treatments will be made possible by uncovering hidden patterns and connections in patient data.

**Hypothesis:**

- Knowledge Discovery in genomics and proteomics can identify novel biomarkers and therapeutic targets for personalized medicine, leading to breakthroughs in disease diagnosis and treatment.

**General Principle:**

"Data-driven discoveries can lead to precision medicine, where insights from diverse sources inform targeted interventions, revolutionizing healthcare outcomes."

## Relationship: Personalized Medicine → Rare Genetic Disorders (RELATES-TO)

The deeper implications of this relationship between Personalized Medicine and Rare Genetic Disorders:

- **Advancements in Rare Genetic Disorder research may lead to breakthroughs in Personalized Medicine**, enabling tailored treatments for diverse populations with complex genetic profiles.
- **Novel Therapeutic Approaches** for Rare Genetic Disorders may also inform and improve Personalized Medicine strategies, benefiting patients with multiple genetic variants or complex diseases.

Hypothesis: Researchers studying Rare Genetic Disorders may discover novel gene-environment interactions that shed light on the intricate mechanisms underlying complex diseases, ultimately informing the development of Precision Medicine approaches.

General Principle: **"Understanding rare genetic disorders through cutting-edge research can catalyze innovations in personalized medicine, leading to more precise and effective treatments for diverse patient populations."**

# Final Synthesized Discovery

After analyzing the provided insights, I've synthesized them into a novel scientific discovery that combines the concepts of materials science, biology, machine learning, and environmental sustainability.

**Breakthrough Idea:**

Title: "Bio-Inspired, Adaptive Materials for Resilient Ecosystems (BAMES)"

Concept: BAMES proposes the development of self-healing, adaptive materials inspired by nature's resilience and ability to thrive in diverse environments. These materials will be designed to mimic the properties of microorganisms, plants, and animals, enabling them to repair and adapt to environmental stresses, pollutants, and changes.

**Key Components:**

1. **Microbial-inspired Self-Healing Materials:** Develop materials that incorporate microorganisms, such as bacteria or fungi, which can break down pollutants and repair damages. These microorganisms will be genetically engineered to produce enzymes that target specific pollutants, creating a self-healing mechanism.
2. **Plant-inspired Structural Integrity:** Incorporate plant-derived materials, such as cellulose or chitin, to enhance the structural integrity of the materials. These materials will provide flexibility, elasticity, and resistance to environmental stresses.
3. **Animal-inspired Adaptive Properties:** Integrate animal-derived materials, such as silk or spider silk, to create adaptive properties that enable the materials to change shape, color, or texture in response to environmental stimuli.
4. **Machine Learning-driven Optimization:** Utilize machine learning algorithms to optimize the design and composition of the materials, ensuring they can adapt to diverse environmental conditions and pollutants.
5. **Environmental Sensing and Feedback Loops:** Embed sensors and feedback loops that monitor environmental conditions, such as temperature, humidity, and pollutant levels, to trigger adaptive responses in the materials.

**Applications:**

1. **Infrastructure Design:** Develop BAMES materials for infrastructure design, such as roads, bridges, and buildings, to create resilient and adaptive structures that can withstand environmental stresses and pollutants.
2. **Water Treatment:** Create BAMES materials for water treatment systems, enabling them to remove pollutants and contaminants more effectively, and adapt to changing water quality conditions.
3. **Air Quality Management:** Develop BAMES materials for air quality management, such as filters and scrubbers, that can capture pollutants and adapt to changing air quality conditions.

4. **Waste Management:** Design BAMES materials for waste management, such as composting systems, that can break down organic waste and adapt to changing environmental conditions.

**Impact:**

The BAMES concept has the potential to revolutionize various fields, including:

1. **Environmental Sustainability:** By creating adaptive materials that can repair and adapt to environmental stresses, BAMES can reduce pollution, conserve resources, and promote sustainable development.
2. **Infrastructure Resilience:** BAMES materials can enhance the resilience of infrastructure systems, reducing maintenance costs and improving public safety.
3. **Economic Benefits:** The development of BAMES materials can create new industries, jobs, and revenue streams, driving economic growth and innovation.
4. **Human Health:** By reducing pollution and promoting sustainable development, BAMES can improve human health and well-being, particularly in vulnerable communities.

**Overarching Themes:**

1. **Integration of Nature-inspired Solutions:** BAMES integrates principles from biology, chemistry, and physics to create adaptive materials that mimic nature's resilience.
2. **Adaptive Materials for Resilient Ecosystems:** The concept focuses on developing materials that can repair and adapt to environmental stresses, pollutants, and changes.
3. **Machine Learning-driven Optimization:** BAMES utilizes machine learning algorithms to optimize material design and composition, ensuring they can adapt to diverse environmental conditions.
4. **Circular Economy and Waste Reduction:** The development of BAMES materials promotes a circular economy by reducing waste, conserving resources, and promoting sustainable development.

**Conclusion:**

The Bio-Inspired, Adaptive Materials for Resilient Ecosystems (BAMES) concept offers a novel approach to creating self-healing, adaptive materials that can repair and adapt to environmental stresses, pollutants, and changes. By integrating principles from biology, machine learning, and environmental sustainability, BAMES has the potential to revolutionize various fields, including infrastructure design, water treatment, air quality management, and waste management.

## Methods

Implementing the Bio-Inspired, Adaptive Materials for Resilient Ecosystems (BAMES) concept requires a multidisciplinary approach, combining expertise from materials science, biology, machine learning, and environmental engineering. Here are some specific methods, approaches, and techniques to consider:

**Materials Synthesis and Characterization:**

1. **Microbial-inspired Self-Healing Materials:**
    - Use microbial fermentation techniques to produce bio-based polymers with self-healing properties.
    - Characterize the mechanical and thermal properties of these materials using techniques like tensile testing, differential scanning calorimetry (DSC), and thermogravimetric analysis (TGA).
2. **Plant-inspired Structural Integrity:**
    - Extract cellulose or chitin from plant sources and blend it with other biopolymers to create composite materials.
    - Evaluate the mechanical properties of these composites using techniques like tensile testing, flexural testing, and impact testing.
3. **Animal-inspired Adaptive Properties:**
    - Use biomimetic techniques to synthesize silk or spider silk proteins and integrate them into polymer matrices.
    - Characterize the optical and electrical properties of these materials using techniques like spectroscopy, ellipsometry, and conductivity measurements.

**Machine Learning-driven Optimization:**

1. **Design of Experiments (DOE):**
   - Use DOE techniques to identify the most critical parameters influencing the performance of BAMES materials.
   - Develop machine learning models to predict the behavior of these materials under different environmental conditions.
2. **Genetic Programming:**
   - Employ genetic programming algorithms to evolve optimal material compositions and structures that maximize their performance.
   - Use evolutionary algorithms to search for the best solutions within a given parameter space.
3. **Neural Networks:**
   - Train neural networks to learn patterns in data related to BAMES material performance and environmental conditions.
   - Use these networks to predict material behavior and optimize material design.

**Environmental Sensing and Feedback Loops:**

1. **Sensors and Actuators:**
   - Develop sensors to detect environmental conditions like temperature, humidity, and pollutant levels.
   - Design actuators that can respond to these conditions by triggering adaptive responses in the materials.
2. **Feedback Control Systems:**
   - Implement feedback control systems that monitor material performance and adjust environmental conditions accordingly.
   - Use machine learning algorithms to optimize the control strategy and minimize energy consumption.

**Challenges and Limitations:**

1. **Scalability:**
   - Scaling up the production of BAMES materials while maintaining their performance and cost-effectiveness is a significant challenge.
2. **Material Degradation:**
   - Understanding the degradation mechanisms of BAMES materials under various environmental conditions is crucial for optimizing their performance.
3. **Interfacial Interactions:**
   - Managing interfacial interactions between different components of BAMES materials is essential for achieving optimal performance.
4. **Standardization:**
   - Standardizing the synthesis, characterization, and testing protocols for BAMES materials is necessary for ensuring reproducibility and comparability across different studies.
5. **Regulatory Frameworks:**
   - Developing regulatory frameworks that accommodate the use of BAMES materials in various applications is an ongoing challenge.

**Future Research Directions:**

1. **Multiscale Modeling:**
   - Developing multiscale models that simulate the behavior of BAMES materials at different length scales (e.g., molecular, mesoscale, macroscale) is essential for understanding their performance.
2. **Biocompatibility and Toxicity:**
   - Investigating the biocompatibility and toxicity of BAMES materials is crucial for ensuring their safe use in various applications.
3. **Energy Harvesting:**
   - Exploring the possibility of harnessing energy from environmental conditions (e.g., solar, wind, kinetic) to power BAMES materials is an exciting area of research.

By addressing these challenges and limitations, researchers can unlock the full potential of the BAMES concept and develop innovative materials that can contribute to a more sustainable future.

## Example

I'd like to focus on the "Animal-inspired Adaptive Properties" component of the BAMES system. Specifically, I'll explore the integration of spider silk-derived materials with machine learning-driven optimization.

**Novel Aspect:**

One highly novel aspect of this system is the potential for the adaptive materials to exhibit "temporal memory," allowing them to learn and remember past environmental conditions and adjust their behavior accordingly. This could be achieved through the incorporation of artificial neural networks (ANNs) within the materials, which would enable them to process and store information about previous environmental stimuli.

**Predicted Unusual Behaviors:**

Based on this concept, here are some predicted unusual behaviors that may occur:

1. **Temporal Memory-based Adaptation:** In response to repeated exposure to high temperatures, the adaptive materials may develop a "memory" of the thermal stress and adjust their structure and properties to become more resistant to heat damage over time. For example, they might undergo phase transitions to form a more stable crystal lattice, reducing their thermal conductivity and increasing their durability.
2. **Self-Sustaining Oscillations:** When exposed to periodic environmental fluctuations, such as changes in humidity or light intensity, the adaptive materials may exhibit self-sustaining oscillations. These oscillations could lead to the emergence of complex patterns and structures, potentially even exhibiting autonomous behavior.
3. **Heterochronic Development:** As the adaptive materials interact with their environment, they may undergo heterochronic development, where different parts of the material develop at different rates. This could result in the formation of complex, dynamic structures that respond to environmental cues in non-linear ways.
4. **Emergent Behavior through Interconnectedness:** The interconnectedness of the adaptive materials, facilitated by the incorporation of ANNs and sensor networks, may give rise to emergent behavior that cannot be predicted from the properties of individual components. For instance, the materials might develop a collective "awareness" of their surroundings, leading to coordinated responses to environmental threats.

**Scientifically Sound Reasoning:**

These predictions are grounded in our understanding of complex systems, nonlinear dynamics, and the behavior of biological and synthetic materials. The integration of ANNs and sensor networks within the adaptive materials allows for the processing and storage of information about environmental stimuli, enabling the materials to adapt and respond to changing conditions.

Furthermore, the incorporation of spider silk-derived materials provides a unique combination of mechanical, optical, and electrical properties, which can be leveraged to create complex, adaptive structures. The potential for temporal memory, self-sustaining oscillations, heterochronic development, and emergent behavior through interconnectedness highlights the innovative and interdisciplinary nature of the BAMES system.

By exploring these novel aspects and predicting unusual behaviors, we can gain a deeper understanding of the potential capabilities and limitations of the BAMES system, ultimately informing its design and application in real-world contexts.

**Supporting Text 2:** Multi-step compositional discovery report based on Graph 2. Steps A-C are conducted using `meta-llama/Llama-3.2-3B-Instruct`, and the Final Step D is conducted using `meta-llama/Llama-3.3-70B-Instruct`.

# Multi-Step Compositional Discovery

## Longest Path

Materials for infrastructure design → Biodegradable Microplastic Materials → Pollution mitigation → Self-healing Materials in Infrastructure Design → Development of novel materials for infrastructure design → Pollution Mitigation → Self-healing materials → Environmental Sustainability → Impact-Resistant Materials → Machine Learning (ML) Algorithms → Predictive Modeling → AI Techniques → Data Analysis → Knowledge Discovery → Personalized Medicine → Rare Genetic Disorders

## Node Definitions (Step A)

**Materials for infrastructure design**: **Definition:** Structural materials for infrastructure design serve as the fundamental components that provide support and durability to various structures such as bridges, buildings, and roads.

**Property/Feature:** High compressive strength, which enables structural materials to withstand significant loads without collapsing or deforming excessively.

**Biodegradable Microplastic Materials**: **Definition:** Biodegradable microplastic materials refer to synthetic polymers designed to break down naturally in the environment, typically through microbial action, reducing plastic waste and environmental pollution.

**Key Property: Degradability**: Biodegradable microplastic materials exhibit varying levels of degradation rates, which can influence their interaction with decomposing organisms, affecting nutrient cycling and ecosystem balance.

**Pollution mitigation**: **Definition:** Pollution Mitigation refers to the intentional actions and technologies used to reduce the amount of pollutants released into the environment, minimizing harm to ecosystems and human health.

**Interacting Property:** Interconnectedness - The implementation of pollution mitigation strategies often requires collaboration between multiple stakeholders, industries, and governments, highlighting the importance of interdisciplinary approaches in addressing this complex issue.

**Self-healing Materials in Infrastructure Design**: **Definition:** Self-healing materials are specially designed polymers, ceramics, or composites engineered to repair cracks and damages autonomously through the use of microcapsules containing healing agents.

**Property/Feature: Microcapsule-based Healing Mechanism**: The self-healing mechanism relies on the controlled release of healing agents from microcapsules, which can be triggered by changes in temperature, pH, or mechanical stress, allowing for potential integration with thermally responsive polymers.

**Development of novel materials for infrastructure design**: **Development of Novel Materials for Infrastructure Design**

Definition: The development of novel materials for infrastructure design involves the creation and synthesis of innovative materials with improved properties to enhance the structural integrity, sustainability, and resilience of built environments.

Property Feature: High strength-to-weight ratio - This property enables the material to support significant loads while minimizing its weight, making it suitable for applications where space is limited, such as in urban construction or transportation systems.

**Pollution Mitigation**: **Definition:** Pollution Mitigation refers to the intentional actions and technologies designed to reduce the adverse effects of pollutants on the environment, human health, and ecosystems. These efforts aim to minimize waste, emissions, and toxic substances in various settings.

**Interacting Property: Carbon Sequestration**, as pollution mitigation strategies often involve techniques to capture and store $CO_2$, which can be integrated with carbon sequestration methods to enhance their effectiveness.

**Self-healing materials**: **Definition:** Self-healing materials are advanced composites capable of autonomously repairing cracks, damages, or degradation through the use of inherent healing mechanisms, such as microcapsules or polymers.

**Key Property: Thermoreversible Gelation**: The ability to reversibly form and break thermoreversible gels, which can encapsulate repair agents and release them upon exposure to heat or light, allowing self-healing processes.

**Environmental Sustainability**: **Definition:** Environmental sustainability refers to the long-term management and conservation of natural resources, minimizing human impact on the environment while promoting ecological balance and resilience. It encompasses various aspects, including energy, water, waste, and biodiversity.

**Key Property/Feature:** Interconnectedness - The concept of environmental sustainability relies heavily on interconnectedness among different ecosystems, species, and human activities, requiring holistic approaches to address environmental challenges.

**Impact-Resistant Materials**: **Definition:** Impact-Resistant Materials are engineered composites designed to absorb and dissipate mechanical energy upon impact, often used in applications where structural integrity is compromised by external forces.

**Interacting Property:** High toughness, allowing materials to deform plastically before failing, enabling potential interactions with concepts related to fracture mechanics, material failure, or energy absorption mechanisms.

**Machine Learning (ML) Algorithms**: **Definition:** Machine Learning (ML) algorithms are computational models that enable systems to learn from data and improve their performance on specific tasks through experience, without being explicitly programmed.

**Key Feature: Adaptability**, which allows ML algorithms to adjust to new data distributions and evolving patterns, making them highly versatile and applicable in various domains.

**Predictive Modeling**: **Predictive Modeling** Definition: A systematic approach to create mathematical representations of complex systems, enabling the forecast and optimization of future events based on historical data and patterns. Property/Feature: **Data Driven**, as it relies heavily on empirical evidence from large datasets to inform predictions and decision-making processes.

**AI Techniques**: Here's a list of AI techniques as building blocks:

**1. Supervised Learning** Definition: A machine learning approach where the model learns from labeled data to make predictions. Property/Feature: Can be combined with **Data Preprocessing** to enhance accuracy and robustness.

**2. Reinforcement Learning** Definition: A type of machine learning where an agent learns through trial and error by interacting with its environment. Property/Feature: Can be integrated with **Exploration Strategies** to balance exploration and exploitation.

**3. Neural Networks** Definition: A complex network of interconnected nodes (neurons) that process and transmit information. Property/Feature: Can be layered upon **Decision Trees** to create hybrid models.

**4. Clustering Algorithms** Definition: Methods used to group similar data points into clusters based on their characteristics. Property/Feature: Can be combined with **Dimensionality Reduction** techniques to reduce noise in high-dimensional datasets.

**5. Natural Language Processing (NLP)** Definition: The subfield of artificial intelligence focused on the interaction between computers and humans in natural language. Property/Feature: Can be paired with **Text Analysis** techniques to extract insights from unstructured text data.

**6. Decision Trees** Definition: A tree-like model that splits data into subsets based on features to predict outcomes. Property/Feature: Can be nested within **Neural Networks** to create ensemble models.

**7. Generative Adversarial Networks (GANs)** Definition: A type of deep learning model consisting of two neural networks that compete to generate new data samples. Property/Feature: Can be used to augment **Supervised Learning** models with generated data.

Let me know if you'd like more!

**Data Analysis**: **Data Analysis** Definition: The systematic examination of data to identify patterns, trends, and relationships within it. Property/Feature: **Interpretability**, which enables analysts to draw meaningful conclusions from complex datasets by extracting insights that can be communicated to others.

Example use case: A data analyst uses interpretability to inform a machine learning model's hyperparameter tuning, allowing the model to better align its predictions with real-world outcomes.

**Knowledge Discovery**: **Knowledge Discovery** A process of identifying patterns, relationships, and insights from data to inform decision-making and advance understanding in various fields. It involves the application of algorithms, statistical methods, and domain expertise to extract meaningful information from complex datasets.

**Interacting Property: Data**: Can be used as input for various machine learning models, enabling the discovery of knowledge through automated analysis and prediction.

Would you like me to define another concept?

**Personalized Medicine**: **Personalized Medicine** Definition: A medical approach tailored to an individual's unique genetic profile, lifestyle, and health status to provide targeted prevention, diagnosis, and treatment of diseases. Feature: Interacts well with 'Genomics' as it relies heavily on the analysis of specific genes and their variations to inform treatment decisions.

**Building Block: Epigenetics** Definition: The study of chemical modifications to DNA or histone proteins that can influence gene expression without altering the underlying DNA sequence. Feature: Interacts well with 'Gene Expression' as epigenetic changes can modulate gene activity, affecting how genes are turned on or off in response to environmental factors.

**Building Block: Synthetic Biology** Definition: An interdisciplinary field that designs and constructs new biological systems, such as microorganisms, to perform specific functions or produce desired products. Feature: Interacts well with 'Biotechnology' as synthetic biology often involves the use of biotechnological tools and techniques to engineer novel biological pathways and organisms.

**Building Block: Metagenomics** Definition: The study of microbial communities through the analysis of their genomic material, without prior knowledge of their composition or function. Feature: Interacts well with 'Microbiome Research' as metagenomics provides insights into the complex interactions between microorganisms and their environment, shedding light on the role of microbes in human health and disease.

**Building Block: CRISPR Gene Editing** Definition: A precise tool for editing genes by making double-stranded breaks in DNA, which can be repaired in a targeted manner to introduce specific changes. Feature: Interacts well with 'Stem Cell Therapy' as CRISPR can be used to modify stem cells, allowing for the development of novel therapies for regenerative medicine and tissue engineering applications.

**Rare Genetic Disorders**: **Definition:** Rare Genetic Disorders refer to conditions caused by mutations in specific genes or genetic variants that affect an individual's development, function, and/or survival, often resulting from inherited traits or environmental factors.

**Property/Feature: Genetic Heterozygosity**: Individuals with rare genetic disorders may exhibit genetic heterozygosity, where they possess two different alleles of the mutated gene, leading to variable expression of symptoms and potential interactions with other genetic variants.

# Pairwise Synergies (Step B)

**Materials for infrastructure design → Biodegradable Microplastic Materials** (IS-A): Compositional Synergy: "Eco-Resilient Infrastructure Design"

Joining Building Blocks A and B, we propose an innovative approach to infrastructure design that incorporates biodegradable microplastics into structural materials. This synergy enables the creation of self-healing, adaptive, and sustainable infrastructure systems.

**Composition:**

- Use biodegradable microplastics as reinforcing fibers or fillers in concrete, enhancing its compressive strength and durability.
- Incorporate microorganisms that accelerate biodegradation, allowing the material to degrade at a controlled rate, releasing nutrients and promoting ecosystem health.
- Design infrastructure systems that integrate green spaces, vegetation, and water management systems to mitigate urban heat islands and promote biodiversity.

**Principle:**

The "Bio-Durability Nexus" principle generalizes this synergy by stating that the integration of biodegradable materials, living organisms, and technological innovations can create resilient, adaptable, and sustainable systems that optimize resource efficiency, minimize waste, and promote ecological balance.

This principle can be applied to various scientific domains, such as:

- Environmental engineering: Developing bioremediation technologies to restore ecosystems.
- Urban planning: Creating green infrastructure that mitigates climate change and promotes urban resilience.
- Medical research: Investigating the use of biodegradable materials in tissue engineering and regenerative medicine.

By embracing the Bio-Durability Nexus, we can foster a holistic understanding of complex systems and develop innovative solutions that harmonize human needs with environmental sustainability.

**Biodegradable Microplastic Materials → Pollution mitigation** (RELATES-TO): **Compositional Synergy:**

Biodegradable microplastic materials (Block A) can be integrated with pollution mitigation strategies (Block B) to create a novel approach for sustainable waste management.

**Composition:** Implement biodegradable microplastics as an eco-friendly alternative for traditional plastics in packaging, textiles, and disposable products. This will not only reduce plastic waste but also provide a means for pollution mitigation by decreasing the amount of microplastics entering waterways.

**Proposed Principle:**

**Synergetic Co-Creation**

This principle proposes that combining seemingly disparate elements from different domains can lead to innovative solutions by leveraging the strengths of each component. By co-creating systems that integrate diverse elements, researchers and practitioners can develop novel approaches that address complex problems more effectively than isolated efforts.

**Application Potential:**

This synergetic co-creation principle can be applied to various scientific domains, such as:

- Environmental science: Combining biotechnology and ecology to develop more effective methods for pollution remediation and waste management.
- Healthcare: Integrating nanotechnology and biomaterials to create more efficient medical devices and treatments.
- Energy: Merging renewable energy sources and advanced materials to enhance energy storage and conversion efficiency.

By embracing synergetic co-creation, scientists and innovators can unlock new possibilities for addressing pressing global challenges.

**Pollution mitigation → Self-healing Materials in Infrastructure Design** (RELATES-TO): **Compositional Synergy:**

Title: "Smart Infrastructure for Sustainable Ecosystems"

Building Blocks A and B come together to create a novel approach that integrates pollution mitigation and self-healing materials in infrastructure design.

Concept: Develop intelligent, adaptive infrastructure systems that utilize self-healing materials to repair damage caused by environmental factors, such as pollution-induced degradation. These systems would incorporate sensors and monitoring technology to detect early signs of deterioration, triggering the release of healing agents from microcapsules embedded within the infrastructure.

**Principle:**

"Self-Sustaining Resilience in Complex Systems (SSRS)"

This principle posits that integrated, adaptive systems can achieve resilience and sustainability by leveraging interconnected components that respond to changing conditions, fostering a dynamic equilibrium between system performance and

environmental impact. By applying SSRS, complex systems can become more resilient to disturbances, adapt to new challenges, and ultimately minimize harm to ecosystems and human health.

**Applications:**

- Smart grids and energy infrastructure
- Water treatment facilities and wastewater management
- Transportation systems and urban planning
- Environmental remediation and ecosystem restoration

By combining building blocks A and B, we can create innovative solutions that not only mitigate pollution but also foster sustainable, resilient ecosystems.

**Self-healing Materials in Infrastructure Design → Development of novel materials for infrastructure design** (INFLUENCES): **Compositional Synergy: "Smart Repairable Infrastructure"**

Combine Self-Healing Materials (Block A) with Novel Materials for Infrastructure Design (Block B):

- Develop self-healing concrete using thermally responsive polymers infused with microcapsules containing healing agents (Block A).
- Integrate this material into novel, high-strength-to-weight ratio composite structures (Block B) for urban construction, transportation systems, or disaster-resistant buildings.

**Proposed Principle: "Autonomous Resilience Integration"**

Generalize this synergy by proposing the following principle:

"Autonomous Resilience Integration (ARI): Designing complex systems that integrate autonomous repair mechanisms with adaptive, optimized material properties to achieve enhanced resilience, sustainability, and performance under various environmental conditions."

This principle can be applied to other scientific domains, such as robotics, biomedical engineering, or energy harvesting, where autonomous self-repair and adaptability are crucial for achieving optimal performance and resilience.

**Development of novel materials for infrastructure design → Pollution Mitigation** (INFLUENCES): **Compositional Synergy: Sustainable Infrastructure Development through Material Innovation and Pollution Reduction**

By integrating the development of novel materials for infrastructure design (Block A) with pollution mitigation strategies (Block B), we can create sustainable infrastructure systems that not only minimize environmental impact but also promote long-term durability and performance.

**Composition:**

- Develop novel materials for infrastructure design (e.g., self-healing concrete, smart pavement) that integrate carbon sequestration capabilities, reducing emissions and waste.
- Implement pollution mitigation technologies (e.g., air quality monitoring, wastewater treatment) within the infrastructure design process.
- Use data analytics and AI to optimize material selection, construction processes, and maintenance strategies, ensuring minimal environmental footprint throughout the lifecycle of the infrastructure.

**Generalized Principle:**

**"Resilient Systems Synthesis":**

This principle posits that the integration of diverse disciplinary approaches can lead to the creation of more resilient, adaptable, and sustainable systems. By combining seemingly disparate elements from different fields, researchers and practitioners can develop innovative solutions that address complex problems at multiple scales.

**Applicability:**

The Resilient Systems Synthesis principle can be applied to various scientific domains, including:

- Environmental engineering: Integrating pollution mitigation strategies with renewable energy systems to create net-positive energy infrastructure.
- Biomedical research: Combining biomaterials development with precision medicine approaches to create personalized healthcare systems.
- Urban planning: Fusing sustainable transportation systems with green infrastructure to create resilient and thriving cities.

By embracing this principle, scientists, engineers, and policymakers can foster collaborative innovation and drive the development of more sustainable, equitable, and resilient systems across various disciplines.

**Pollution Mitigation → Self-healing materials** (RELATES-TO): **Composition Synergy:**

Combining Pollution Mitigation (A) with Self-healing Materials (B), we propose a novel approach: **Environmental Self-healing Systems (E-SHS)**.

In E-SHS, self-healing materials are engineered to not only repair cracks and damages but also actively mitigate environmental pollutants. This is achieved by incorporating thermoreversible gelation properties into pollution mitigation strategies.

The composition synergy enables pollutants like oil spills or chemical leaks to be encapsulated within thermoreversible gels, which then break down and dissipate harmlessly when exposed to heat or light. Simultaneously, the self-healing material repairs any damage caused by the pollutant's interaction with its surroundings.

**Generalized Principle:**

This synergy can be generalized as the **"Healing-Gating Principle"**: By integrating intrinsic self-healing mechanisms with adaptive control systems, complex systems can adaptively respond to disturbances and repair themselves while minimizing further damage. This principle can be applied to various domains, including robotics, biomedical engineering, and environmental sustainability.

By embracing the Healing-Gating Principle, scientists can develop more resilient, adaptive, and sustainable solutions for tackling complex problems across diverse fields.

**Self-healing materials → Environmental Sustainability** (RELATES-TO): **Compositional Synergy:**

Integrating Self-Healing Materials (A) with Environmental Sustainability (B) yields a novel approach to "Eco-Repair" systems.

**Concept:** Eco-Repair Systems harness the self-healing capabilities of advanced materials to restore degraded ecosystems, fostering environmental sustainability.

**Mechanism:**

1. Thermoreversible gelation-based self-healing materials are used to create durable, biocompatible barriers for damaged ecosystems.
2. These barriers are designed to respond to environmental stressors, releasing repair agents to stimulate local recovery.
3. By monitoring ecosystem health, the system adapts its healing strategy, ensuring efficient resource allocation and minimizing unintended consequences.

**Generalized Principle:**

The compositional synergy between Self-Healing Materials and Environmental Sustainability can be generalized as follows:

"The adaptive integration of autonomous repair mechanisms into complex systems can promote resilience, stability, and symbiotic relationships within intricate networks, thereby enhancing overall environmental sustainability."

**Environmental Sustainability → Impact-Resistant Materials** (INFLUENCES): Compositional Synergy:

Introducing "Eco-Toughened Materials" - A novel approach that combines the principles of Environmental Sustainability (A) and Impact-Resistant Materials (B).

**Concept:** Eco-Toughened Materials are engineered composites designed to not only withstand environmental impacts but also promote ecological balance and minimize harm to ecosystems.

**Composition:**

- Sustainable raw materials from renewable sources (e.g., plant-based polymers)
- Impact-resistant polymer matrices infused with eco-friendly additives (e.g., recycled plastics, biodegradable fillers)

**Properties:**

- Enhanced toughness and durability
- Reduced environmental footprint
- Improved recyclability and upcycling capabilities

**Principle:** Generalizable to other domains, this synergy can be formulated as:

**"Material Resilience + Ecological Responsiveness = Adaptive Ecosystems"**

This principle highlights the importance of integrating material properties with ecological considerations to create adaptive systems that mitigate environmental impacts while maintaining their functionality and performance.

**Impact-Resistant Materials → Machine Learning (ML) Algorithms** (RELATES-TO): **Compositional Synergy:**

By combining Impact-Resistant Materials (A) and Machine Learning (ML) Algorithms (B), we can develop **Damage Forecasting Systems** for critical infrastructure, such as bridges, buildings, or pipelines. These systems utilize ML algorithms to analyze sensor data and predict the likelihood of damage due to environmental factors, operational stress, or potential impacts. The predicted damage scenarios inform maintenance schedules, reducing downtime and ensuring structural integrity.

**Principle:**

The proposed principle, "Synergistic Resilience," posits that by integrating adaptive, data-driven approaches (e.g., ML) with robust, impact-resistant designs (e.g., A), complex systems can achieve enhanced resilience against unforeseen events. This synergy enables systems to adapt to changing conditions, predict potential failures, and respond proactively to mitigate damage and ensure continued functionality.

This principle can be generalized to other scientific domains, such as:

- Predictive Maintenance: Integrating sensor data analysis (ML) with durable design principles (A) to anticipate equipment failures and schedule necessary repairs.
- Climate Change Adaptation: Combining climate modeling (ML) with sustainable, impact-resistant construction techniques (A) to build resilient infrastructure that can withstand projected climate-related stresses.

**Machine Learning (ML) Algorithms → Predictive Modeling** (RELATES-TO): **Compositional Synergy:**

Combine Building Blocks A (Machine Learning Algorithms) and B (Predictive Modeling) to form **Explainable Predictive Models (EPMs)**.

**Concept:** EPMs leverage machine learning algorithms to create predictive models that provide transparent and interpretable results, enabling data-driven decision-making by providing insights into the underlying mechanisms and relationships.

**Composition:**

1. Use machine learning algorithms (Block A) to develop predictive models (Block B) that forecast future outcomes based on historical data.
2. Implement techniques like feature importance, partial dependence plots, and SHAP values to provide interpretability and explainability of model outputs.

**Proposed Principle:**

**"Synergy via Analogical Mapping":**

This principle proposes that combining seemingly disparate concepts from different domains can lead to innovative solutions by leveraging analogical thinking. By mapping analogous structures, properties, or principles between domains, researchers can identify novel connections and create new approaches that transcend traditional boundaries.

Applicability:

- This principle can be applied to various scientific domains, such as physics and biology, where combining concepts from different fields can lead to breakthroughs in understanding complex phenomena.
- It can also be used in interdisciplinary research, where integrating ideas from multiple disciplines can foster creative problem-solving and innovation.

**Predictive Modeling → AI Techniques** (RELATES-TO): **Compositional Synergy:**

Combine Predictive Modeling (A) with AI Techniques (B) to develop **AI-Driven Predictive Systems**.

- Integrate Supervised Learning and Decision Trees to create accurate predictions (Combines Data-Driven properties)
- Use Reinforcement Learning and Exploration Strategies to optimize decision-making processes (Combines trial-and-error learning with analytical thinking)
- Leverage Neural Networks and Clustering Algorithms to improve model robustness and adaptability (Leverages hierarchical structure for better feature representation)
- Apply NLP and Text Analysis to extract valuable insights from unstructured data (Leverages human-computer interaction)

This synergy enables predictive modeling to become more intelligent, adaptive, and informed by real-world experiences.

**Proposed Principle:**

"**Hybridization**: Combining diverse disciplines and techniques to create novel, adaptive, and context-sensitive solutions by leveraging the strengths of each component."

This principle generalizes the synergy between Predictive Modeling and AI Techniques, suggesting that other scientific domains or situations can benefit from integrating complementary approaches to achieve more effective and efficient results.

**AI Techniques → Data Analysis** (RELATES-TO): **Compositional Synergy:**

By combining building block (A) - AI Techniques with building block (B) - Data Analysis, we can develop **Explainable Machine Learning (XML)** approaches.

XML aims to bridge the gap between the complexity of machine learning models and human understanding, ensuring that insights are interpretable and actionable.

**Composition Example:**

1. Utilize supervised learning (block A) to train a machine learning model on annotated data.
2. Apply clustering algorithms (block A) to group similar data points, then integrate dimensionality reduction techniques (block A) to reduce noise.
3. Feed the preprocessed data into a decision tree (block A), which is then nested within a neural network (block A) for enhanced prediction accuracy.
4. Employ natural language processing (block A) to analyze unstructured text data, and pair it with text analysis techniques (block B) to extract meaningful insights.
5. Use reinforcement learning (block A) to balance exploration and exploitation during hyperparameter tuning, guided by interpretability (block B).

**Principle:**

The synergy between AI Techniques and Data Analysis can be generalized to **Scientific Discovery through Hybridization** (SDH). This principle involves combining distinct scientific disciplines to foster innovative approaches that leverage the strengths of each domain.

SDH encourages researchers to adopt a multidisciplinary mindset, integrating concepts from physics, biology, mathematics, and computer science to tackle complex problems and drive breakthroughs.

By embracing SDH, scientists can develop novel theories, models, and methods that push the boundaries of human knowledge and understanding.

**Data Analysis → Knowledge Discovery** (IS-A): I'd like to propose combining the two building blocks into a new concept: **Explainable Insights**

Definition: The systematic examination of data to derive interpretable knowledge that informs decision-making and advances understanding in various fields.

Property/Feature: By leveraging explainability and knowledge discovery, Explainable Insights enable analysts to distill complex data into actionable recommendations that are grounded in transparent and understandable reasoning.

Composition:

- Building block (A): Data Analysis provides the foundation for examining data.
- Building block (B): Knowledge Discovery applies algorithms and statistical methods to uncover hidden patterns and relationships.

Synergy: By integrating explainability into the knowledge discovery process, Explainable Insights empower analysts to communicate complex findings effectively, ensuring that insights are both actionable and trustworthy.

Proposed Principle: The principle of "Transparency-Driven Discovery" states that any scientific approach should prioritize transparency in data analysis and interpretation, thereby fostering trustworthiness and reliability in the derived insights.

This principle can be applied to other domains where data-driven decision-making is critical, such as policy development, public health, or business strategy.

**Knowledge Discovery → Personalized Medicine** (INFLUENCES): **Composition of Building Blocks**

Combining "Knowledge Discovery" (A) with "Personalized Medicine" (B), we can create a new concept: **Precision Medicine Informatics**.

Precision Medicine Informatics leverages advanced analytical techniques and computational models to integrate vast amounts of data, including genomics, transcriptomics, and clinical information, to create personalized treatment plans.

This integration enables healthcare professionals to make informed decisions about patient care, optimizing treatment outcomes and improving overall health.

**Proposed Principle: Synergy through Data-Driven Integration**

The proposed principle is: **"Synergy emerges when diverse disciplines converge through data-driven integration, enabling the creation of novel, context-dependent solutions."**

This principle highlights the potential for synergy across seemingly disparate domains when they are united by a shared focus on data-driven integration. By recognizing the power of converging diverse perspectives and approaches, scientists and practitioners can unlock innovative solutions, foster interdisciplinary collaboration, and drive progress in various fields.

**Personalized Medicine → Rare Genetic Disorders** (RELATES-TO): **Composition of Building Blocks**

Combining the building blocks (A) and (B), we can propose a new concept:

**Precision Medicine for Rare Genetic Disorders**

This approach integrates personalized medicine, epigenetics, synthetic biology, metagenomics, and CRISPR gene editing to develop targeted treatments for rare genetic disorders. By analyzing an individual's unique genetic profile, lifestyle, and health status using genomics and epigenetics, clinicians can identify the root causes of the disorder. This information is then used to design and construct novel biological systems, such as microorganisms, to produce therapeutic compounds or repair damaged tissues. Additionally, metagenomics and microbiome research provide insights into the complex interactions between microorganisms and their environment, allowing for a more comprehensive understanding of the disorder's mechanisms.

**Proposed Principle: Synergetic Adaptation**

The compositional synergy between the building blocks can be generalized into a principle:

**Synergetic Adaptation**: When combining diverse approaches and disciplines, adaptability becomes a key factor in achieving successful outcomes. By embracing complexity, uncertainty, and interdependencies, scientists can create novel solutions that synergistically integrate disparate elements, fostering innovation and breakthroughs in various fields.

This principle can be applied to other scientific domains, such as:

- Integrating machine learning and human-centered design to improve user experience
- Combining materials science and social sciences to develop sustainable technologies
- Fusing computer vision and neuroscience to advance our understanding of cognitive processes

By embracing synergetic adaptation, scientists can overcome the limitations of traditional disciplinary boundaries and unlock innovative solutions that transform various fields.

## Bridge Synergies (Step C)

After analyzing the provided synergy statements, I have identified three central themes or patterns:

**Theme 1: Intersection of Environmental Sustainability and Technological Innovation**

This theme revolves around the combination of environmentally conscious approaches with cutting-edge technologies to create more sustainable solutions.

**Bridge Synergy 1** The intersection of environmental sustainability and technological innovation can be strengthened by exploring the synergies between biodegradable microplastic materials and self-healing materials. By developing materials that can repair and adapt to environmental changes, we can create infrastructure systems that not only minimize waste but also promote ecological balance. This, in turn, enables the development of more resilient and sustainable ecosystems, ultimately leading to a reduced carbon footprint. Furthermore, the application of machine learning and AI techniques can help optimize the design and deployment of these materials, ensuring their effectiveness in real-world applications.

**Theme 2: Holistic Understanding of Complex Systems**

This theme emphasizes the importance of considering multiple factors and perspectives when designing and developing complex systems, leading to a more comprehensive understanding of their behavior and performance.

**Bridge Synergy 2** The synergy between predictive modeling and AI techniques can be reinforced by integrating explainable insights into the knowledge discovery process. By focusing on transparency and interpretability, we can create more reliable and trustworthy models that capture the complexities of real-world systems. This, in turn, enables a more holistic understanding of these systems, allowing us to identify potential vulnerabilities and optimize their performance. Moreover, the application of hybridization principles can facilitate the convergence of diverse disciplines, fostering innovative solutions that consider multiple factors and perspectives.

**Theme 3: Convergence of Diverse Disciplines**

This theme highlights the potential benefits of combining insights and approaches from multiple scientific domains to drive breakthroughs and innovation.

**Bridge Synergy 3** The synergy between knowledge discovery and personalized medicine can be strengthened by exploring the intersection of genomics, epigenetics, and synthetic biology. By leveraging advanced analytical techniques and computational models, we can develop novel, context-dependent solutions that tailor treatment plans to individual patients. This, in turn, enables the creation of more effective and targeted therapies, ultimately leading to improved patient outcomes. Furthermore, the application of synergetic adaptation principles can facilitate the convergence of diverse disciplines, fostering innovative solutions that consider multiple factors and perspectives.

## Expanded Final Discovery (Step D)

**Introduction to EcoCycle: A Novel Framework for Sustainable Infrastructure Development**

EcoCycle is a groundbreaking, interdisciplinary approach that converges environmental sustainability, technological innovation, and holistic understanding of complex systems. This novel framework integrates insights from biodegradable microplastic materials, self-healing materials, predictive modeling, AI techniques, knowledge discovery, and personalized medicine to create a comprehensive methodology for developing sustainable infrastructure.

**Composition of EcoCycle**

1. **Biodegradable Microplastic Materials**: Provide the foundation for creating sustainable, eco-friendly infrastructure components.
2. **Self-Healing Materials**: Enable the development of adaptive, resilient systems that can repair and maintain themselves.
3. **Predictive Modeling**: Utilize advanced analytical techniques to forecast environmental impacts, optimize resource allocation, and identify potential vulnerabilities.
4. **AI Techniques**: Integrate machine learning algorithms to enhance the accuracy of predictive models, optimize system performance, and facilitate real-time monitoring.
5. **Knowledge Discovery**: Apply data-driven approaches to uncover hidden patterns, relationships, and insights that inform infrastructure design and development.
6. **Personalized Medicine**: Leverage genomics, epigenetics, and synthetic biology to develop targeted, context-dependent solutions for maintaining ecosystem health.

**Synergy between Components**

The synergy between biodegradable microplastic materials and self-healing materials enables the creation of infrastructure components that not only minimize waste but also promote ecological balance. Predictive modeling and AI techniques work in tandem to optimize system performance, predict potential failures, and facilitate proactive maintenance. Knowledge discovery informs the development of novel materials, designs, and technologies that converge to create sustainable infrastructure.

**EcoCycle Framework**

The EcoCycle framework consists of four interconnected modules:

1. **Design and Development**: Integrates biodegradable microplastic materials, self-healing materials, and predictive modeling to create sustainable infrastructure designs.
2. **Performance Optimization**: Utilizes AI techniques, knowledge discovery, and personalized medicine to optimize system performance, predict potential failures, and facilitate proactive maintenance.
3. **Real-Time Monitoring**: Employs machine learning algorithms and data analytics to monitor system performance, detect anomalies, and provide insights for continuous improvement.
4. **Adaptive Evolution**: Enables the infrastructure to adapt to changing environmental conditions, integrating new technologies, and evolving designs to ensure long-term sustainability.

**Cross-Domain Applications**

EcoCycle has far-reaching implications across various disciplines, including:

1. **Environmental Engineering**: Develops sustainable infrastructure systems that minimize waste, promote ecological balance, and mitigate the effects of climate change.
2. **Urban Planning**: Designs resilient cities that integrate green spaces, sustainable transportation systems, and eco-friendly buildings.
3. **Biomedical Research**: Applies personalized medicine approaches to develop targeted therapies for maintaining ecosystem health and addressing rare genetic disorders.
4. **Materials Science**: Creates novel, sustainable materials that converge to develop adaptive, resilient systems.

**Conclusion**

EcoCycle represents a groundbreaking convergence of environmental sustainability, technological innovation, and holistic understanding of complex systems. By integrating insights from diverse disciplines, this novel framework provides a comprehensive methodology for developing sustainable infrastructure that minimizes waste, promotes ecological balance, and ensures long-term resilience. As a cross-domain approach, EcoCycle has the potential to transform various fields, driving innovation and breakthroughs in environmental sustainability, urban planning, biomedical research, and materials science.

**Supporting Text 3:** Multi-step compositional discovery report based on Graph 2. Steps A-C are conducted using `meta-llama/Llama-3.2-3B-Instruct`, and the Final Step D is conducted using `o1-pro`.

# Multi-Step Compositional Discovery

## Longest Path

Materials for infrastructure design → Biodegradable Microplastic Materials → Pollution mitigation → Self-healing Materials in Infrastructure Design → Development of novel materials for infrastructure design → Pollution Mitigation → Self-healing materials → Environmental Sustainability → Impact-Resistant Materials → Machine Learning (ML) Algorithms → Predictive Modeling → AI Techniques → Data Analysis → Knowledge Discovery → Personalized Medicine → Rare Genetic Disorders

## Node Definitions (Step A)

**Materials for infrastructure design**: **Definition:** Structural materials for infrastructure design serve as the fundamental components that provide support and durability to various structures such as bridges, buildings, and roads.

**Property/Feature:** High compressive strength, which enables structural materials to withstand significant loads without collapsing or deforming excessively.

**Biodegradable Microplastic Materials**: **Definition:** Biodegradable microplastic materials refer to synthetic polymers designed to break down naturally in the environment, typically through microbial action, reducing plastic waste and environmental pollution.

**Key Property: Degradability**: Biodegradable microplastic materials exhibit varying levels of degradation rates, which can influence their interaction with decomposing organisms, affecting nutrient cycling and ecosystem balance.

**Pollution mitigation**: **Definition:** Pollution Mitigation refers to the intentional actions and technologies used to reduce the amount of pollutants released into the environment, minimizing harm to ecosystems and human health.

**Interacting Property:** Interconnectedness - The implementation of pollution mitigation strategies often requires collaboration between multiple stakeholders, industries, and governments, highlighting the importance of interdisciplinary approaches in addressing this complex issue.

**Self-healing Materials in Infrastructure Design**: **Definition:** Self-healing materials are specially designed polymers, ceramics, or composites engineered to repair cracks and damages autonomously through the use of microcapsules containing healing agents.

**Property/Feature: Microcapsule-based Healing Mechanism**: The self-healing mechanism relies on the controlled release of healing agents from microcapsules, which can be triggered by changes in temperature, pH, or mechanical stress, allowing for potential integration with thermally responsive polymers.

**Development of novel materials for infrastructure design**: **Development of Novel Materials for Infrastructure Design**

Definition: The development of novel materials for infrastructure design involves the creation and synthesis of innovative materials with improved properties to enhance the structural integrity, sustainability, and resilience of built environments.

Property Feature: High strength-to-weight ratio - This property enables the material to support significant loads while minimizing its weight, making it suitable for applications where space is limited, such as in urban construction or transportation systems.

**Pollution Mitigation**: **Definition:** Pollution Mitigation refers to the intentional actions and technologies designed to reduce the adverse effects of pollutants on the environment, human health, and ecosystems. These efforts aim to minimize waste, emissions, and toxic substances in various settings.

**Interacting Property: Carbon Sequestration**, as pollution mitigation strategies often involve techniques to capture and store CO2, which can be integrated with carbon sequestration methods to enhance their effectiveness.

**Self-healing materials**: **Definition:** Self-healing materials are advanced composites capable of autonomously repairing cracks, damages, or degradation through the use of inherent healing mechanisms, such as microcapsules or polymers.

**Key Property: Thermoreversible Gelation**: The ability to reversibly form and break thermoreversible gels, which can encapsulate repair agents and release them upon exposure to heat or light, allowing self-healing processes.

**Environmental Sustainability**: **Definition:** Environmental sustainability refers to the long-term management and conservation of natural resources, minimizing human impact on the environment while promoting ecological balance and resilience. It encompasses various aspects, including energy, water, waste, and biodiversity.

**Key Property/Feature:** Interconnectedness - The concept of environmental sustainability relies heavily on interconnectedness among different ecosystems, species, and human activities, requiring holistic approaches to address environmental challenges.

**Impact-Resistant Materials**: **Definition:** Impact-Resistant Materials are engineered composites designed to absorb and dissipate mechanical energy upon impact, often used in applications where structural integrity is compromised by external forces.

**Interacting Property:** High toughness, allowing materials to deform plastically before failing, enabling potential interactions with concepts related to fracture mechanics, material failure, or energy absorption mechanisms.

**Machine Learning (ML) Algorithms**: **Definition:** Machine Learning (ML) algorithms are computational models that enable systems to learn from data and improve their performance on specific tasks through experience, without being explicitly programmed.

**Key Feature: Adaptability**, which allows ML algorithms to adjust to new data distributions and evolving patterns, making them highly versatile and applicable in various domains.

**Predictive Modeling**: **Predictive Modeling** Definition: A systematic approach to create mathematical representations of complex systems, enabling the forecast and optimization of future events based on historical data and patterns. Property/Feature: **Data Driven**, as it relies heavily on empirical evidence from large datasets to inform predictions and decision-making processes.

**AI Techniques**: Here's a list of AI techniques as building blocks:

**1. Supervised Learning** Definition: A machine learning approach where the model learns from labeled data to make predictions. Property/Feature: Can be combined with **Data Preprocessing** to enhance accuracy and robustness.

**2. Reinforcement Learning** Definition: A type of machine learning where an agent learns through trial and error by interacting with its environment. Property/Feature: Can be integrated with **Exploration Strategies** to balance exploration and exploitation.

**3. Neural Networks** Definition: A complex network of interconnected nodes (neurons) that process and transmit information. Property/Feature: Can be layered upon **Decision Trees** to create hybrid models.

**4. Clustering Algorithms** Definition: Methods used to group similar data points into clusters based on their characteristics. Property/Feature: Can be combined with **Dimensionality Reduction** techniques to reduce noise in high-dimensional datasets.

**5. Natural Language Processing (NLP)** Definition: The subfield of artificial intelligence focused on the interaction between computers and humans in natural language. Property/Feature: Can be paired with **Text Analysis** techniques to extract insights from unstructured text data.

**6. Decision Trees** Definition: A tree-like model that splits data into subsets based on features to predict outcomes. Property/Feature: Can be nested within **Neural Networks** to create ensemble models.

**7. Generative Adversarial Networks (GANs)** Definition: A type of deep learning model consisting of two neural networks that compete to generate new data samples. Property/Feature: Can be used to augment **Supervised Learning** models with generated data.

Let me know if you'd like more!

**Data Analysis**: **Data Analysis** Definition: The systematic examination of data to identify patterns, trends, and relationships within it. Property/Feature: **Interpretability**, which enables analysts to draw meaningful conclusions from complex datasets by extracting insights that can be communicated to others.

Example use case: A data analyst uses interpretability to inform a machine learning model's hyperparameter tuning, allowing the model to better align its predictions with real-world outcomes.

**Knowledge Discovery**: **Knowledge Discovery** A process of identifying patterns, relationships, and insights from data to inform decision-making and advance understanding in various fields. It involves the application of algorithms, statistical methods, and domain expertise to extract meaningful information from complex datasets.

**Interacting Property: Data**: Can be used as input for various machine learning models, enabling the discovery of knowledge through automated analysis and prediction.

Would you like me to define another concept?

**Personalized Medicine**: **Personalized Medicine** Definition: A medical approach tailored to an individual's unique genetic profile, lifestyle, and health status to provide targeted prevention, diagnosis, and treatment of diseases. Feature: Interacts well with 'Genomics' as it relies heavily on the analysis of specific genes and their variations to inform treatment decisions.

**Building Block: Epigenetics** Definition: The study of chemical modifications to DNA or histone proteins that can influence gene expression without altering the underlying DNA sequence. Feature: Interacts well with 'Gene Expression' as epigenetic changes can modulate gene activity, affecting how genes are turned on or off in response to environmental factors.

**Building Block: Synthetic Biology** Definition: An interdisciplinary field that designs and constructs new biological systems, such as microorganisms, to perform specific functions or produce desired products. Feature: Interacts well with 'Biotechnology' as synthetic biology often involves the use of biotechnological tools and techniques to engineer novel biological pathways and organisms.

**Building Block: Metagenomics** Definition: The study of microbial communities through the analysis of their genomic material, without prior knowledge of their composition or function. Feature: Interacts well with 'Microbiome Research' as metagenomics provides insights into the complex interactions between microorganisms and their environment, shedding light on the role of microbes in human health and disease.

**Building Block: CRISPR Gene Editing** Definition: A precise tool for editing genes by making double-stranded breaks in DNA, which can be repaired in a targeted manner to introduce specific changes. Feature: Interacts well with 'Stem Cell Therapy' as CRISPR can be used to modify stem cells, allowing for the development of novel therapies for regenerative medicine and tissue engineering applications.

**Rare Genetic Disorders**: **Definition:** Rare Genetic Disorders refer to conditions caused by mutations in specific genes or genetic variants that affect an individual's development, function, and/or survival, often resulting from inherited traits or environmental factors.

**Property/Feature: Genetic Heterozygosity**: Individuals with rare genetic disorders may exhibit genetic heterozygosity, where they possess two different alleles of the mutated gene, leading to variable expression of symptoms and potential interactions with other genetic variants.

## Pairwise Synergies (Step B)

**Materials for infrastructure design → Biodegradable Microplastic Materials** (IS-A): Compositional Synergy: "Eco-Resilient Infrastructure Design"

Joining Building Blocks A and B, we propose an innovative approach to infrastructure design that incorporates biodegradable microplastics into structural materials. This synergy enables the creation of self-healing, adaptive, and sustainable infrastructure systems.

**Composition:**

- Use biodegradable microplastics as reinforcing fibers or fillers in concrete, enhancing its compressive strength and durability.
- Incorporate microorganisms that accelerate biodegradation, allowing the material to degrade at a controlled rate, releasing nutrients and promoting ecosystem health.
- Design infrastructure systems that integrate green spaces, vegetation, and water management systems to mitigate urban heat islands and promote biodiversity.

**Principle:**

The "Bio-Durability Nexus" principle generalizes this synergy by stating that the integration of biodegradable materials, living organisms, and technological innovations can create resilient, adaptable, and sustainable systems that optimize resource efficiency, minimize waste, and promote ecological balance.

This principle can be applied to various scientific domains, such as:

- Environmental engineering: Developing bioremediation technologies to restore ecosystems.
- Urban planning: Creating green infrastructure that mitigates climate change and promotes urban resilience.
- Medical research: Investigating the use of biodegradable materials in tissue engineering and regenerative medicine.

By embracing the Bio-Durability Nexus, we can foster a holistic understanding of complex systems and develop innovative solutions that harmonize human needs with environmental sustainability.

**Biodegradable Microplastic Materials → Pollution mitigation** (RELATES-TO): **Compositional Synergy:**

Biodegradable microplastic materials (Block A) can be integrated with pollution mitigation strategies (Block B) to create a novel approach for sustainable waste management.

**Composition:** Implement biodegradable microplastics as an eco-friendly alternative for traditional plastics in packaging, textiles, and disposable products. This will not only reduce plastic waste but also provide a means for pollution mitigation by decreasing the amount of microplastics entering waterways.

**Proposed Principle:**

**Synergetic Co-Creation**

This principle proposes that combining seemingly disparate elements from different domains can lead to innovative solutions by leveraging the strengths of each component. By co-creating systems that integrate diverse elements, researchers and practitioners can develop novel approaches that address complex problems more effectively than isolated efforts.

**Application Potential:**

This synergetic co-creation principle can be applied to various scientific domains, such as:

- Environmental science: Combining biotechnology and ecology to develop more effective methods for pollution remediation and waste management.
- Healthcare: Integrating nanotechnology and biomaterials to create more efficient medical devices and treatments.
- Energy: Merging renewable energy sources and advanced materials to enhance energy storage and conversion efficiency.

By embracing synergetic co-creation, scientists and innovators can unlock new possibilities for addressing pressing global challenges.

**Pollution mitigation → Self-healing Materials in Infrastructure Design** (RELATES-TO): **Compositional Synergy:**

Title: "Smart Infrastructure for Sustainable Ecosystems"

Building Blocks A and B come together to create a novel approach that integrates pollution mitigation and self-healing materials in infrastructure design.

Concept: Develop intelligent, adaptive infrastructure systems that utilize self-healing materials to repair damage caused by environmental factors, such as pollution-induced degradation. These systems would incorporate sensors and monitoring technology to detect early signs of deterioration, triggering the release of healing agents from microcapsules embedded within the infrastructure.

**Principle:**

"Self-Sustaining Resilience in Complex Systems (SSRS)"

This principle posits that integrated, adaptive systems can achieve resilience and sustainability by leveraging interconnected components that respond to changing conditions, fostering a dynamic equilibrium between system performance and environmental impact. By applying SSRS, complex systems can become more resilient to disturbances, adapt to new challenges, and ultimately minimize harm to ecosystems and human health.

**Applications:**

- Smart grids and energy infrastructure
- Water treatment facilities and wastewater management
- Transportation systems and urban planning
- Environmental remediation and ecosystem restoration

By combining building blocks A and B, we can create innovative solutions that not only mitigate pollution but also foster sustainable, resilient ecosystems.

**Self-healing Materials in Infrastructure Design → Development of novel materials for infrastructure design** (INFLUENCES): **Compositional Synergy: "Smart Repairable Infrastructure"**

Combine Self-Healing Materials (Block A) with Novel Materials for Infrastructure Design (Block B):

- Develop self-healing concrete using thermally responsive polymers infused with microcapsules containing healing agents (Block A).
- Integrate this material into novel, high-strength-to-weight ratio composite structures (Block B) for urban construction, transportation systems, or disaster-resistant buildings.

**Proposed Principle: "Autonomous Resilience Integration"**

Generalize this synergy by proposing the following principle:

"Autonomous Resilience Integration (ARI): Designing complex systems that integrate autonomous repair mechanisms with adaptive, optimized material properties to achieve enhanced resilience, sustainability, and performance under various environmental conditions."

This principle can be applied to other scientific domains, such as robotics, biomedical engineering, or energy harvesting, where autonomous self-repair and adaptability are crucial for achieving optimal performance and resilience.

**Development of novel materials for infrastructure design → Pollution Mitigation** (INFLUENCES): **Compositional Synergy: Sustainable Infrastructure Development through Material Innovation and Pollution Reduction**

By integrating the development of novel materials for infrastructure design (Block A) with pollution mitigation strategies (Block B), we can create sustainable infrastructure systems that not only minimize environmental impact but also promote long-term durability and performance.

**Composition:**

- Develop novel materials for infrastructure design (e.g., self-healing concrete, smart pavement) that integrate carbon sequestration capabilities, reducing emissions and waste.
- Implement pollution mitigation technologies (e.g., air quality monitoring, wastewater treatment) within the infrastructure design process.
- Use data analytics and AI to optimize material selection, construction processes, and maintenance strategies, ensuring minimal environmental footprint throughout the lifecycle of the infrastructure.

**Generalized Principle:**

**"Resilient Systems Synthesis":**

This principle posits that the integration of diverse disciplinary approaches can lead to the creation of more resilient, adaptable, and sustainable systems. By combining seemingly disparate elements from different fields, researchers and practitioners can develop innovative solutions that address complex problems at multiple scales.

**Applicability:**

The Resilient Systems Synthesis principle can be applied to various scientific domains, including:

- Environmental engineering: Integrating pollution mitigation strategies with renewable energy systems to create net-positive energy infrastructure.
- Biomedical research: Combining biomaterials development with precision medicine approaches to create personalized healthcare systems.

- Urban planning: Fusing sustainable transportation systems with green infrastructure to create resilient and thriving cities.

By embracing this principle, scientists, engineers, and policymakers can foster collaborative innovation and drive the development of more sustainable, equitable, and resilient systems across various disciplines.

**Pollution Mitigation → Self-healing materials** (RELATES-TO): **Composition Synergy:**

Combining Pollution Mitigation (A) with Self-healing Materials (B), we propose a novel approach: **Environmental Self-healing Systems (E-SHS)**.

In E-SHS, self-healing materials are engineered to not only repair cracks and damages but also actively mitigate environmental pollutants. This is achieved by incorporating thermoreversible gelation properties into pollution mitigation strategies.

The composition synergy enables pollutants like oil spills or chemical leaks to be encapsulated within thermoreversible gels, which then break down and dissipate harmlessly when exposed to heat or light. Simultaneously, the self-healing material repairs any damage caused by the pollutant's interaction with its surroundings.

**Generalized Principle:**

This synergy can be generalized as the **"Healing-Gating Principle"**: By integrating intrinsic self-healing mechanisms with adaptive control systems, complex systems can adaptively respond to disturbances and repair themselves while minimizing further damage. This principle can be applied to various domains, including robotics, biomedical engineering, and environmental sustainability.

By embracing the Healing-Gating Principle, scientists can develop more resilient, adaptive, and sustainable solutions for tackling complex problems across diverse fields.

**Self-healing materials → Environmental Sustainability** (RELATES-TO): **Compositional Synergy:**

Integrating Self-Healing Materials (A) with Environmental Sustainability (B) yields a novel approach to "Eco-Repair" systems.

**Concept:** Eco-Repair Systems harness the self-healing capabilities of advanced materials to restore degraded ecosystems, fostering environmental sustainability.

**Mechanism:**

1. Thermoreversible gelation-based self-healing materials are used to create durable, biocompatible barriers for damaged ecosystems.
2. These barriers are designed to respond to environmental stressors, releasing repair agents to stimulate local recovery.
3. By monitoring ecosystem health, the system adapts its healing strategy, ensuring efficient resource allocation and minimizing unintended consequences.

**Generalized Principle:**

The compositional synergy between Self-Healing Materials and Environmental Sustainability can be generalized as follows:

"The adaptive integration of autonomous repair mechanisms into complex systems can promote resilience, stability, and symbiotic relationships within intricate networks, thereby enhancing overall environmental sustainability."

**Environmental Sustainability → Impact-Resistant Materials** (INFLUENCES): Compositional Synergy:

Introducing "Eco-Toughened Materials" - A novel approach that combines the principles of Environmental Sustainability (A) and Impact-Resistant Materials (B).

**Concept:** Eco-Toughened Materials are engineered composites designed to not only withstand environmental impacts but also promote ecological balance and minimize harm to ecosystems.

**Composition:**

- Sustainable raw materials from renewable sources (e.g., plant-based polymers)

- Impact-resistant polymer matrices infused with eco-friendly additives (e.g., recycled plastics, biodegradable fillers)

**Properties:**

- Enhanced toughness and durability
- Reduced environmental footprint
- Improved recyclability and upcycling capabilities

**Principle:** Generalizable to other domains, this synergy can be formulated as:

**"Material Resilience + Ecological Responsiveness = Adaptive Ecosystems"**

This principle highlights the importance of integrating material properties with ecological considerations to create adaptive systems that mitigate environmental impacts while maintaining their functionality and performance.

**Impact-Resistant Materials → Machine Learning (ML) Algorithms** (RELATES-TO): **Compositional Synergy:**

By combining Impact-Resistant Materials (A) and Machine Learning (ML) Algorithms (B), we can develop **Damage Forecasting Systems** for critical infrastructure, such as bridges, buildings, or pipelines. These systems utilize ML algorithms to analyze sensor data and predict the likelihood of damage due to environmental factors, operational stress, or potential impacts. The predicted damage scenarios inform maintenance schedules, reducing downtime and ensuring structural integrity.

**Principle:**

The proposed principle, "Synergistic Resilience," posits that by integrating adaptive, data-driven approaches (e.g., ML) with robust, impact-resistant designs (e.g., A), complex systems can achieve enhanced resilience against unforeseen events. This synergy enables systems to adapt to changing conditions, predict potential failures, and respond proactively to mitigate damage and ensure continued functionality.

This principle can be generalized to other scientific domains, such as:

- Predictive Maintenance: Integrating sensor data analysis (ML) with durable design principles (A) to anticipate equipment failures and schedule necessary repairs.
- Climate Change Adaptation: Combining climate modeling (ML) with sustainable, impact-resistant construction techniques (A) to build resilient infrastructure that can withstand projected climate-related stresses.

**Machine Learning (ML) Algorithms → Predictive Modeling** (RELATES-TO): **Compositional Synergy:**

Combine Building Blocks A (Machine Learning Algorithms) and B (Predictive Modeling) to form **Explainable Predictive Models (EPMs)**.

**Concept:** EPMs leverage machine learning algorithms to create predictive models that provide transparent and interpretable results, enabling data-driven decision-making by providing insights into the underlying mechanisms and relationships.

**Composition:**

1. Use machine learning algorithms (Block A) to develop predictive models (Block B) that forecast future outcomes based on historical data.
2. Implement techniques like feature importance, partial dependence plots, and SHAP values to provide interpretability and explainability of model outputs.

**Proposed Principle:**

**"Synergy via Analogical Mapping":**

This principle proposes that combining seemingly disparate concepts from different domains can lead to innovative solutions by leveraging analogical thinking. By mapping analogous structures, properties, or principles between domains, researchers can identify novel connections and create new approaches that transcend traditional boundaries.

Applicability:

- This principle can be applied to various scientific domains, such as physics and biology, where combining concepts from different fields can lead to breakthroughs in understanding complex phenomena.
- It can also be used in interdisciplinary research, where integrating ideas from multiple disciplines can foster creative problem-solving and innovation.

**Predictive Modeling → AI Techniques** (RELATES-TO): **Compositional Synergy:**

Combine Predictive Modeling (A) with AI Techniques (B) to develop **AI-Driven Predictive Systems**.

- Integrate Supervised Learning and Decision Trees to create accurate predictions (Combines Data-Driven properties)
- Use Reinforcement Learning and Exploration Strategies to optimize decision-making processes (Combines trial-and-error learning with analytical thinking)
- Leverage Neural Networks and Clustering Algorithms to improve model robustness and adaptability (Leverages hierarchical structure for better feature representation)
- Apply NLP and Text Analysis to extract valuable insights from unstructured data (Leverages human-computer interaction)

This synergy enables predictive modeling to become more intelligent, adaptive, and informed by real-world experiences.

**Proposed Principle:**

"**Hybridization**: Combining diverse disciplines and techniques to create novel, adaptive, and context-sensitive solutions by leveraging the strengths of each component."

This principle generalizes the synergy between Predictive Modeling and AI Techniques, suggesting that other scientific domains or situations can benefit from integrating complementary approaches to achieve more effective and efficient results.

**AI Techniques → Data Analysis** (RELATES-TO): **Compositional Synergy:**

By combining building block (A) - AI Techniques with building block (B) - Data Analysis, we can develop **Explainable Machine Learning (XML)** approaches.

XML aims to bridge the gap between the complexity of machine learning models and human understanding, ensuring that insights are interpretable and actionable.

**Composition Example:**

1. Utilize supervised learning (block A) to train a machine learning model on annotated data.
2. Apply clustering algorithms (block A) to group similar data points, then integrate dimensionality reduction techniques (block A) to reduce noise.
3. Feed the preprocessed data into a decision tree (block A), which is then nested within a neural network (block A) for enhanced prediction accuracy.
4. Employ natural language processing (block A) to analyze unstructured text data, and pair it with text analysis techniques (block B) to extract meaningful insights.
5. Use reinforcement learning (block A) to balance exploration and exploitation during hyperparameter tuning, guided by interpretability (block B).

**Principle:**

The synergy between AI Techniques and Data Analysis can be generalized to **Scientific Discovery through Hybridization** (SDH). This principle involves combining distinct scientific disciplines to foster innovative approaches that leverage the strengths of each domain.

SDH encourages researchers to adopt a multidisciplinary mindset, integrating concepts from physics, biology, mathematics, and computer science to tackle complex problems and drive breakthroughs.

By embracing SDH, scientists can develop novel theories, models, and methods that push the boundaries of human knowledge and understanding.

**Data Analysis → Knowledge Discovery** (IS-A): I'd like to propose combining the two building blocks into a new concept: **Explainable Insights**

Definition: The systematic examination of data to derive interpretable knowledge that informs decision-making and advances understanding in various fields.

Property/Feature: By leveraging explainability and knowledge discovery, Explainable Insights enable analysts to distill complex data into actionable recommendations that are grounded in transparent and understandable reasoning.

Composition:

- Building block (A): Data Analysis provides the foundation for examining data.
- Building block (B): Knowledge Discovery applies algorithms and statistical methods to uncover hidden patterns and relationships.

Synergy: By integrating explainability into the knowledge discovery process, Explainable Insights empower analysts to communicate complex findings effectively, ensuring that insights are both actionable and trustworthy.

Proposed Principle: The principle of "Transparency-Driven Discovery" states that any scientific approach should prioritize transparency in data analysis and interpretation, thereby fostering trustworthiness and reliability in the derived insights.

This principle can be applied to other domains where data-driven decision-making is critical, such as policy development, public health, or business strategy.

**Knowledge Discovery → Personalized Medicine** (INFLUENCES): **Composition of Building Blocks**

Combining "Knowledge Discovery" (A) with "Personalized Medicine" (B), we can create a new concept: **Precision Medicine Informatics**.

Precision Medicine Informatics leverages advanced analytical techniques and computational models to integrate vast amounts of data, including genomics, transcriptomics, and clinical information, to create personalized treatment plans.

This integration enables healthcare professionals to make informed decisions about patient care, optimizing treatment outcomes and improving overall health.

**Proposed Principle: Synergy through Data-Driven Integration**

The proposed principle is: **"Synergy emerges when diverse disciplines converge through data-driven integration, enabling the creation of novel, context-dependent solutions."**

This principle highlights the potential for synergy across seemingly disparate domains when they are united by a shared focus on data-driven integration. By recognizing the power of converging diverse perspectives and approaches, scientists and practitioners can unlock innovative solutions, foster interdisciplinary collaboration, and drive progress in various fields.

**Personalized Medicine → Rare Genetic Disorders** (RELATES-TO): **Composition of Building Blocks**

Combining the building blocks (A) and (B), we can propose a new concept:

**Precision Medicine for Rare Genetic Disorders**

This approach integrates personalized medicine, epigenetics, synthetic biology, metagenomics, and CRISPR gene editing to develop targeted treatments for rare genetic disorders. By analyzing an individual's unique genetic profile, lifestyle, and health status using genomics and epigenetics, clinicians can identify the root causes of the disorder. This information is then used to design and construct novel biological systems, such as microorganisms, to produce therapeutic compounds or repair damaged tissues. Additionally, metagenomics and microbiome research provide insights into the complex interactions between microorganisms and their environment, allowing for a more comprehensive understanding of the disorder's mechanisms.

**Proposed Principle: Synergetic Adaptation**

The compositional synergy between the building blocks can be generalized into a principle:

**Synergetic Adaptation**: When combining diverse approaches and disciplines, adaptability becomes a key factor in achieving successful outcomes. By embracing complexity, uncertainty, and interdependencies, scientists can create novel solutions that synergistically integrate disparate elements, fostering innovation and breakthroughs in various fields.

This principle can be applied to other scientific domains, such as:

- Integrating machine learning and human-centered design to improve user experience
- Combining materials science and social sciences to develop sustainable technologies
- Fusing computer vision and neuroscience to advance our understanding of cognitive processes

By embracing synergetic adaptation, scientists can overcome the limitations of traditional disciplinary boundaries and unlock innovative solutions that transform various fields.

## Bridge Synergies (Step C)

After analyzing the provided synergy statements, I have identified three central themes or patterns:

**Theme 1: Intersection of Environmental Sustainability and Technological Innovation**

This theme revolves around the combination of environmentally conscious approaches with cutting-edge technologies to create more sustainable solutions.

**Bridge Synergy 1** The intersection of environmental sustainability and technological innovation can be strengthened by exploring the synergies between biodegradable microplastic materials and self-healing materials. By developing materials that can repair and adapt to environmental changes, we can create infrastructure systems that not only minimize waste but also promote ecological balance. This, in turn, enables the development of more resilient and sustainable ecosystems, ultimately leading to a reduced carbon footprint. Furthermore, the application of machine learning and AI techniques can help optimize the design and deployment of these materials, ensuring their effectiveness in real-world applications.

**Theme 2: Holistic Understanding of Complex Systems**

This theme emphasizes the importance of considering multiple factors and perspectives when designing and developing complex systems, leading to a more comprehensive understanding of their behavior and performance.

**Bridge Synergy 2** The synergy between predictive modeling and AI techniques can be reinforced by integrating explainable insights into the knowledge discovery process. By focusing on transparency and interpretability, we can create more reliable and trustworthy models that capture the complexities of real-world systems. This, in turn, enables a more holistic understanding of these systems, allowing us to identify potential vulnerabilities and optimize their performance. Moreover, the application of hybridization principles can facilitate the convergence of diverse disciplines, fostering innovative solutions that consider multiple factors and perspectives.

**Theme 3: Convergence of Diverse Disciplines**

This theme highlights the potential benefits of combining insights and approaches from multiple scientific domains to drive breakthroughs and innovation.

**Bridge Synergy 3** The synergy between knowledge discovery and personalized medicine can be strengthened by exploring the intersection of genomics, epigenetics, and synthetic biology. By leveraging advanced analytical techniques and computational models, we can develop novel, context-dependent solutions that tailor treatment plans to individual patients. This, in turn, enables the creation of more effective and targeted therapies, ultimately leading to improved patient outcomes. Furthermore, the application of synergetic adaptation principles can facilitate the convergence of diverse disciplines, fostering innovative solutions that consider multiple factors and perspectives.

## Expanded Final Discovery (Step D)

Below is a **newly imagined concept** that weaves together the core ideas, synergies, and bridging themes presented so far. Rather than merely re-stating prior synergy statements, it **extends** them into a higher-level, integrated framework.

---

### Introducing the Bio-Adaptive Resilient Infrastructure (BARI) System

**High-Level Vision:**
BARI is an infrastructure ecosystem that unites **biodegradable materials, self-healing composites, pollution mitigation**,

and **AI-driven predictive modeling** to create structures capable of **actively adapting** to environmental conditions while minimizing ecological impact. It goes beyond traditional "smart infrastructure" by integrating living biological processes (e.g., engineered microbes) with advanced materials and real-time machine learning feedback loops.

## 1. Core Building Blocks and How They Contribute

1. **Biodegradable Microplastic Materials**
   - **Role in BARI:** Act as *temporary reinforcement* or fillers in structural composites. When a certain threshold of degradation is reached, these microplastics naturally break down, releasing nutrients that can support local microbial consortia (used for pollution mitigation or self-healing actions).
2. **Self-Healing Materials**
   - **Role in BARI:** Provide *autonomous repair mechanisms*. Microcapsules containing healing agents respond to mechanical stress or changes in pH/temperature. By embedding biodegradable fibers alongside these microcapsules, the structure can "refresh" itself without producing permanent plastic residues.
3. **Pollution Mitigation Technologies**
   - **Role in BARI:** Sensors and catalytic surfaces embedded in the infrastructure detect and neutralize pollutants (e.g., $CO_2$ capture coatings or bio-remediating microorganisms). These same microorganisms can be harnessed to trigger self-healing or accelerate the breakdown of the biodegradable microplastics.
4. **Environmental Sustainability**
   - **Role in BARI:** Frames the *overall system goals* of reduced carbon footprint, minimized waste, and ecosystem health. Design criteria require that any intervention (from selecting raw materials to planning maintenance) upholds ecological balance.
5. **Impact-Resistant Materials**
   - **Role in BARI:** Strengthen critical components subject to extreme forces (e.g., load-bearing beams in earthquake-prone regions). By merging impact-resistance with self-healing, BARI can withstand acute shocks and *recover* without requiring extensive manual repairs.
6. **Machine Learning (ML) Algorithms**
   - **Role in BARI:** Power the *intelligence layer*. ML models process continuous data from sensors (stress, strain, air quality, moisture content, etc.) to predict imminent material fatigue or pollutant build-up. This triggers targeted interventions—such as activating the self-healing sequence, adjusting microbial consortia, or scheduling maintenance.
7. **Predictive Modeling & AI Techniques**
   - **Role in BARI:** Elevate ML by including *explainable and interpretable* modeling (e.g., "Explainable Predictive Models"). Decision trees, neural networks, and clustering methods help identify patterns in structural performance. Transparent feedback loops foster trust among engineers, policymakers, and the public.
8. **Data Analysis & Knowledge Discovery**
   - **Role in BARI:** Uncover deeper trends from operational data across multiple sites (e.g., comparing different climates or stress profiles). Over time, cross-site insights inform *adaptive design improvements*, ensuring that the system evolves with each new deployment.
9. **Personalized Medicine & Rare Genetic Disorders** (Cross-Domain Insight)
   - **Role in BARI (Indirect):** Techniques from personalized medicine (e.g., CRISPR-based specificity and *adaptive biological interventions*) inspire the use of *engineered microbes* that precisely target pollutants or micro-cracks. The principle of tailoring treatments to a *patient's unique genetic make-up* parallels customizing microbial consortia to a region's unique ecosystem.

---

## 2. How the Synergy Statements Build on Each Other

1. **From Biodegradable Plastics + Self-Healing**
   Bridging Pattern #1 highlighted the intersection of **environmental sustainability and technological innovation**. BARI **goes further** by incorporating biodegradable fibers that don't merely vanish after use but actively **fuel** microbial or chemical healing processes.
2. **From Predictive Modeling + AI**
   Bridging Pattern #2 noted the **holistic understanding of complex systems** achieved through data-driven insights. BARI **extends** this with a *layered architecture* of sensors and ML models that not only forecast damage or pollution but also *autonomously coordinate* healing events. This ensures real-time adaptation rather than post-hoc intervention.
3. **From Convergence of Diverse Disciplines**
   Bridging Pattern #3 identified how **knowledge discovery** in domains like personalized medicine can spark breakthroughs elsewhere. In BARI, we borrow advanced *bio-engineering* ideas (e.g., CRISPR-like precision) to cultivate **tailored microbial colonies** for pollutant breakdown and microcrack repair, integrating them seamlessly into the material's design.

### 3. Mechanics of the BARI System

1. **Layered Material Design**
   - **Outer Layer:** Impact-resistant composite that tolerates weather extremes and mechanical shocks.
   - **Middle Layer:** Self-healing resin with microcapsules containing healing agents; also includes *biodegradable microplastic fillers* that degrade when triggered.
   - **Inner Core:** Reservoir for microbial cultures specialized in **pollution mitigation** or additional sealing/repair tasks (analogous to "bio-foam" in advanced concretes).
2. **Dynamic Sensor Grid & Data Pipeline**
   - Deployed sensors track humidity, pollutant levels, pH, strain, and temperature.
   - ML-driven analytics (running on edge devices or a central hub) **predict** when/where microcracks are likely to form, or when pollutant concentrations risk harming the material.
3. **Adaptive Healing & Degradation Process**
   - When the system forecasts a critical stress threshold (via predictive modeling), it sends a command to **thermoreversible gel pockets** or microcapsules.
   - These pockets heat up (or shift pH) to release healing agents; simultaneously, specific sections of the biodegradable microplastics break down, feeding beneficial microbes that accelerate sealing.
4. **Pollutant Capture & Carbon Sequestration**
   - **Embedded catalytic surfaces** or engineered microorganisms convert $CO_2$ or other pollutants into stable, harmless by-products.
   - By tying these processes into the AI monitoring platform, BARI can adjust microbial consortia in different environmental conditions (e.g., higher nitrogen pollution vs. carbon pollution).
5. **Lifecycle Intelligence & Continuous Improvement**
   - All sensor data feeds into an *explainable knowledge discovery pipeline*, which flags patterns in structural fatigue, local climate conditions, or pollutant sources.
   - Updated models improve the *next generation* of BARI designs, effectively *learning* from each deployment site.

---

### 4. A Cross-Domain Application Example

**"Bio-Responsive Protective Barriers" in Coastal Areas**

- **Context:** Coastal zones face harsh saltwater corrosion, heavy storm impacts, and rising pollution from marine microplastics.
- **BARI Application:**
  - **Hybrid Material**: Construct sea walls using the *layered approach* (impact-resistant outer shell + biodegradable microplastics + self-healing resin).
  - **Engineered Microbes**: Introduce salt-tolerant strains that *metabolize* microplastics on a controlled schedule, reinforcing the self-healing matrix when cracks form due to wave impacts.
  - **AI Monitoring**: Predict *storm surge damage* and coordinate early release of healing agents, preventing catastrophic failures.
  - **Pollution Mitigation**: Same microbial strains break down additional marine pollutants, *cleaning* adjacent waters in tandem with structural maintenance.

Here, the synergy of **infrastructure resilience**, **pollution mitigation**, **and advanced AI** forms a self-regulating barrier that *both* protects the coast *and* contributes to healthier local marine ecosystems.

---

## Why BARI is Novel

- **Deep Bio-Integration**
  Prior synergy statements discussed using biodegradable plastics or self-healing materials. BARI **elevates** this to a system-level design, wherein *engineered biological processes are integral* to both pollution mitigation and structural self-repair.
- **Adaptive, AI-Guided Maintenance**
  While predictive modeling for damage or pollution has been referenced, BARI integrates real-time sensor feedback with *automated triggers* for healing or degradation, rather than relying on static intervals or manual inspections.

- **Cross-Pollination from Personalized Medicine**
  By applying the notion of *tailored interventions* from rare genetic disorders (where each patient's genome determines treatment), BARI similarly customizes the *microbial consortia and healing chemistry* for each infrastructure site's environmental "genetic" profile (e.g., local climate, pollutant types, usage patterns).
- **Holistic Lifecycle Approach**
  Beyond construction, BARI systems evolve *through continuous data gathering*, feeding back into new designs and improved microbial or polymer formulations. This mirrors how knowledge discovery fosters iterative progress in scientific fields.

---

## Conclusion

The **Bio-Adaptive Resilient Infrastructure (BARI)** concept offers a **multi-layered, bio-inspired system** that not only meets modern engineering challenges—load-bearing strength, impact resistance, durability—but also embraces **sustainability and adaptability**. By **blending biodegradable fibers, self-healing materials, pollution-mitigating microbes, and AI-driven predictive modeling**, BARI addresses:

- **Environmental Sustainability**: Minimizing waste and actively cleaning pollutants.
- **Structural Resilience**: Self-repair capabilities under dynamic and harsh conditions.
- **Holistic Complexity**: Intelligent, data-informed feedback loops that help the system adapt to changing environments or usage patterns.
- **Cross-Domain Innovation**: Drawing inspiration from personalized medicine to tailor microbial processes for each unique ecosystem.

In this way, BARI **embodies** the bridging themes—combining environmental responsibility with technological innovation, fostering a holistic understanding of infrastructure as an interconnected system, and converging insights from multiple disciplines (materials science, AI, biology, and more) to forge **truly groundbreaking solutions**.

**Supporting Text 4:** Research proposal developed by SciAgents, based on Graph 2. Slight edits were done for formatting, including highlighting key parts. Sections that relate to the proposed "adaptive learning cycle and feedback loop" are highlighted in red.

**TASK:**

Create a research idea around impact resistant materials and resilience. Rate the novelty and feasibility in the end.

**RESPONSE:**

```
Impact Resistant Materials -- IS-A -- Materials -- IS-A -- Impact-Resistant
Materials -- INFLUENCES -- Modular Infrastructure Systems -- RELATES-TO -- Self-
Healing Materials -- RELATES-TO -- Long-term Sustainability and Environmental
Footprint of Infrastructure -- RELATES-TO -- Self-Healing Materials -- RELATES-TO --
Infrastructure -- IS-A -- Infrastructure Resilience -- RELATES-TO -- Smart
Infrastructure -- RELATES-TO -- Impact-Resistant Materials -- RELATES-TO -- Machine
Learning Algorithms -- RELATES-TO -- Impact-Resistant Materials -- RELATES-TO --
Resilience
```

**Definitions**:

• Impact Resistant Materials: Specialized materials engineered to endure high-energy impacts and reduce damage, making them crucial for safety and durability in various applications.
• Materials: The physical substances used in the creation and construction of objects or structures with varying properties, including strength, flexibility, and thermal stability.
• Modular Infrastructure Systems: Systems designed with standardized, interchangeable components that enable flexible, scalable, and rapid construction of infrastructure, often enhancing maintainability and resilience.
• Self-Healing Materials: Advanced materials that possess the ability to autonomously repair micro-damage, thereby extending their service life and reducing the need for maintenance.
• Long-term Sustainability and Environmental Footprint of Infrastructure: A measure of how infrastructure impacts environmental resources and ecosystems over its lifespan, incorporating energy efficiency, resource conservation, and ecological compatibility.
• Infrastructure: The fundamental physical systems and structures (such as transportation networks, utilities, and buildings) that support societal functions.
• Infrastructure Resilience: The capacity of infrastructure to withstand, adapt, and recover quickly from adverse events such as natural disasters, ensuring continuous functionality.
• Smart Infrastructure: Infrastructure enhanced with digital technologies like sensors and IoT devices that enable real-time monitoring, data-driven decision-making, and improved operational performance.
• Machine Learning Algorithms: Computational methods that enable systems to learn from data, identify patterns, and make informed predictions or decisions, often used for optimization in engineering and design.
• Resilience: The ability of a system, material, or structure to absorb disturbances, adapt to change, and recover effectively from disruptions.

## Relationships:

• Impact Resistant Materials -- IS-A -- Materials: This relationship classifies impact resistant materials as a subtype of materials, emphasizing their foundational basis in material science and their specialized performance under impact conditions.
• Materials -- IS-A -- Impact Resistant Materials: Although this appears circular, it reinforces the conceptual categorization, highlighting that among all materials, those designed for impact resistance emerge as a distinct category due to their enhanced protective attributes.
• Impact Resistant Materials -- INFLUENCES -- Modular Infrastructure Systems: This indicates that the properties of impact resistant materials can shape the design and performance of modular infrastructure systems, contributing to safer, more durable, and adaptable structures.
• Modular Infrastructure Systems -- RELATES-TO -- Self-Healing Materials: Modular systems may integrate

self-healing materials to further their longevity and ease of maintenance, thereby improving the overall system's sustainability and reducing long-term costs.
• Self-Healing Materials -- RELATES-TO -- Long-term Sustainability and Environmental Footprint of Infrastructure: The integration of self-healing materials can enhance sustainability by lowering maintenance requirements and resource consumption, ultimately reducing the environmental footprint throughout the infrastructure's lifespan.
• Long-term Sustainability and Environmental Footprint of Infrastructure -- RELATES-TO -- Self-Healing Materials: This reciprocal relationship emphasizes that sustainable infrastructure considerations can drive the adoption of self-healing materials, as both aim to achieve longevity and ecological viability.
• Self-Healing Materials -- RELATES-TO -- Infrastructure: Self-healing capabilities directly bolster infrastructure by providing resilience against minor damages and extending the useful life of structural components.
• Infrastructure -- IS-A -- Infrastructure Resilience: This classification suggests that resilient infrastructure is an evolved subset of conventional infrastructure, designed specifically to meet the demands of disruption and recovery.
• Infrastructure Resilience -- RELATES-TO -- Smart Infrastructure: The connection highlights that smart technologies (such as sensors and automated control systems) can enhance infrastructure resilience through proactive monitoring, efficient resource allocation, and rapid response to anomalies.
• Smart Infrastructure -- RELATES-TO -- Impact Resistant Materials: Smart infrastructure can incorporate impact resistant materials to protect physical systems from damages, thereby integrating advanced material science into intelligent design for improved safety and performance.
• Impact Resistant Materials -- RELATES-TO -- Machine Learning Algorithms: This relationship underscores the role of machine learning in optimizing the design, testing, and predictive maintenance of impact resistant materials, leading to innovations in how these materials are developed and applied in real-world scenarios.
• Machine Learning Algorithms -- RELATES-TO -- Impact Resistant Materials: As a reciprocal connection, machine learning not only aids in material design but also benefits from the performance data gathered from impact resistant materials, creating a feedback loop for continuous improvement.
• Impact Resistant Materials -- RELATES-TO -- Resilience: This direct link reinforces that the use of impact resistant materials is instrumental in enhancing overall resilience, as they enable structures and systems to better withstand and recover from physical shocks and stressors.

**Hypothesis**: Integrating machine learning-optimized self-healing microcapsules into a novel carbon nanotube reinforced polymer matrix will yield a composite material that not only exhibits superior impact resistance but also self-healing capabilities, thereby significantly enhancing infrastructure resilience and sustainability. Specifically, we hypothesize that by embedding these microcapsules and smart sensors within a material tailored for modular infrastructure systems, it is possible to achieve a composite that can autonomously detect, react to, and repair micro-damages induced by impact events. <span style="color:red">Machine learning algorithms will optimize the dispersion, healing agent concentration, and sensor feedback loops to dynamically adjust the internal stress distributions, predicting and counteracting damage propagation.</span> The resulting material is expected to reduce the environmental footprint associated with maintenance and repairs, while continuously adapting to real-time stress conditions, thus providing a multi-scale solution for infrastructure resilience.",

**Outcome**: The research is expected to yield a composite material with a 50% improvement in impact resistance compared to conventional materials such as high-strength polyethylene composites. Laboratory testing should show self-healing efficiency recovering up to 80% of mechanical properties after microcrack formation. In quantitative terms, initial impact energy absorption could surpass 200 J/m² and yield strain improvements of approximately 30% over standard materials. Additionally, when integrated into modular infrastructure systems, the smart composite is projected to extend structural lifetime by 50% and reduce the overall maintenance cycle frequency by roughly 30%, thus substantially lowering the lifecycle environmental impact.

**Mechanisms**: At the molecular level, the composite will feature a polymer matrix reinforced with carbon nanotubes to provide a robust structural framework. Dispersed within the matrix are microcapsules containing a healing agent (e.g., a dicyclopentadiene-based resin) which is released upon damage detection. Embedded smart nanosensors continuously monitor strain and crack propagation, transmitting data to integrated machine learning systems. These algorithms analyze patterns to predict damage evolution and optimize the timing and extent of healing agent release. The self-healing reaction is based on a rapid oxidative polymerization that seals microcracks, restoring mechanical integrity. On a macroscale, the synergy among impact-resistant

reinforcement, autonomous healing, and adaptive response through real-time sensor feedback ensures distributed energy dissipation across the modular infrastructure system.

**Design principles**: The design of this material is based on several novel principles: (a) Multi-scale integration of reinforcement and self-healing components to address both micro and macro damage; (b) Embedding smart sensors for continuous real-time damage assessment enabling a dynamic response; (c) Use of machine learning algorithms to optimize material properties by adjusting microcapsule distribution, healing agent concentration, and sensor calibration based on stress profiles; (d) Modularity of the composite to allow for seamless integration into existing and new modular infrastructure systems; (e) Bi-directional feedback between the material performance data and the ML models, creating a self-optimizing system; and (f) A design that prioritizes environmental sustainability by reducing the need for repair and maintenance, thereby minimizing resource consumption and environmental disruption.

**Unexpected properties**: It is anticipated that the integration of self-healing microcapsules could lead to emergent non-linear stress redistribution effects, where the activation of healing not only repairs damage but also locally alters the mechanical properties to form 'reinforcement zones' that better deflect subsequent impacts. Unexpectedly, the use of smart sensors in conjunction with machine learning might reveal adaptive behavior under cyclic loading, such as a progressive increase in energy dissipation efficiency over time. Moreover, the chemical interactions triggered during the healing process may lead to the formation of microstructures that further enhance durability, potentially creating a self-improving feedback loop that evolves the material properties in response to long-term operational conditions.

**Comparison**: In contrast with traditional impact resistant materials such as ultra-high molecular weight polyethylene (UHMWPE) and standard carbon fiber composites, the proposed composite offers several advantages. Traditional materials typically lack autonomous damage remediation and require manual intervention upon damage. Quantitatively, while conventional materials may absorb impact energies in the range of 120-150 J/m², our composite aims to achieve a minimum of 200 J/m² with an expected 50% increase in tolerance. Additionally, the integrated self-healing component addresses the common limitation of irreversible damage propagation seen in standard composites. When compared to other self-healing materials available today, which may recover only 50-60% of their original mechanical strength, our design is optimized to recover up to 80% of the lost properties. Finally, the incorporation of smart sensor networks and machine learning for active real-time optimization is a significant leap ahead of passive materials, positioning our research at the forefront of smart infrastructure development.

**Novelty**: The novelty of this approach lies in its interdisciplinary fusion: it uniquely integrates advanced materials science, self-healing chemistry, embedded sensor technology, and machine learning to create a dynamic, self-optimizing composite. While current research often treats impact resistance, self-healing, and smart infrastructure as separate challenges, our proposal marries these elements into a single, cohesive system tailored for modular infrastructures. This holistic approach not only advances the field of impact resistant materials but also offers a scalable solution for long-term sustainability and resilience in infrastructure design. The adoption of machine learning for real-time damage prediction and response is particularly groundbreaking, enabling the system to continuously improve its performance with operational learning, thereby pushing the boundaries of adaptive material design.

**Expanded Hypothesis**

Integrating machine learning-optimized self-healing microcapsules into a carbon nanotube (CNT) reinforced polymer matrix is hypothesized to produce a composite material with multifaceted performance improvements—specifically, enhanced impact resistance and the ability to autonomously heal micro-damage. This hypothesis can be deconstructed into several integrated, quantifiable components:

1. Nanostructured Reinforcement:
   Incorporating CNTs at an optimized loading of around 1–2 wt% into an epoxy or similar high-performance polymer is expected to improve the tensile strength and fracture toughness by at least 30–50% compared to the pure matrix. The CNTs, with typical diameters of 10–20 nm and lengths ranging from 1–5 μm, serve as bridges across crack paths, redistributing localized stresses. Uniform dispersion

of CNTs, possibly achieved by high shear mixing or ultrasonication (processing time of 30–60 minutes at 20 kHz), is critical to avoid agglomeration and ensure optimal load transfer.

2. Self-Healing Microcapsules Integration:
   The self-healing component comprises microcapsules with a controlled size distribution (typically 50–200 µm in diameter) filled with a healing agent such as dicyclopentadiene (DCPD) or a furan-maleimide resin system. These microcapsules are strategically embedded at a volume fraction of about 8–12% within the composite. Upon experiencing a local strain exceeding a threshold of around 1–2%, the capsules rupture, releasing the healing agent which then rapidly cures via oxidative or thermally triggered polymerization. Laboratory benchmarks suggest that such systems can recover up to 80% of the mechanical strength lost due to microcrack propagation.

3. Embedded Smart Sensing and Machine Learning Optimization:
   Integrating nanoscale strain sensors—potentially using graphene-based sensors or PVDF fibers with sensitivities calibrated for strain detection in the range of 0.1–5%—enables real-time monitoring of internal stresses and micro-crack formation. The sensor data, sampled at millisecond intervals, is fed into a machine learning model. Advanced algorithms, such as convolutional neural networks or recurrent neural networks, will be trained on datasets compiled from dynamic mechanical analyses (e.g., impact tests with energy absorption measured in $J/m^2$ and cyclic fatigue tests) to predict damage evolution. The ML system will adjust parameters such as microcapsule dispersion density and healing agent concentration by correlating sensor feedback with finite element analysis (FEA) models of stress distribution. This real-time adaptive loop aims to maintain an optimal internal state that pre-empts the transition from micro-damage to catastrophic failure.

4. Synergistic Effect on Infrastructure Resilience and Sustainability:
   The hypothesis posits that the combined effect of CNT reinforcement, self-healing microcapsules, and adaptive sensor feedback will lead to a composite with an impact energy absorption capability exceeding 200 $J/m^2$. Additional expected outcomes include a yield strain improvement of approximately 30% over standard high-strength composites and an extension of structural lifetime by about 50% when implemented in modular infrastructure systems. The reduction in maintenance cycles (approximately 30% fewer interventions) directly translates to an improved environmental footprint by lessening resource consumption and reducing lifecycle energy input for repairs.

5. Validation through Simulation and Experimental Studies:
   To test the hypothesis, a multi-tiered approach will be used. Computational simulations, including FEA and computational fluid dynamics (CFD) for healing agent flow, will model the stress distribution and healing kinetics. These will be validated through controlled laboratory experiments such as drop-weight impact tests, 3-point bending, and cyclic fatigue loading. Real-time data acquisition from embedded sensors will allow calibration of the machine learning models. The iterative process will fine-tune the composite formulation, microcapsule properties, and sensor integration until the desired performance metrics are consistently achieved.

In summary, by uniting advanced materials engineering with cutting-edge machine learning techniques, this hypothesis envisions a self-optimizing composite system. The interplay between nanoscale reinforcement, autonomous healing, and real-time adaptive control is expected to elevate infrastructure resilience and sustainability beyond what is achievable with conventional composites. This multi-scale, interdisciplinary approach is anticipated to not only mitigate immediate impact damage but also delay the onset of long-term material degradation, thereby offering a robust solution for modern, smart, and sustainable infrastructure systems.

**Expanded Outcome**

The anticipated outcome of this project is a breakthrough composite material that exhibits quantifiable improvements in both impact resistance and resilience, with far-reaching implications for modular infrastructure systems. Laboratory evaluations are projected to yield the following results:

1. Enhanced Mechanical Performance:
    • Impact energy absorption is expected to reach or exceed 200 J/m²—an improvement of at least 50% over current high-strength composites like ultra-high molecular weight polyethylene (UHMWPE) and conventional carbon fiber composites which typically absorb 120–150 J/m².
    • The composite is designed to exhibit a yield strain improvement of approximately 30% relative to standard materials. For instance, if the baseline yield strain of a conventional composite is 2.5%, the enhanced material is anticipated to attain near 3.25% under similar conditions.
    • Tensile strength enhancements are projected to be within the range of 30–50% due to the integrated carbon nanotube reinforcement and optimized microcapsule distribution, verified through standardized tensile testing (ASTM D3039) and three-point bending tests (ASTM D790).

2. Self-Healing Efficiency and Recovery:
    • Post-impact microcrack testing is expected to demonstrate that the self-healing mechanism recovers up to 80% of the original mechanical properties. For example, if an impact event reduces the initial strength from 1000 MPa to 600 MPa, the healing reaction should restore the properties back to approximately 960 MPa after completion of the healing cycle.
    • The healing kinetics are projected to be rapid, with full microcrack sealing observed within 10–15 minutes at ambient conditions, facilitated by the rapid oxidative polymerization of the healing agent (e.g., a dicyclopentadiene-based resin with the formula $C_{10}H_{12}$).

3. Integration into Modular Infrastructure:
    • When incorporated into selection modules for infrastructure systems, the composite is estimated to extend the structural lifetime by 50%. In practical terms, if a typical infrastructure element has a service life of 50 years, the composite could potentially extend that life to 75 years under analogous environmental stressors.
    • A reduction in the overall maintenance cycle by roughly 30% is predicted, meaning that if a conventional unit requires repair interventions every 5 years, the new material might only necessitate maintenance every 7 years. This will lower lifecycle maintenance costs and reduce resource consumption significantly.

4. Real-Time Adaptive Response and Feedback Efficiency:
    • Embedded sensor networks are expected to deliver high-resolution strain and damage data at a sampling rate of 1 kHz. Data-driven adjustments by machine learning algorithms could successfully recalibrate microcapsule release thresholds and sensor sensitivity in situ, ensuring that the composite consistently operates at its optimal performance state even under variable loading conditions. • Resulting improvements in damage prediction accuracy and healing activation could elevate system reliability. Algorithms incorporating input from finite element analyses (e.g., mesh densities of 0.5–1.0 mm in stress-critical regions) are anticipated to reduce unexpected failure events by more than 40% compared to systems without adaptive feedback.

5. Economic and Environmental Impact:
    • Quantitative lifecycle assessments predict that extended longevity and reduced maintenance interventions will cut downtimes and material replacement requirements. It is estimated that the composite's superior performance could lower the overall carbon footprint by 25–30% across its service cycle, taking into account reduced material wastage, energy consumption for repairs, and associated logistical expenses. • The up-front cost might be higher due to advanced processing; however, when averaged over the extended operational life and reduced maintenance frequency, the cost per unit time is projected to be competitive with, or superior to, the costs of conventional materials.

6. Experimental Validation and Simulation Feedback:
    • Controlled experiments such as drop-weight impact tests, fatigue tests mimicking cyclic loading (e.g., $10^5$ cycles at 0.1–5% strain), and real-time sensor data tracking will provide a robust dataset validating these outcomes. • Parallel simulation studies using finite element modeling (with element sizes as small as 100 µm in critical areas) and computational fluid dynamics for healing agent dispersion reinforce the empirical findings, ensuring the outcomes are reproducible under scaled-up modular conditions.

In summary, the outcomes are envisioned to manifest as a composite material that not only achieves dramatic improvements in impact energy absorption and yield strain but also integrates self-healing capabilities that restore mechanical properties rapidly after damage. This translates to a material system capable of extending the lifespan of infrastructure by 50%, reducing maintenance cycles by 30%, and lowering the overall environmental impact by nearly 30%. The rigorous quantitative evaluation of these outcomes will establish a new benchmark in smart, resilient materials that combine advanced nanostructure engineering with real-time adaptive control mechanisms.

**Expanded Mechanism**

1. Molecular and Nanoscale Reinforcement
   • The polymer matrix (e.g., an epoxy system) is reinforced with 1–2 wt% carbon nanotubes (CNTs) that typically have diameters of 10–20 nm and lengths in the range of 1–5 µm. The CNTs form a percolated network that works as a stress-bridging element, redistributing local loads and preventing the rapid propagation of cracks. High shear mixing and ultrasonication (30–60 minutes at 20 kHz) are optimized to achieve uniform dispersion, creating interfacial bonds that enhance tensile strength and fracture toughness by 30–50%.
   • At the nanoscale, the CNT network also interacts with the polymer chains, promoting energy dissipation through mechanisms such as pull-out and crack deflection. Finite element simulations with mesh sizes as small as 100 µm in stress-critical regions help model these interactions, ensuring that the simulated CNT bridging effects match experimental observations.

2. Self-Healing Mechanism via Microcapsule Integration
   • Microcapsules with diameters between 50–200 µm, embedded at a volume fraction of around 8–12%, contain a healing agent such as dicyclopentadiene (DCPD, $C_{10}H_{12}$). The microcapsule shells are engineered to rupture when local strains exceed 1–2%, a threshold determined through combined mechanical testing and simulation studies.
   • Upon rupture, the healing agent is released and undergoes rapid oxidative polymerization (or thermally initiated curing) to seal microcracks. Laboratory benchmarks indicate that such reactions complete within 10–15 minutes under ambient conditions, with restoration of up to 80% of the lost mechanical strength. Real-time monitoring of the polymerization kinetics, including reaction rate constants and extent of conversion, is carried out using spectroscopic techniques like FTIR.

3. Embedded Sensor Feedback and Machine Learning Control
   • Nanoscale strain sensors (e.g., graphene-based or PVDF fibers) are integrated within the composite, having sensitivities calibrated for strain detection in the range of 0.1–5%. These sensors sample data at rates up to 1 kHz, capturing dynamic changes in the material's response to impact.
   • Sensor outputs are fed into machine learning algorithms (such as convolutional neural networks or recurrent networks) trained on datasets from controlled dynamic mechanical analysis (impact energy in J/m², cyclic fatigue data over $10^5$ cycles, etc.). These models correlate sensor readings with finite element analysis (FEA) predictions to forecast damage evolution within the composite structure.
   • <span style="color:red">The system employs a closed-loop control where the ML model can suggest adjustments to operational parameters—like recalibrating microcapsule release thresholds or modulating healing agent concentration—in real-time.</span> This adaptive response is critical for managing stress distributions across the composite, ensuring that localized damage does not escalate into catastrophic failure.

4. Multi-Scale Integration and Synergistic Damage Mitigation
   • On the microscale, the combined structural reinforcement from CNTs and the localized repair provided by the healing agent work in tandem. The CNT network dissipates stress and limits crack propagation while the healing agent fills microcrack voids, effectively establishing "reinforcement zones" that locally modulate mechanical properties.
   • On the macroscale, the composite is envisioned to behave as a self-optimizing system integrated into modular infrastructures. The sensor-ML-optimization loop continuously adjusts the composite's internal state in response to real-time load variations, which is critical for maintaining an impact energy absorption capacity of 200 J/m² or more.
   • Simulations incorporating computational fluid dynamics (CFD) for the healing agent dispersion further

refine the understanding of healing kinetics and flow dynamics within the fractured regions, ensuring that the healing reaction is uniformly effective across the material.

5. Process and Reaction Conditions
• Processing conditions, such as curing temperatures (typically 80–120°C for epoxies) and controlled mixing times, are optimized to ensure proper integration of CNTs, microcapsules, and sensor elements.
• The oxidative polymerization of DCPD is catalyzed under ambient oxygen levels, with reaction rate constants (k) experimentally determined through calorimetric studies; these constants are integrated into the ML models to predict healing time scales accurately.
• The overall system is designed so that the composite, upon damage, transitions seamlessly from absorbing impact energy to initiating repair, thereby continuously adapting its mechanical and structural properties in both short-term and long-term scenarios.

In summary, the mechanism involves a sophisticated, multi-scale interplay: nanoscale CNT reinforcement dissipates energy and bridges cracks, microcapsules deliver a healing agent on-demand via strain-triggered rupture, and an embedded sensor network provides high-frequency data to machine learning algorithms that optimize the internal state of the material. This integration results in a composite that not only withstands high-energy impacts but also actively repairs itself, ensuring robust and adaptive performance for smart, resilient infrastructure systems.

**Expanded Design Principles**

1. Multi-Scale Integration and Synergistic Architecture
• The design is structured to operate effectively across several scales—from the molecular reinforcement level to the macroscopic behavior in infrastructure settings. At the nanoscale, carbon nanotubes (CNTs) (1–2 wt%, diameters of 10–20 nm, lengths of 1–5 µm) are uniformly dispersed within an epoxy matrix using high shear mixing and ultrasonication (30–60 min at 20 kHz). This creates an interconnected network that dissipates energy and bridges crack tips.
• On the microscale, self-healing microcapsules (50–200 µm in diameter, at an 8–12% volume fraction) containing a healing agent (e.g., DCPD, $C_{10}H_{12}$) enable on-demand repair of damage. The design principle emphasizes a homogenous spatial distribution to guarantee that any emerging microcrack will encounter a healing capsule.
• At the macroscale, the composite is engineered for integration into modular infrastructure systems. The material design is standardized to allow plugging into existing systems, facilitating easy replacement or upgrading while delivering consistent resilience and durability.

2. Adaptive Sensor Network and Real-Time Feedback
• Smart sensors (e.g., graphene-based strain sensors or PVDF fibers) are embedded within the composite. These sensors are calibrated to detect strain in the range of 0.1–5% with high-frequency data sampling (up to 1 kHz) to capture dynamic impact events accurately.
• The design mandates bi-directional communication between the sensors and a machine learning control system. This model uses real-time feedback to adjust healing response thresholds—potentially modifying microcapsule rupture criteria (typically at 1–2% strain) and sensor calibration based on evolving stress maps generated from finite element analysis (FEA) simulations with element mesh sizes as small as 100 µm.

3. Closed-Loop Machine Learning Optimization
• Advanced algorithms (e.g., convolutional neural networks and recurrent neural networks) are embedded within the design framework to continuously analyze sensor data against simulation predictions. This allows the system to optimize microcapsule dispersion, healing agent concentration, and overall internal stress distribution dynamically.
• A feedback loop is implemented where operational parameters are fine-tuned. For example, if sensor inputs from cyclic fatigue tests (over $10^5$ cycles) indicate a tendency toward early microcrack formation, the ML algorithm can suggest adjustments in processing conditions such as further dispersion steps or re-calibration of sensor sensitivity.
• This real-time adaptability ensures that the composite maintains an optimal energy absorption

capacity (targeting values exceeding 200 J/m²) and a yield strain improvement of approximately 30% over traditional composites.

4. Modularity and Scalable Fabrication
    • The design incorporates modularity, emphasizing that composite panels or units can be produced with standardized dimensions and interlocking features. This modular approach simplifies integration into existing infrastructure systems and supports scalability in manufacturing.
    • Processing parameters, such as curing temperatures (80–120°C for epoxies) and mixing protocols, are optimized for batch-to-batch consistency. The design promotes reproducibility of the composite's properties—ensuring that each module exhibits similar impact resistance performance and self-healing efficiency.

5. Environmental Sustainability and Lifecycle Optimization
    • Sustainability is a guiding design principle: by integrating self-healing capabilities and reducing the need for manual repairs, the composite is engineered to lower its overall environmental footprint. Quantitatively, the design targets a 30% reduction in maintenance interventions and a 25–30% reduction in lifecycle carbon footprint.
    • The composite's design includes provisions for a closed-loop lifecycle assessment. Data on maintenance frequency, energy required for repairs, and material wastage are continuously logged and analyzed via embedded ML algorithms, supporting adjustments in future production batches to further minimize environmental impact.

6. Quantitative Performance Metrics and Validation Protocols
    • The design principles are rigorously defined using quantifiable targets: impact energy absorption should exceed 200 J/m², yield strain should improve by 30%, and self-healing should recover up to 80% of lost mechanical strength.
    • Validation protocols include standardized laboratory tests such as ASTM D3039 for tensile strength, ASTM D790 for flexural performance, drop-weight impact tests with energy absorption measurements, and fatigue tests across $10^5$ cycles.
    • Finite element modeling (with mesh sizes down to 100 $\mu$m) and computational fluid dynamics (CFD) for assessing healing agent distribution are integrated into the design process, ensuring robust correlation between simulation and experimental data.

7. Integration of Emerging Technologies
    • The design leverages the convergence of advanced materials science, nano-engineering, sensor technology, and artificial intelligence. This interdisciplinary approach promotes a self-optimizing composite system that dynamically adapts to changing operational conditions, pushing the envelope of current impact resistant material technologies.
    • The use of machine learning not only enhances damage prediction based on real-time sensor data but also facilitates continuous improvement in the composite's formulation by learning from accumulated operational data, creating a self-evolving material system.

In summary, these design principles marry precision engineering at the nanoscale with intelligent, adaptive control at the macroscale. They ensure that the composite is not only highly impact resistant and self-healing but also sustainable and readily integrable into modern, modular infrastructure systems. By rigorously quantifying performance targets and embedding real-time optimization platforms, the design establishes a solid framework for the development of next-generation smart resilient materials.

**Expanded Unexpected Properties**

1. Emergent Non-linear Stress Redistribution and "Reinforcement Zones":
    The integration of self-healing microcapsules may lead to the formation of localized "reinforcement zones" that are not explicitly designed but emerge under specific stress conditions. When microcapsules rupture and release their healing agent (e.g., dicyclopentadiene, $C_{10}H_{12}$), the exothermic oxidative polymerization can increase cross-link density locally. This results in micro-regions with a 10–15% higher Young's modulus relative to the surrounding matrix. Finite element analyses incorporating

these zones—with mesh elements as small as 100 $\mu m$—predict that these regions can effectively divert load and enhance the overall impact resistance by redistributing stresses non-linearly across the composite.

2. Adaptive Behavior Under Cyclic Loading:
Embedded smart sensors, operating at sampling rates up to 1 kHz, may capture subtle variations in strain and acoustic emissions during cyclic loading tests (e.g., $10^5$ cycles). Machine learning algorithms, trained on dynamic mechanical data, might reveal that repeated micro-healing cycles induce a self-adaptive increase in energy dissipation. Over the course of 50,000 cycles, damping factors measured via dynamic mechanical analysis (DMA) could increase from an initial value of 0.25 to approximately 0.30, indicating a 20–30% improvement in energy dissipation efficiency. This emergent behavior, where the material "learns" to better redistribute energy, was not initially predicted by static design models.

3. Unforeseen Microstructural Evolution from Healing Chemistry:
The chemical reactions initiated during the healing process could lead to secondary microstructural modifications. High-resolution scanning electron microscopy (SEM) and transmission electron microscopy (TEM) analyses may reveal the formation of nanostructured interphases—fibrillar or lamellar structures approximately 100 nm in scale—that form when healing agents interact with residual functional groups on the carbon nanotubes or the matrix. These microstructures can enhance the interfacial bonding beyond anticipated levels, potentially increasing fracture toughness by an additional 10–20% relative to base values achieved through CNT reinforcement alone.

4. Self-Improving Long-Term Feedback Loop:
<span style="color:red">Over prolonged operational use, a closed-loop feedback system combining sensor data and machine learning optimization may trigger a self-improving response.</span> As the composite undergoes repeated loading and healing, gradual changes in sensor outputs (e.g., increased frequency of acoustic emissions or subtle shifts in thermal profiles) could signal evolving microstructural properties. In response, the ML algorithm might adjust healing agent release thresholds or recalibrate sensor sensitivity, contributing to a cumulative improvement in fatigue resistance by an estimated 15–20%. This adaptive process creates a self-evolving material system that dynamically optimizes its performance, a capability that was an unexpected yet highly advantageous emergent property.

5. Potential for Phase Separation and Localized Hardening:
Under certain processing or operational conditions, localized phase separation between the polymer matrix and healing agent residues might occur. Analysis via differential scanning calorimetry (DSC) and X-ray diffraction (XRD) could reveal the emergence of crystalline phases within an otherwise amorphous matrix. Such microstructural hardening can induce anisotropic mechanical properties, potentially improving impact resistance in directions perpendicular to an impact load while slightly compromising ductility. This directional strength adaptation is an unpredicted property that warrants further investigation.

In summary, while the primary design targets enhanced impact resistance and self-healing capabilities, the composite is expected to exhibit several unforeseen properties. These include local stiffness enhancement via reinforcement zones, adaptive energy dissipation under cyclic loading, secondary microstructural evolution from healing reactions, a long-term self-improving feedback loop, and possible phase separation effects leading to localized hardening. Each of these emergent phenomena not only contributes to the overall resilience of the material but also provides new pathways for further optimizing the design of smart, impact resistant composites for modular infrastructure systems.

**Expanded Comparison**

1. Quantitative Mechanical Performance
• Conventional materials such as ultra-high molecular weight polyethylene (UHMWPE) and standard carbon fiber composites typically absorb impact energies in the range of 120–150 J/m². In contrast, our proposed composite is rigorously engineered to exceed 200 J/m², representing at least a 50% improvement in energy absorption. Moreover, while traditional composites may exhibit a yield strain

around 2.5%, our design aims for a yield strain improvement of approximately 30%, reaching up to 3.25% under standardized ASTM D3039 tensile tests. These quantifiable performance gains are rooted in both the nanostructured reinforcement (CNT addition at 1–2 wt%) and the dynamic energy-dissipating mechanisms inherent in our design.

2. Self-Healing Capability and Recovery Rates
  • Traditional self-healing materials that have been reported in the literature typically recover only 50–60% of their original mechanical strength following damage. By integrating microcapsules containing dicyclopentadiene (DCPD, $C_{10}H_{12}$) at an optimized volume fraction of 8–12%, our composite is designed to recover up to 80% of its lost properties within 10–15 minutes of damage occurrence, as tracked in real-time by embedded sensors. This rapid and significant healing response not only surpasses the performance of conventional systems but also directly mitigates irreversible damage propagation common in standard high-strength composites.

3. Active Real-Time Sensing and Adaptive Feedback
  • Unlike passive materials that respond to damage post-facto, our composite incorporates an integrated network of strain sensors (e.g., graphene-based sensors or PVDF fibers with sensitivity in the 0.1–5% strain range, sampling at up to 1 kHz) coupled with machine learning algorithms. This configuration offers a dynamic, closed-loop system that continuously monitors internal stress distributions and forecasts potential damage using finite element analysis with mesh sizes as small as 100 $\mu$m. Such real-time optimization not only allows for immediate microcapsule activation at critical thresholds (around 1–2% strain) but also adjusts healing responses based on evolving damage patterns—an advantage that traditional materials, which lack this feedback capability, cannot match.

4. Structural Integration and Modular Application
  • Traditional impact resistant materials and self-healing systems are often deployed as monolithic entities with limited scalability in modular infrastructure applications. Our composite is specifically designed with standardized interlocking modules that can seamlessly integrate into existing infrastructure systems. This modularity, combined with the precise tailoring of processing conditions (such as curing temperatures between 80–120°C), promotes uniform performance across large-scale implementations. The modular design is also conducive to systematic quality control, ensuring that each unit consistently meets stringent performance targets, an engineering feat less emphasized in conventional material designs.

5. Economic and Lifecycle Benefits
  • On a lifecycle basis, our composite is projected to extend the operational lifetime of infrastructure elements by 50% (e.g., from 50 to 75 years) and reduce maintenance interventions by around 30% compared to the recurring manual repairs needed for UHMWPE and carbon fiber composites. This translates into a lower lifecycle carbon footprint (estimated reductions of 25–30%), decreased downtime, and less resource-intensive replacement cycles. While the upfront processing might incur higher costs due to advanced dispersion techniques and sensor integration, the overall economic benefits—when quantified as cost per operational year—position our composite as a competitive alternative to conventional materials.

6. Comparison with State-of-the-Art Self-Healing Systems
  • Even among self-healing materials, many current systems operate with lower microcapsule volume fractions (typically 5–10%) and rely on passive healing triggers that do not account for variable load conditions. Our approach not only increases the microcapsule volume fraction to an optimized 8–12% but also integrates smart sensor feedback and machine learning-driven adaptive control. This results in a self-healing process that is both faster (healing reaction complete within 10–15 minutes) and more effective (restoring up to 80% of mechanical strength) than existing systems that are often limited to slower, less robust responses.

In summary, the proposed composite demonstrates clear and quantifiable advantages over traditional and current state-of-the-art materials. With superior impact energy absorption (exceeding 200 J/m²), rapid and near-complete self-healing capabilities, adaptive real-time monitoring, and scalable modular integration, our design

positions itself as a transformative solution for smart, resilient infrastructure. These innovations not only enhance mechanical performance but also offer significant improvements in economic and environmental sustainability relative to existing impact resistant and self-healing material technologies.

**Expanded Novelty**

This research proposal is novel on multiple fronts by synergistically integrating advanced materials science, self-healing chemistry, embedded sensor technology, and machine learning-based optimization. Unlike prior research that addresses these domains separately, this project presents a unified, self-optimizing composite system designed specifically for modular infrastructure applications.

1. Interdisciplinary Fusion:
   • The concept amalgamates carbon nanotube (CNT) reinforcement, microencapsulated healing agents, and smart sensor feedback into a single material system. This integration allows simultaneous enhancement of impact resistance (targeting >200 J/m² energy absorption) and autonomous self-healing (with an 80% recovery of mechanical strength post-damage).
   • The novelty emerges from combining nanoscale engineering techniques (e.g., high-shear ultrasonic dispersion of CNTs at 1–2 wt% achieving 10–20 nm diameters and 1–5 µm lengths) with microscale self-healing strategies (microcapsules of 50–200 µm containing dicyclopentadiene, $C_{10}H_{12}$, at an 8–12% volume fraction) and the adoption of high-resolution strain sensors (sampling at 1 kHz).

2. Adaptive Closed-Loop Machine Learning Control:
   • Embedded sensor networks (comprising graphene-based or PVDF sensors) continuously monitor strain and damage metrics (0.1–5% strain detection range) and deliver real-time data to machine learning algorithms such as convolutional neural networks or recurrent networks.
   • This adaptive loop is unprecedented in self-healing composites, as it uses continuous input from finite element analysis (with mesh elements down to 100 µm) to dynamically adjust healing parameters. For instance, the rupture thresholds of microcapsules (triggered at 1–2% local strain) can be recalibrated in real time, ensuring optimal activation of the healing mechanism based on actual operational stresses.

3. Scalability and Modular Integration:
   • The composite is specifically engineered to be modular, allowing standardization into interlocking units for scalable deployment in infrastructure. This addresses a significant gap in current material designs that often focus on laboratory-scale prototypes rather than full-scale implementation.
   • The modular design, combined with consistent processing conditions (e.g., curing temperatures maintained between 80–120°C for epoxy systems), enables the transition from bench-scale evaluation to field implementation, thus meeting real-world demands for long-term resilience and maintenance reduction (targeting a 30% decrease in interventions).

4. Enhanced Predictive Capabilities Through Data-Driven Feedback:
   • The integration of FTIR, SEM, and real-time DMA data into machine learning frameworks offers an unprecedented level of predictive control. The system not only detects damage events but also forecasts progression, leveraging datasets from dynamic mechanical analyses and cyclic fatigue tests (up to $10^5$ cycles).
   • This ability to predict micro-damage evolution and adjust internal healing kinetics (e.g., by modulating oxidative polymerization rate constants during DCPD curing) is a significant departure from static, pre-programmed self-healing systems.

5. Emergent Self-Improving Behavior:
   • A particularly groundbreaking aspect is the development of a self-improving feedback loop, where repeated operational cycles and sensor-driven adjustments lead to the formation of reinforcement zones. These areas exhibit locally enhanced stiffness (10–15% higher Young's modulus) and improved energy dissipation characteristics. • This emergent phenomenon, supported by simulation results and experimental validation, indicates that the material system not only heals but also adapts over time to achieve higher resilience—an adaptive capability that pushes the boundaries of current material design paradigms.

6. Quantitative Performance Milestones:
   • The proposal sets stringent quantitative targets: a 50% improvement in impact energy absorption (achieving >200 J/m²), a 30% increase in yield strain (from 2.5% to 3.25%), and a reduction in maintenance cycles by approximately 30%.
   • These metrics, supported by rigorous testing protocols (ASTM D3039, ASTM D790, drop-weight impact tests, and fatigue testing), provide clear, objective benchmarks that underscore the novelty and practicality of the approach.

Overall, the innovative integration of real-time sensor feedback with machine learning-controlled adaptive healing within a nanoscale reinforced, modular composite framework marks a decisive shift from conventional passive materials to smart, self-optimizing systems. This comprehensive, interdisciplinary strategy positions the proposed research at the forefront of next-generation resilient infrastructure, promising a transformative impact by combining adaptive material properties with scalable, sustainable designs.

Below is a comprehensive synthesis of the proposal, followed by a critical review and detailed outlines for pursuing key scientific questions through multiscale modeling, AI models, and synthetic biology.

**Summary:**

This proposal presents an innovative composite material designed for impact resistance and resilience in modular infrastructures by integrating carbon nanotube (CNT)-reinforced polymer matrices with self-healing microcapsules, embedded sensor networks, and closed-loop machine learning (ML) controls. The material leverages nanoscale CNT networks to enhance tensile strength and fracture toughness, while microscale microcapsules containing a healing agent (e.g., dicyclopentadiene) are triggered upon exceeding a critical strain to autonomously polymerize and restore mechanical properties. Smart sensors embedded throughout the composite collect high-frequency strain data that feed into ML algorithms (using convolutional or recurrent neural network architectures) to predict damage evolution and adapt healing parameters in real time. With rigorous targets—surpassing 200 J/m² in impact energy absorbed, showing a 30% improvement in yield strain, and reducing maintenance cycles—the design is optimized via multi-scale integration principles and simulation techniques (finite element analysis, computational fluid dynamics) to enable scalable, sustainable, and adaptive infrastructure materials.

**Critical Scientific Review:**

The proposal's strength lies in its interdisciplinary approach, bridging advanced nanomaterial engineering, real-time sensor networks, machine learning optimization, and self-healing chemistry to achieve unprecedented levels of impact resistance and resilience. The well-defined quantitative targets for mechanical performance and healing efficiency, coupled with standardized testing (ASTM protocols, drop-weight impact tests, and cyclic fatigue experiments), impart robust scientific merit and a clear path to scalability in modular infrastructure applications. Moreover, the integration of machine learning to provide adaptive control in response to live sensor data is a particularly modern advancement. However, the complexity of integrating multiscale phenomena—from molecular interactions to macroscopic structural behavior—presents challenges in model validation and real-time data processing. While the proposed closed-loop system is innovative, uncertainties remain regarding sensor durability, data latency in harsh environments, and the consistency of microcapsule rupture kinetics under variable operating conditions. To strengthen the proposal, it is recommended that detailed validation strategies include both high-fidelity simulations and accelerated failure tests, and that contingency plans be developed for potential discrepancies between simulation and experimental data, particularly regarding the emergent non-linear properties and reinforcement zone formation.

**(1) Most Impactful Scientific Question for Multiscale Modeling:**

Scientific Question: How can we optimize the interplay among nanoscale CNT reinforcement, microscale healing via microcapsules, and macroscale structural behavior to maximize impact absorption and self-healing efficiency in the composite?

Key Steps:

• Define Representative Volume Elements (RVE): Develop multi-scale representative models capturing the CNT distribution in the polymer matrix, the dispersion and rupture mechanics of microcapsules, and sensor placement.

• Molecular and Micromechanical Simulation: Use atomistic or coarse-grained simulations to understand CNT-polymer interactions and healing agent dynamics at the nanoscale, then integrate these insights into micromechanical finite element models.

• Coupled Physical Modeling: Construct coupled finite element (FEA) and computational fluid dynamics (CFD) simulations that mimic the release and polymerization kinetics of the healing agent following microcapsule rupture. Employ mesh refinement (down to 100 $\mu$m) in high-stress regions.

• Validation Against Experimental Data: Iteratively refine model parameters by comparing simulation outputs (e.g., stress distribution, healing kinetics) with laboratory experiments including impact and fatigue tests.

• Sensitivity and Optimization Analysis: Perform a sensitivity analysis to determine critical parameters (e.g., CNT loading, microcapsule volume fraction, rupture threshold) and optimize them to achieve target impact energy absorption and self-healing recovery.

• Integration into Macroscale Infrastructure Models: Extend the models to simulate full-scale modular components under real-world loading scenarios, incorporating feedback from sensor algorithms.

Unique Aspects: The novelty lies in linking nanoscale material behavior with microscale healing and macroscale structural performance in an integrated simulation framework, which can predict emergent phenomena such as reinforcement zones and adaptive energy dissipation across scales.

**(2) Most Impactful Scientific Question for AI Models:**

Scientific Question: Can real-time machine learning models accurately predict damage evolution and optimize healing response parameters within a self-healing composite, using high-frequency sensor data in dynamic impact scenarios?

Key Steps:

• Data Collection and Preprocessing: Assemble a comprehensive dataset from dynamic mechanical tests—impact energy absorption, cyclic fatigue data, and real-time sensor outputs sampling at up to 1 kHz. Include simulated data from FE and CFD models to cover a wide parameter space.

• Model Architecture Selection: Experiment with hybrid deep learning architectures that combine spatial feature extraction (using convolutional neural networks) with temporal dynamics (using recurrent neural networks or LSTMs). Consider integrating physics-informed neural networks to incorporate known material behavior.

• Model Training and Validation: Train the models on historical and simulated datasets, calibrate sensor readings to known damage events, and validate predictions against controlled experimental tests. Use cross-validation techniques to assess generalizability.

• Online Adaptive Inference: Develop an inference pipeline that processes sensor data in near-real-time to predict impending damage events and adjust healing parameters (e.g., microcapsule activation thresholds) dynamically. Incorporate feedback loops where the model learns continuously from new data.

• Integration with Control Systems: Connect the ML output to the embedded sensor network and control modules that can trigger real-time adjustments in microcapsule rupture criteria or healing agent release.

• Simulation and Field Testing: Initially validate the AI system in virtual setups replicating different failure modes, then progress to pilot field testing in scaled modular infrastructure components.

Unique Aspects: This approach integrates real-time high-frequency sensor data with deep learning models to not just detect damage but to forecast its evolution and directly control healing mechanisms, creating an autonomous, continuously learning material system.

**(3) Most Impactful Scientific Question for Synthetic Biology:**

Scientific Question: Can synthetic biology be employed to engineer bio-inspired self-healing agents or living materials that autonomously produce repair agents, thereby extending the functionality and lifespan of composite structures?

Key Steps:

• Identification of Biological Systems: Investigate organisms (e.g., bacteria or yeast) naturally capable of producing polymerizable compounds or extracellular polymeric substances. Identify genetic pathways relevant to fast, autonomous polymerization processes.

• Genetic Engineering: Engineer these organisms by editing genetic circuits to optimize the production, deposition, and rapid curing of bio-based healing agents compatible with the polymer matrix. This may involve the expression of enzymes that catalyze oxidative polymerization reactions.

• Integration with Material Matrix: Develop protocols for encapsulating these engineered organisms or their secreted products into microcapsule-like carriers that can be incorporated into the composite material without compromising mechanical properties.

• Controlled Activation and Sensing: Design inducible systems using synthetic promoters that trigger healing agent production upon mechanical stress (using mechanosensitive promoters) or in response to chemical signals from damage.

• Testing in Simulated Environments: Validate the performance under controlled laboratory conditions using cyclic loading and impact tests, monitoring the healing efficiency and long-term viability of the biological components.

• Environmental and Lifecycle Assessment: Assess the sustainability, potential self-regenerative capacity, and environmental impacts of integrating synthetic biology into composite materials. Develop life-cycle models to compare with traditional self-healing systems.

Unique Aspects: Utilizing engineered microorganisms or bio-inspired systems to create living self-healing composites is an avant-garde approach. It promises an on-demand production of healing agents that can continuously replenish themselves, potentially leading to systems that not only heal but adapt biologically over time—thereby opening up new frontiers in sustainable and regenerative material design.

In conclusion, this proposal deftly combines multiscale materials engineering, real-time data-driven adaptive control, and potentially transformative synthetic biology approaches to create next-generation impact resistant, self-healing composite materials for resilient infrastructure. The outlined scientific questions and respective detailed methodologies illustrate how each discipline—multiscale modeling, AI, and synthetic biology—can uniquely contribute to advancing the state-of-the-art in smart materials without a direct assessment of novelty or feasibility ratings.

}

\end{document}